\documentclass{article} 
\usepackage{iclr2027_conference,times}

\usepackage{amsmath,amsfonts,bm}

\def\eqref#1{equation~\ref{#1}}

\def\1{\bm{1}}

\DeclareMathAlphabet{\mathsfit}{\encodingdefault}{\sfdefault}{m}{sl}
\SetMathAlphabet{\mathsfit}{bold}{\encodingdefault}{\sfdefault}{bx}{n}

\usepackage{iftex}
\ifPDFTeX
  \usepackage[utf8]{inputenc}
\fi
\usepackage[T1]{fontenc}
\usepackage{fix-cm} 
\usepackage{inconsolata} 
\usepackage{booktabs}
\usepackage{amssymb}
\usepackage{nicefrac}
\usepackage{microtype}
\usepackage{xcolor}
\usepackage{graphicx}
\usepackage{array}
\usepackage{tabularx}
\newcolumntype{Y}{>{\raggedright\arraybackslash}X}
\newcolumntype{Z}{>{\centering\arraybackslash}X}
\newcommand{\adeappendixtablestyle}{%
  \small\setlength{\tabcolsep}{5pt}\renewcommand{\arraystretch}{1.15}}
\usepackage{enumitem}
\usepackage{float}
\usepackage{placeins}
\usepackage{tikz}
\usepackage{algpseudocode}

\usepackage{tcolorbox}
\tcbuselibrary{skins,breakable,listings}
\definecolor{adeboxink}{HTML}{263747}
\definecolor{adeblue}{HTML}{436B94}
\definecolor{adeteal}{HTML}{397B78}
\definecolor{adeviolet}{HTML}{75638D}
\tcbset{adepanel/.style={
  enhanced,breakable,lines before break=8,boxrule=0.45pt,leftrule=2pt,arc=2pt,
  left=9pt,right=9pt,top=7pt,bottom=7pt,
  before skip=10pt,after skip=10pt,
  coltext=adeboxink,coltitle=adeboxink,
  fonttitle=\sffamily\bfseries\small,
  toptitle=5pt,bottomtitle=5pt,
  fontupper=\small
}}
\newtcolorbox{adepromptbox}[2][adeblue]{
  adepanel,colback=#1!3!white,colframe=#1!45!white,
  colbacktitle=#1!11!white,title={#2}
}
\newtcolorbox{adecontractbox}[1]{
  adepanel,boxrule=0.6pt,leftrule=0.6pt,arc=0pt,
  colback=white,colframe=adeboxink!45!white,
  colbacktitle=adeboxink!7!white,title={Task contract: #1}
}
\definecolor{adecodeink}{HTML}{28323C}
\definecolor{adecodekeyword}{HTML}{355F87}
\definecolor{adecodestring}{HTML}{476B58}
\definecolor{adecodecomment}{HTML}{68727C}
\lstdefinestyle{adecode}{
  basicstyle=\ttfamily\fontsize{8.5}{10.5}\selectfont\color{adecodeink},
  keywordstyle=\color{adecodekeyword}\bfseries,
  commentstyle=\color{adecodecomment},
  stringstyle=\color{adecodestring},
  columns=fullflexible,keepspaces=true,showstringspaces=false,
  breaklines=true,breakatwhitespace=true,breakautoindent=true,
  breakindent=1em,upquote=true,tabsize=4,
  aboveskip=0pt,belowskip=0pt
}
\tcbset{adecodepanel/.style={
  enhanced,breakable,lines before break=6,
  boxrule=0.35pt,arc=0pt,
  colback=black!2!white,colframe=black!20!white,
  colbacktitle=black!5!white,coltitle=adecodeink,
  fonttitle=\sffamily\small\bfseries,
  left=9pt,right=9pt,top=7pt,bottom=7pt,
  toptitle=5pt,bottomtitle=5pt,
  before skip=9pt,after skip=9pt
}}
\newtcblisting{adecodebox}[2][adeteal]{
  adecodepanel,title={#2},title after break={#2 (continued)},
  listing only,listing options={style=adecode}
}
\newtcblisting{adestrategycode}[1]{
  adecodepanel,title={#1},title after break={#1 (continued)},
  listing only,listing options={style=adecode,language=Python}
}

\newfloat{algorithm}{!t}{loa}
\renewcommand{\thealgorithm}{\arabic{algorithm}}

\newcommand{\name}{\textsc{ADE}}
\newcommand{\adecomponent}[1]{\textit{#1}}

\newcommand{\code}[1]{\texttt{#1}}

\newcolumntype{L}[1]{>{\raggedright\arraybackslash}p{#1}}
\newcolumntype{C}[1]{>{\centering\arraybackslash}p{#1}}

\title{Agentic Data Engineering for LLMs: System Design and a Controlled Empirical Study}

\newcommand{\adeauthormark}[1]{%
  \textsuperscript{\normalfont\ensuremath{\vphantom{1,*\dagger}#1}}}
\newcommand{\adeauthorline}[1]{%
  \begingroup
  \sbox0{#1}%
  \ifdim\wd0>\linewidth
    \resizebox{\linewidth}{!}{\usebox0}%
  \else
    \usebox0%
  \fi
  \endgroup}
\author{%
\parbox[t]{\dimexpr\textwidth-2\tabcolsep\relax}{%
\centering
\adeauthorline{%
\textbf{Ruomeng Ding}\adeauthormark{1,*}\enspace
\textbf{Qianli Shen}\adeauthormark{2}\enspace
\textbf{Zhaoyang Han}\adeauthormark{3,*}\enspace
\textbf{Meixin Chen}\adeauthormark{2}\enspace
\textbf{Daoyuan Chen}\adeauthormark{2,\dagger}\enspace
\textbf{Yaliang Li}\adeauthormark{2,\dagger}}\\[0.65em]
\adeauthorline{\normalfont\small
\adeauthormark{1}University of North Carolina at Chapel Hill\quad
\adeauthormark{2}Alibaba Group\quad
\adeauthormark{3}Peking University}%
}}

\usepackage{xurl}
\usepackage{hyperref}
\newcommand{\Qianli}[2][]{}
\newcommand{\qianli}[1]{}
\newcommand{\ruomeng}[1]{}

\usepackage{fancyhdr}
\iclrfinalcopy
\begin{document}

\maketitle
\begingroup
\renewcommand{\thefootnote}{\fnsymbol{footnote}}
\footnotetext[1]{Work done during internships at Alibaba Group.}
\footnotetext[2]{Correspondence to:
\texttt{\{daoyuanchen.cdy, yaliang.li\}@alibaba-inc.com}.}
\endgroup
\fancyhead{}
\vspace{-10pt}


\begin{abstract}
LLM data engineering requires iterative experimentation to determine how data
selection, rewards, and curricula affect model learning, making it a natural
setting for AI-for-AI (AI4AI) research. Can agents use experimental feedback to progressively improve strategies, and do the resulting strategies improve
target-model performance on held-out data?
We study how agents use experimental feedback to develop data-engineering
strategies under fixed training and evaluation protocols.
Our \emph{Agentic Data Engineering} (\name{}) framework lets agents pursue,
revise, or combine research directions using cumulative findings.
Strategies are selected by in-loop scores and evaluated on held-out benchmarks.
Across four tasks---math and code data selection, math reward design, and
math curriculum learning---selected strategies achieve absolute gains over generic
baselines of up to 13.34\% on AIME25 Pass@32 and 5.85\% averaged across four
held-out benchmarks under fixed training conditions. In-loop scores improve
over successive experiments, while held-out trajectories show that later
in-loop gains do not always improve generalization. Comparisons of research
organization and feedback access favor combining evidence analysis with
cumulative findings. Research trajectories show how earlier findings inform
strategy revisions and combinations across research directions. 
Our code is released at {\urlstyle{same}\url{https://github.com/datajuicer/agentic-data-engineering}}.
\end{abstract}

\section{Introduction}
\label{sec:introduction}

AI-for-AI (AI4AI) seeks to use AI systems to automate and improve the research
process behind model development, including the design of training data,
learning algorithms, and feedback \citep{ai4aisurvey2026}.
LLM data engineering offers a concrete setting for studying this research
capability. Data selection determines which examples a model learns from,
curriculum design determines their allocation over training, and reward design
determines the feedback assigned to model behavior.
The value of these interventions depends on what a target model learns under
given training conditions, making downstream experiments central to strategy
development. Recent work on data-centric improvement likewise frames automation
as an iterative loop in which evaluation guides successive data interventions
\citep{dcrisurvey2026}.
Developing such strategies requires agents to propose changes, interpret their
effects, and decide what to investigate next. At the same time, a fixed training
paradigm makes these consequential research decisions experimentally comparable.

Assessing this research capability requires examining both how strategies
develop and their resulting effectiveness, motivating two questions.
{\textbf{Question 1}: \textit{Can agents use experimental feedback to guide subsequent
research decisions and develop progressively stronger strategies?}}
Autonomy over execution is not sufficient for AI4AI; the relevant autonomy
is over evidence-conditioned research decisions
\citep{whatismissing2026}. A strong final strategy may reflect best-of-many
sampling rather than feedback-guided research.
{\textbf{Question 2}: \textit{Can agent-developed strategies improve target-model performance
on evaluation data withheld from the research process?}} Repeatedly using
evaluation feedback to develop and select strategies can favor the particular
examples or metrics exposed during search \citep{rethinkingharnesseval2026,rrsi2026}.
Answering these questions requires assessing progress over experiments,
the links between prior feedback and subsequent decisions, and held-out
effectiveness.

Existing AI4AI systems differ in intervention scope, evaluation boundaries,
and control over research decisions. AutoResearch lets the agent modify and
run training code while keeping the evaluator fixed \citep{autoresearch2026}.
RSIBench-Data separates researcher agents from fixed training infrastructure,
but checkpoint selection and final evaluation use the same task subset
\citep{rsibenchdata2026}. DataMaster uses a predefined score balancing branch
performance and exploration to choose which agent-proposed branch to execute
next \citep{datamaster2026}. Designs for iterative search and evidence reuse
also vary, spanning reflective experiment archives, shared findings, and
reusable diagnostic skills \citep{poise2026,datamaster2026,evotrainer2026}.
Our study examines agent-directed research under fixed experimental
protocols: Agents have full authority to decide which directions to pursue,
revise, or combine, while the Harness enforces a fixed protocol and training
setup (Table~\ref{tab:study-boundaries}).

\newsavebox{\adeintrotable}
\begin{table}[!t]
  \centering
  \small
  \vspace{-2em}
  \begin{minipage}[t]{0.56\linewidth}
    \centering
    \vspace{0pt}
    \caption{Scope and research boundaries.}
    \label{tab:study-boundaries}
  \sbox{\adeintrotable}{%
    \fontsize{8.5}{10}\selectfont%
    \setlength{\tabcolsep}{4pt}%
    \renewcommand{\arraystretch}{1.0}%
    \resizebox{0.95\linewidth}{!}{%
    \begin{tabularx}{\linewidth}{@{}lY@{}}
      \toprule
      \textbf{Dimension} & \textbf{Our Work (\name{})} \\
      \midrule
      Intervention Scope
        & Data Selection, Reward Design,\newline and Curriculum Learning;\newline
          Fixed Model and Training Paradigm. \\
      \addlinespace[4pt]
      \textcolor{adeteal}{\textbf{Q1:}} Research Capability
        & \textcolor{adeteal}{\textbf{P1:}} Give agents control over\newline research decisions under\newline fixed experimental protocols. \\
      \addlinespace[4pt]
      \textcolor{adeblue}{\textbf{Q2:}} Effectiveness
        & \textcolor{adeblue}{\textbf{P2:}} Select in-loop; evaluate held-out. \\
      \bottomrule
    \end{tabularx}%
    }%
  }
    \usebox{\adeintrotable}
  \end{minipage}\hfill
  \begin{minipage}[t]{0.4\linewidth}
    \vspace{0pt}
    \centering
    \resizebox{\linewidth}{!}{%
\begin{tikzpicture}[x=1cm,y=1cm,
  every node/.style={font=\fontsize{8.5}{9.5}\selectfont,align=center},
  role/.style={draw=adeboxink!65,rounded corners=2pt,fill=white,
    inner xsep=4pt,inner ysep=4pt},
  flow/.style={->,>=stealth,line width=0.7pt,draw=adeboxink},
  lab/.style={font=\fontsize{7.5}{8.5}\selectfont,fill=white,inner sep=1pt}]
  \path[use as bounding box] (0,0) rectangle (6.8,4.05);
  \draw[violet!55!black,dashed,rounded corners=2pt]
    (0.08,1.55) rectangle (6.72,2.57);
  \node[lab,text=violet!75!black] at (3.4,1.71) {Harness-enforced protocol};
  \node[role,fill=adeteal!7] (agent) at (3.4,3.65) {Research agents};
  \node[role] (training) at (0.98,2.12) {Model\\training};
  \node[role] (eval) at (5.82,2.12) {In-loop\\evaluation};
  \draw[flow] (agent.south west) -- (training.north);
  \node[lab] at (1.05,3.08) {Research decisions};
  \draw[flow] (training.east) -- node[lab,above] {Model checkpoint} (eval.west);
  \draw[flow,draw=adeteal] (eval.north) -- (agent.south east);
  \node[lab,text=adeteal] at (5.95,3.08) {Feedback};
  \node[text=adeteal,font=\fontsize{8}{9}\selectfont\bfseries]
    at (3.4,2.87) {Q1: Research\\capability};
  \draw[adeboxink!30,dashed] (0.08,1.43) -- (6.72,1.43);
  \node[text=adeblue,font=\fontsize{8}{9}\selectfont\bfseries]
    at (2.4,1.1) {Q2: Effectiveness};
  \node[role,draw=adeblue,fill=adeblue!6] (held) at (2.4,0.4)
    {Held-out\\evaluation};
  \draw[flow,draw=adeblue] (eval.south) -- (5.82,0.4) --
    node[lab,above,text=adeblue] {Frozen selection} (held.east);
\end{tikzpicture}}\par
    \makeatletter\def\@captype{figure}\makeatother
      \vspace{-0.4em}
    \caption{Research loop and evaluation.}
    \label{fig:intro-teaser}
  \end{minipage}
  \vspace{-1.5em}
\end{table}

We introduce \emph{Agentic Data Engineering} (\name{}) through two principles
(Figure~\ref{fig:intro-teaser}).
{\textbf{Principle 1}: \textit{Give agents control over research decisions under fixed experimental protocols ($\rightarrow${Q1}).}}
Agents use cumulative memory of prior findings to choose subsequent research
directions. The Harness validates actions and controls information access;
the Engine executes admitted experiments under task-specific conditions.
We examine how Agents use feedback to develop stronger strategies.
{\textbf{Principle 2}: \textit{Select in-loop; evaluate held-out ($\rightarrow${Q2}).}}
Strategies are selected using only in-loop scores; the selected checkpoint is
then frozen and evaluated on disjoint held-out data. These results never enter
the research process, providing an independent assessment of effectiveness.

Across four tasks---math and code data selection, math reward design, and
math curriculum learning---\name{} improves target-model performance over generic
baselines, with absolute gains up to 13.34\% on AIME25 Pass@32 and 5.85\% averaged across four held-out benchmarks. Successive experiments
progressively improve strategies: for math data selection, mean
in-loop Pass@32 rises from 45.56\% for the first completed strategies to a final
54.44\%. Ablations of research organization and feedback access favor the complete
\name{} system, which exceeds the strongest alternative per metric by 6.67\%
in in-loop Pass@32 and 2.62\% averaged across four held-out
benchmarks. Research trajectories link earlier findings to later revisions
and combinations.

Our contributions are threefold:
\begin{itemize}[leftmargin=8pt,label=\textbullet,itemindent=0pt,itemsep=0.1em,topsep=0.3em]
  \item We introduce \name{}, giving agents control over research decisions
    under fixed experimental protocols and separating in-loop selection
    from held-out evaluation.
  \item We evaluate progressive strategy improvement and held-out
    effectiveness across four tasks under fixed training conditions,
    examining both sequential and parallel research.
  \item We assess research organization and feedback use through
    ablations and trace how earlier findings inform subsequent
    strategy revisions and combinations.
\end{itemize}

\vspace{-1em}
\section{Related Work}
\label{related-work}
\vspace{-0.5em}
\textbf{Agentic data engineering and model training.}
DataEnvGym and Curation-Bench study feedback-driven data generation and
selection \citep{dataenvgym2025,curationbench2026}. DataEvolver evolves
preparation pipelines; Autodata also optimizes the data scientist's prompts
\citep{dataevolver2026,autodata2026}. DataMaster and DataAgent investigate
data research with fixed training stacks \citep{datamaster2026,exploringade2026}.
FT-Dojo and AutoTrainess cover broader fine-tuning workflows
\citep{ftdojo2026,autotrainess2026}; A-Evolve-Training and EvoTrainer revise
research policies and diagnostics \citep{aevolvetraining2026,evotrainer2026}.

\textbf{Search and adaptation in AI4AI systems.}
AutoResearch, MARS, and POISE study iterative experiments, reflective search,
and algorithm archives \citep{autoresearch2026,mars2026,poise2026}.
Frontis-MA1 trains research operators; HELIX evolves harnesses and exports
training evidence \citep{frontisma12026,helix2026}.
STOP and Darwin G\"odel Machine modify improver code; ScienceBuddy alternates
harness revision and model training \citep{stop2024,dgm2026,sciencebuddy2026}.
\name{} fixes researcher models and the Harness to examine evidence-guided
strategy revision within data engineering.

\textbf{Evaluating AI4AI research.}
PostTrainBench examines contamination; RSIBench-Data distinguishes discoveries
from sustained improvement \citep{posttrainbench2026,rsibenchdata2026}.
\citet{whatismissing2026} find execution gains without high-level strategy
revision. Heuresis assesses held-out search performance; AI4AI-Bench
distinguishes run-side edits from learning-algorithm changes
\citep{heuresis2026,ai4aibench2026}. Xcientist and AutoResearchEval examine
claim--implementation alignment and research trajectories
\citep{xcientist2026,autoresearcheval2026}. These complement the AI4AI Survey's
discussion of reliable improvement \citep{ai4aisurvey2026}.
Appendix~\ref{app:research-boundaries} provides an extended comparison.
\vspace{-0.5em}

\vspace{-0.5em}
\section{Experimental Framework for Agentic Data Engineering}
\label{sec:system}
\begin{figure}[!t]
  \centering
  \vspace{-2em}
\includegraphics[width=0.99\linewidth]{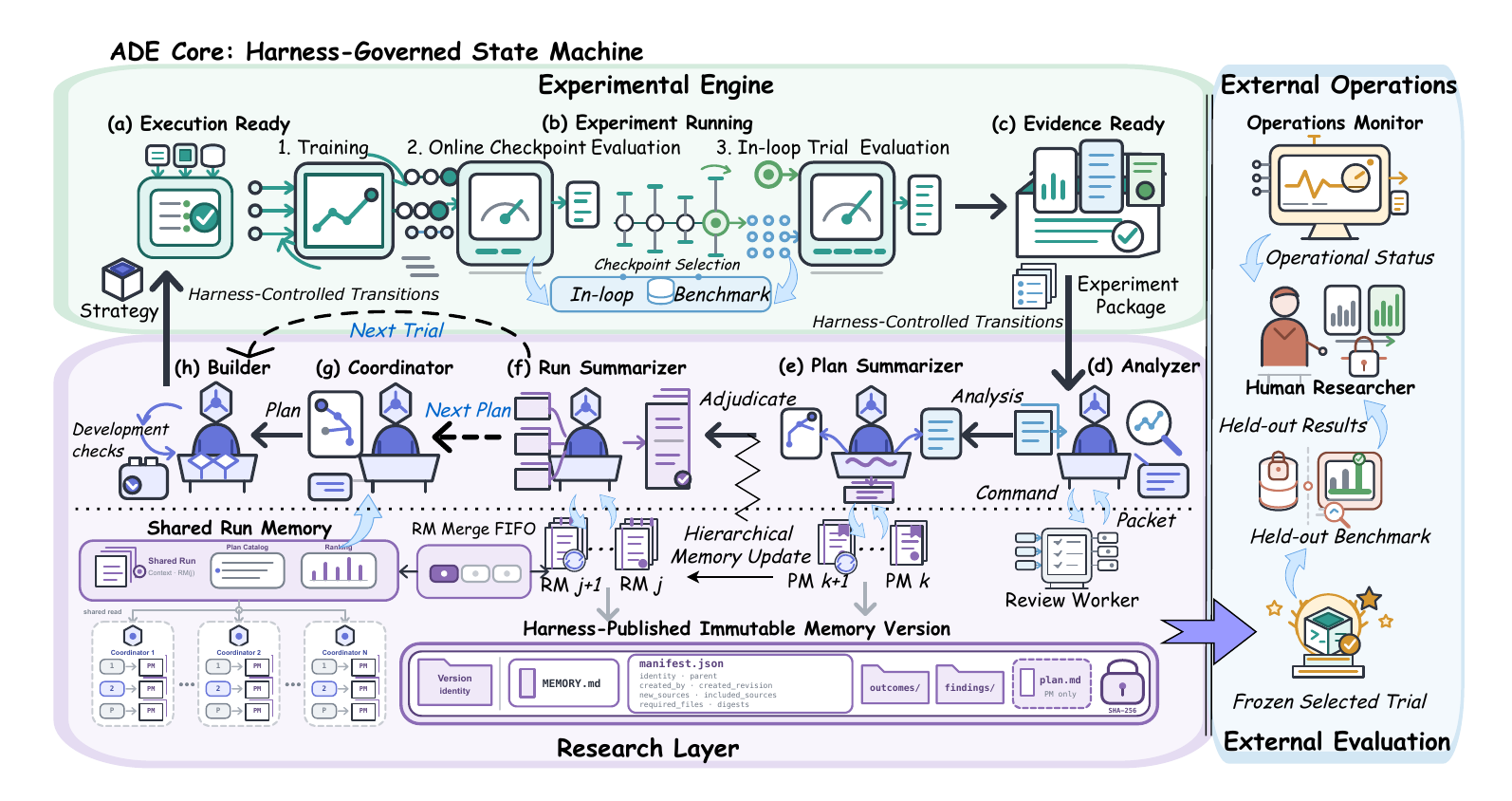}
  \vspace{-1.5em}
  \caption{\name{} under Harness control: Agents develop strategies using In-loop feedback and Memory; the Engine executes controlled experiments. Oversight and held-out evaluation remain external.}
  \label{fig:ade-method-overview}
  \vspace{-1em}
\end{figure}

\vspace{-0.5em}
\textbf{Overview.}
\name{} develops strategies through a \emph{Harness}-governed state machine
(Figure~\ref{fig:ade-method-overview}) with task-defined strategy interfaces and
experimental contracts. Under \textcolor{adeteal}{\textbf{Principle 1}}, \emph{Agents} use in-loop feedback from
\emph{Experiment Packages} and cumulative \emph{Memory} to direct research; the
\emph{Experimental Engine} executes admitted strategies under fixed experimental protocols; the
Harness validates actions, controls information access, and commits state
transitions. Under \textcolor{adeblue}{\textbf{Principle 2}}, selection follows in-loop scores;
\emph{External Evaluation} assesses the selected checkpoint on held-out data.

A \emph{Run} is one bounded research process containing \emph{Plans}, each
specifying a research direction and its commitment. A \emph{Trial} is one
realization and experimental evaluation of a Plan's fixed commitment.
We next describe the experimental contract
(Section~\ref{sec:controlled-execution}), agent research and Memory
(Section~\ref{sec:strategy-proposal}), Harness enforcement
(Section~\ref{sec:harness-protocol}), and external evaluation and oversight
(Section~\ref{sec:heldout-oversight}). Appendix~\ref{app:formal-protocol}
formalizes the state transitions.

\vspace{-0.5em}
\subsection{Experimental Contract and Controlled Execution}
\label{sec:controlled-execution}

\textbf{The experimental contract defines the intervention boundary.}
Each task specifies a strategy interface with authorized inputs, required
outputs, and constraints. Agents express each strategy as an executable
implementation of this interface. Across Trials of the same task, the
experimental configuration $\mathcal B$ fixes the target model, candidate data and evaluation splits,
training budget, optimization recipe, checkpoint rule, evaluator, and resource
policy. The admitted strategy is the only agent-controlled intervention; the
Engine executes it under $\mathcal B$
(Figure~\ref{fig:ade-method-overview}a). Task-specific interfaces thus connect
different interventions to a common research workflow.

\textbf{Checkpoint selection and Trial comparison use separate evaluation budgets.}
For Trial $k$ in Plan $p$, the Engine trains the target model and performs two
stages of evaluation on the same in-loop dataset $\mathcal D^{\mathrm{in}}$
(Figure~\ref{fig:ade-method-overview}b). Online evaluation selects checkpoint
$\theta_{p,k}^{\star}$ from training trajectory $\Theta_{p,k}$; offline
evaluation scores this checkpoint to produce the Trial's in-loop
performance record $s_{p,k}$:
{\small
\begin{equation}
  \theta_{p,k}^{\star}
  =\arg\max_{\theta\in\Theta_{p,k}}
    S_{\mathrm{online}}
    (\theta;\mathcal D^{\mathrm{in}},K_{\mathrm{online}}),
  \qquad
  s_{p,k}
  =S_{\mathrm{in}}
    (\theta_{p,k}^{\star};\mathcal D^{\mathrm{in}},K_{\mathrm{offline}}),
  \quad K_{\mathrm{online}}\leq K_{\mathrm{offline}}.
  \label{eq:engine-checkpoint-and-in-loop-evaluation}
\end{equation}}
Here $K_{\mathrm{online}}$ and $K_{\mathrm{offline}}$ are the numbers of sampled
responses per example, fixed across Trials. Using fewer samples for repeated
online checks reduces selection cost; a larger offline budget can improve the
reliability of Trial comparisons. Both evaluations remain in-loop; offline evaluation is distinct from held-out evaluation
(Section~\ref{sec:heldout-oversight}).

\textbf{Structured experimental feedback supports subsequent research.}
The Engine returns an immutable Experiment Package containing the in-loop
score, selected checkpoint, executed strategy, and
task-permitted execution, training, and model-behavior records
(Figure~\ref{fig:ade-method-overview}c). Agents inspect these records alongside
in-loop performance to analyze strategy execution and target-model responses.
Each record identifies its source Trial. The package supplies feedback for
analysis and subsequent Memory updates; the Harness enforces its information
boundaries (Section~\ref{sec:harness-protocol}).
Appendix~\ref{app:artifact-and-package} specifies the task interfaces and
package schemas.


\vspace{-0.5em}
\subsection{Agent-Directed Research with Cumulative Memory}
\label{sec:strategy-proposal}

\textbf{Agents choose research directions using prior findings.}
The \emph{Coordinator} (Figure~\ref{fig:ade-method-overview}g) draws on
\emph{Run Memory} (RM), a \emph{Plan Catalog} of existing directions, and the
current Trial ranking. It chooses between exploring a new direction and
revisiting, challenging, or combining earlier Plans; rankings inform but do not
dictate this choice. Each proposed Plan cites its motivating findings and
specifies a hypothesis, mechanism,
expected observations, rejection conditions, and controls. The accepted
commitment defines what subsequent Trials within the Plan test.

\textbf{Builders implement strategies under a fixed research commitment.}
The \emph{Builder} (Figure~\ref{fig:ade-method-overview}h) implements the
strategy from a fixed Plan snapshot with its commitment and available
\emph{Plan Memory} (PM), plus approved prior strategies records.
Bounded Harness feedback guides implementation revisions within the same Trial. These development checks do not train the target model;
drafts and probes enter neither Trial rankings nor Memory. Later Trials refine the strategy using updated PM under that commitment;
changing the hypothesis or research direction requires a new Plan.

\textbf{Independent analysis precedes hypothesis assessment.}
The \emph{Analyzer} (Figure~\ref{fig:ade-method-overview}d) receives the
Experiment Package without the Plan, hypothesis, or Memory. It can request
bounded record-level review from a \emph{Review Worker}, then synthesizes
observations and experimental evidence independently of the expected answer.
The \emph{Plan Summarizer}
(Figure~\ref{fig:ade-method-overview}e) uses these findings and the Plan's
commitment to assess how the observations relate to the proposed hypothesis.
Each Trial keeps this model-generated assessment separate from Harness-provided conclusions on
implementation fidelity and improvement over the frozen in-loop comparison
target.
Appendix~\ref{app:agent-protocols} specifies the role contracts and outcome
categories.

\textbf{Cumulative Memory connects findings to subsequent decisions.}
After each terminal Trial, the Plan Summarizer updates Plan Memory using its
prior contents, the Plan, and the Trial's results and analysis. Plan Memory preserves the direction's findings,
contradictions, and unresolved questions. The \emph{Run Summarizer}
(Figure~\ref{fig:ade-method-overview}f) integrates this update with existing
Run Memory and current Run-level results for subsequent Coordinator decisions.
Findings retain their source Trials, scope,
and assessment status, making their use in later research decisions traceable;
the Harness governs when updates become available to other Plans
(Section~\ref{sec:harness-protocol}).

\vspace{-0.5em}
\subsection{Harness Enforcement of the Research Protocol}
\label{sec:harness-protocol}

\textbf{Agents propose state changes; the Harness commits them.}
Before advancing the state machine, the Harness validates agent-submitted
research decisions, executable strategies, analyses, and Memory updates against
the calling role, current state, input versions, and required outputs
(Figure~\ref{fig:ade-method-overview}, Harness-Controlled Transitions).
Before execution, it checks the strategy's interface, authorized inputs,
constraints, and executable behavior, then binds the accepted strategy to
fixed configuration $\mathcal B$ in a frozen Engine command. Admission establishes
protocol compliance and executability. For system state $\mathcal S_t$ and protocol
$\mathcal P$:
\begin{equation}
  \underbrace{
    u_t\sim\pi_{\mathrm{agent}}
    \bigl(\,\cdot\mid\operatorname{View}_{\mathcal P}(\mathcal S_t)\bigr)
  }_{\text{agent-proposed research action}}
  \quad \mathrel{\text{vs.}} \quad
  \underbrace{
    u_t\in\mathcal A_{\mathcal P}(\mathcal S_t),
    \qquad
    (\mathcal S_{t+1},o_t)
    =\mathsf H_{\mathcal P}(\mathcal S_t,u_t)
  }_{\text{Harness-enforced state transition}}.
  \label{eq:research-agency-vs-experimental-control}
\end{equation}
Here $\operatorname{View}_{\mathcal P}(\mathcal S_t)$ is the role's supplied
information, $\mathcal A_{\mathcal P}(\mathcal S_t)$ the admissible actions,
and $\mathsf H_{\mathcal P}$ the transition function that commits the next
state and returns observations $o_t$.

\textbf{Role-specific inputs define the information boundary.}
The Harness builds role-specific input packages from a fixed research-state
version: Run-level findings for Coordinators, Plan snapshots and approved
source strategies for Builders, and Experiment Packages for Analyzers
(without the Plan, hypothesis, or Memory). The Engine receives the executable
strategy, fixed configuration, and data references. The Harness validates
review requests against permitted records and task requirements, dispatches
the Review Worker, and returns its findings to the Analyzer. Agents are
instructed to use only supplied inputs. Appendix~\ref{app:agent-protocols}
specifies role contracts and information-access rules.

\textbf{Versioned commits preserve the basis of research decisions.}
Plans execute asynchronously, while each planning call retains a fixed snapshot
of shared findings. The Harness commits Plan Memory before queuing its update
to Run Memory, processes Run Memory updates one at a time, and enables sharing
and strategy reuse only after both updates are committed. Later calls can use
new findings; running calls retain their original inputs. Recovery preserves
the frozen inputs and committed state of the interrupted action.
Appendices~\ref{app:update-queue} and~\ref{app:recovery-semantics} specify the
commit and recovery rules.

\subsection{Held-Out Evaluation and Operational Oversight}
\label{sec:heldout-oversight}

\textbf{Final selection uses only in-loop scores.}
Following \textcolor{adeblue}{Principle 2}, \name{} ranks Plans by their best
Trials under the task's ordered in-loop metrics. It selects the highest-ranked
Plan and freezes its best Trial's executable strategy, training result, and
selected checkpoint. External Evaluation assesses that same checkpoint on
$\mathcal D^{\mathrm{hold}}$, disjoint from the in-loop dataset
$\mathcal D^{\mathrm{in}}$ (Figure~\ref{fig:ade-method-overview}, External
Evaluation). Let $\tau_p^{\star}$ denote the best eligible Trial in Plan $p$:
\begin{equation}
  \widehat p
  =\arg\max_p S_{\mathrm{in}}(\tau_p^{\star}),
  \qquad
  e^{\mathrm{hold}}
  =\operatorname{Eval}_{\mathrm{hold}}(\tau_{\widehat p}^{\star}).
  \label{eq:in-loop-selection-held-out-evaluation}
\end{equation}
Here $S_{\mathrm{in}}$ uses the offline in-loop scores from
Section~\ref{sec:controlled-execution}. 

\textbf{Held-out results remain outside subsequent research decisions.}
Held-out evaluation tests whether strategy improvements extend to data
withheld from research. Held-out
examples, responses, and scores are stored in an external evaluation record
and excluded from checkpoint selection, Plan and Trial rankings, Memory, and subsequent
research decisions. If they guide a later revision, they become in-loop
feedback and that revision requires fresh held-out evaluation. Held-out results
assess effectiveness within the configured evaluation scope.

\textbf{Operational oversight remains outside research decisions.}
The \emph{Operations Monitor} reports resource availability and run status to
the \emph{Human Researcher} (Figure~\ref{fig:ade-method-overview}).
After recoverable failures, such as worker crashes or stalled training,
recovery replays the frozen experiment, preserving its Plan, strategy,
configuration, and recorded results. Recovery actions are logged separately
from research Memory. The Human Researcher uses domain expertise and held-out
results to review \name{}'s final strategy and decide whether to adopt it.
Appendix~\ref{app:recovery-semantics} specifies operational boundaries.

Together, these mechanisms give agents control over research decisions under
fixed experimental protocols (\textcolor{adeteal}{P1}) and separate in-loop selection
from held-out evaluation (\textcolor{adeblue}{P2}).

\section{Evaluating Strategy Effectiveness and Research Capability}
\label{sec:empirical}

\label{sec:scope-rqs}
We evaluate \name{} through four research questions. RQ1 examines strategy
effectiveness, while RQ2--RQ4 examine research progress,
organization, and feedback use during strategy development.
\begin{itemize}[leftmargin=*,label=\textbullet,itemsep=0.1em,topsep=0.2em]
  \item \textbf{RQ1: Strategy effectiveness.} Do \name{}-developed strategies improve target-model performance over generic baselines on both in-loop and held-out benchmarks?
  \item \textbf{RQ2: Progressive strategy improvement.} Does \name{} discover stronger strategies over successive experiments, and how does research capacity affect progress and computational cost?
  \item \textbf{RQ3: Research organization and experimental feedback.} How does strategy quality compare across research organizations, feedback settings, and independent search?
  \item \textbf{RQ4: Feedback-guided research trajectories.} How do findings from earlier experiments inform subsequent research decisions and strategy revisions?
\end{itemize}

\begin{table}[!t]
  \centering
  \vspace{-1em}
  \small
  \setlength{\tabcolsep}{4pt}
  \caption{Controlled settings for the four data-engineering tasks.}
  \label{tab:experimental-settings}
  \resizebox{0.95\linewidth}{!}{%
  \begin{tabular}{@{}lcccc@{}}
    \toprule
    \textbf{Task} & \textbf{Domain} & \textbf{Training paradigm} & \textbf{Agent-controlled strategy} & \textbf{Generic baseline} \\
    \midrule
    Data Selection & Math & Long-CoT SFT & Subset selection & Random subset \\
    Data Selection & Code & Long-CoT SFT & Subset selection & Random subset \\
    Reward Design & Math & RFT & Reward + group credit & Outcome-only reward \\
    Curriculum Learning & Math & RFT & Training-data schedule & Random ordering \\
    \bottomrule
  \end{tabular}%
  }
\vspace{-0.5em}
\end{table}

\begin{table}[!t]
  \centering
  \vspace{-1em}
  \caption{Main results for in-loop-selected strategies (\%). Cells report Pass@$K$ / Average@$K$.}
\label{tab:effectiveness-results}
\scriptsize\setlength{\tabcolsep}{2.5pt}
\renewcommand{\arraystretch}{0.85}
\resizebox{0.95\linewidth}{!}{%
\begin{tabular}{@{}lc|ccccc@{}}
\toprule
\textbf{Method} & \textbf{In-loop} & \multicolumn{5}{c}{\textbf{Held-out Benchmark}} \\
\midrule
\textbf{Data Selection (Math)} & \multicolumn{1}{c|}{AIME24} & \multicolumn{1}{c}{MATH-500} & \multicolumn{1}{c}{MATH-S} & \multicolumn{1}{c}{AIME25} & \multicolumn{1}{c}{HMMT} & \multicolumn{1}{c}{Macro Avg.} \\
\addlinespace[1pt]
Base model & 36.67 / 11.15 & 85.80 / 76.00 & 86.00 / 78.00 & 30.00 / 7.71 & 16.67 / 2.29 & 54.62 / 41.00 \\
Generic Baseline & 43.33 / 12.60 & 87.60 / 78.00 & 88.00 / 77.87 & 33.33 / 11.15 & 30.00 / \textbf{7.40} & 59.73 / 43.61 \\
ADE ($C=1$) & \textbf{56.67} / \textbf{17.81} & \textbf{90.20} / \textbf{80.00} & \textbf{88.80} / \textbf{79.73} & \textbf{46.67} / 14.90 & \textbf{36.67} / 6.67 & \textbf{65.58} / \textbf{45.32} \\
ADE ($C=3$) & \textbf{56.67} / 16.98 & 87.20 / 77.33 & 87.60 / 78.73 & 43.33 / \textbf{16.25} & \textbf{36.67} / 7.08 & 63.70 / 44.85 \\
\midrule
\textbf{Data Selection (Code)} & \multicolumn{1}{c|}{LCB $\leq$03/30} & \multicolumn{1}{c}{LCB $>$03/30} & \multicolumn{1}{c}{HumanEval+} & \multicolumn{1}{c}{Codeforces} & \multicolumn{1}{c}{CodeElo} & \multicolumn{1}{c}{Macro Avg.} \\
\addlinespace[1pt]
Base model & 40.55 / 27.49 & 24.26 / 15.95 & 85.98 / 74.39 & 7.28 / 3.93 & 8.58 / 4.25 & 31.53 / 24.63 \\
Generic Baseline & 39.86 / 26.35 & 25.85 / 18.44 & 85.98 / 70.53 & \textbf{8.57} / 4.85 & 9.07 / 5.23 & 32.37 / 24.76 \\
ADE ($C=1$) & 45.79 / 31.74 & 30.39 / 20.11 & 87.20 / 72.76 & 7.71 / 4.00 & 8.82 / 4.66 & 33.53 / 25.38 \\
ADE ($C=3$) & \textbf{48.06} / \textbf{33.03} & \textbf{32.20} / \textbf{22.07} & \textbf{89.02} / \textbf{76.02} & \textbf{8.57} / \textbf{5.28} & \textbf{10.54} / \textbf{5.39} & \textbf{35.08} / \textbf{27.19} \\
\midrule
\textbf{Reward Design (Math)} & \multicolumn{1}{c|}{MATH-S} & \multicolumn{1}{c}{MATH-500} & \multicolumn{1}{c}{AIME24} & \multicolumn{1}{c}{AIME25} & \multicolumn{1}{c}{HMMT} & \multicolumn{1}{c}{Macro Avg.} \\
\addlinespace[1pt]
Base model & 39.40 / 20.87 & 36.60 / 19.93 & 0.00 / 0.00 & 3.33 / 0.10 & 0.00 / 0.00 & 9.98 / 5.01 \\
Generic Baseline & 49.00 / 34.13 & \textbf{46.80} / 30.13 & 3.33 / 0.21 & 0.00 / 0.00 & 0.00 / 0.00 & 12.53 / 7.59 \\
ADE ($C=1$) & 50.20 / 36.33 & 46.40 / \textbf{30.80} & 3.33 / 0.10 & \textbf{6.67} / \textbf{0.21} & \textbf{3.33} / \textbf{0.10} & 14.93 / \textbf{7.80} \\
ADE ($C=3$) & \textbf{52.40} / \textbf{36.67} & 42.00 / 27.93 & \textbf{13.33} / \textbf{0.52} & \textbf{6.67} / \textbf{0.21} & 0.00 / 0.00 & \textbf{15.50} / 7.17 \\
\midrule
\textbf{Curriculum Learning (Math)} & \multicolumn{1}{c|}{MATH-S} & \multicolumn{1}{c}{MATH-500} & \multicolumn{1}{c}{AIME24} & \multicolumn{1}{c}{AIME25} & \multicolumn{1}{c}{HMMT} & \multicolumn{1}{c}{Macro Avg.} \\
\addlinespace[1pt]
Base model & 39.40 / 20.87 & 36.60 / 19.93 & 0.00 / 0.00 & \textbf{3.33} / \textbf{0.10} & 0.00 / 0.00 & 9.98 / 5.01 \\
Generic Baseline & 49.00 / 34.13 & 46.80 / 30.13 & 3.33 / 0.21 & 0.00 / 0.00 & 0.00 / 0.00 & 12.53 / 7.59 \\
ADE ($C=1$) & 51.80 / 36.40 & 45.80 / 30.60 & \textbf{6.67} / 0.31 & 0.00 / 0.00 & \textbf{3.33} / \textbf{0.10} & 13.95 / 7.75 \\
ADE ($C=3$) & \textbf{54.40} / \textbf{37.13} & \textbf{47.60} / \textbf{31.07} & \textbf{6.67} / \textbf{0.42} & 0.00 / 0.00 & \textbf{3.33} / \textbf{0.10} & \textbf{14.40} / \textbf{7.90} \\
\bottomrule
\end{tabular}}
\vspace{-1.5em}
\end{table}

\textbf{Experimental setup.}
\label{sec:experimental-design}
We evaluate \name{} on four data-engineering tasks
(Table~\ref{tab:experimental-settings}). Both long chain-of-thought SFT
(Long-CoT SFT) tasks use
Qwen2.5-7B-Instruct and select 384 trajectories from a shared fixed mixed-domain
pool of 3,840 OpenThoughts trajectories \citep{guha2025openthoughts}, with task-specific
evaluation objectives. Both RFT tasks use Qwen2.5-0.5B, 3,072 MATH problems
\citep{hendrycks2021math}, and 100 policy optimization steps; curriculum learning
varies a precomputed data schedule with fixed outcome-only reward.
Base model denotes the target before training. We compare $C=1$ versus $C=3$
concurrent Coordinator lanes, budgeting five Plans per Coordinator and one Trial
per Plan, with three Runs per full ADE setting. Research agents use GPT-5.6 Sol
in Codex with extra high reasoning effort; Qwen3.6-35B-A3B provides record-level
review. Appendix~\ref{app:experimental-configuration} details training,
agent budgets, and comparison protocols.

\textbf{Evaluation.}
In-loop scores select checkpoints and rank Plans by their best Trials.
Held-out evaluation assesses the selected checkpoint independently of research.
We report Pass@$K$ (fraction of problems with at least one correct response
among $K$ samples, measuring coverage) / Average@$K$ (mean fraction of correct
responses per problem, measuring average correctness). SFT prioritizes Pass@$K$
and RFT Average@$K$, with the other metric breaking ties.
Reported in-loop and held-out scores use $K=32$ on AIME/HMMT and $3$ elsewhere.
The table column Macro Avg. reports Pass@$K$ / Average@$K$ averaged equally
across each task's four held-out benchmarks, using benchmark-specific $K$. Bold marks
per-metric maxima within each task. Appendix~\ref{app:benchmark-details}
details benchmark splits and evaluation protocols.
Table~\ref{tab:effectiveness-results} reports ADE's Best strategy across three
Runs, selected by in-loop scores. Across-Run means and standard deviations
appear in Appendix~\ref{app:complete-effectiveness}, Table~\ref{tab:average-results}.

\subsection{RQ1: Strategy Effectiveness}
\label{sec:rq1}

\textbf{Selected strategies improve performance across all four tasks.}
Table~\ref{tab:effectiveness-results} summarizes strategy effectiveness.
Both $C=1$ and $C=3$ improve in-loop Pass@$K$ and Average@$K$ and held-out
Macro Avg. Pass@$K$ over the generic baselines on all four tasks. These gains
show that the discovered strategies benefit target-model learning beyond the in-loop benchmarks.
Appendix~\ref{app:complete-effectiveness} reports individual Run results and their aggregate performance.

\begin{figure*}[!t]
  \centering
  \includegraphics[width=0.97\linewidth]{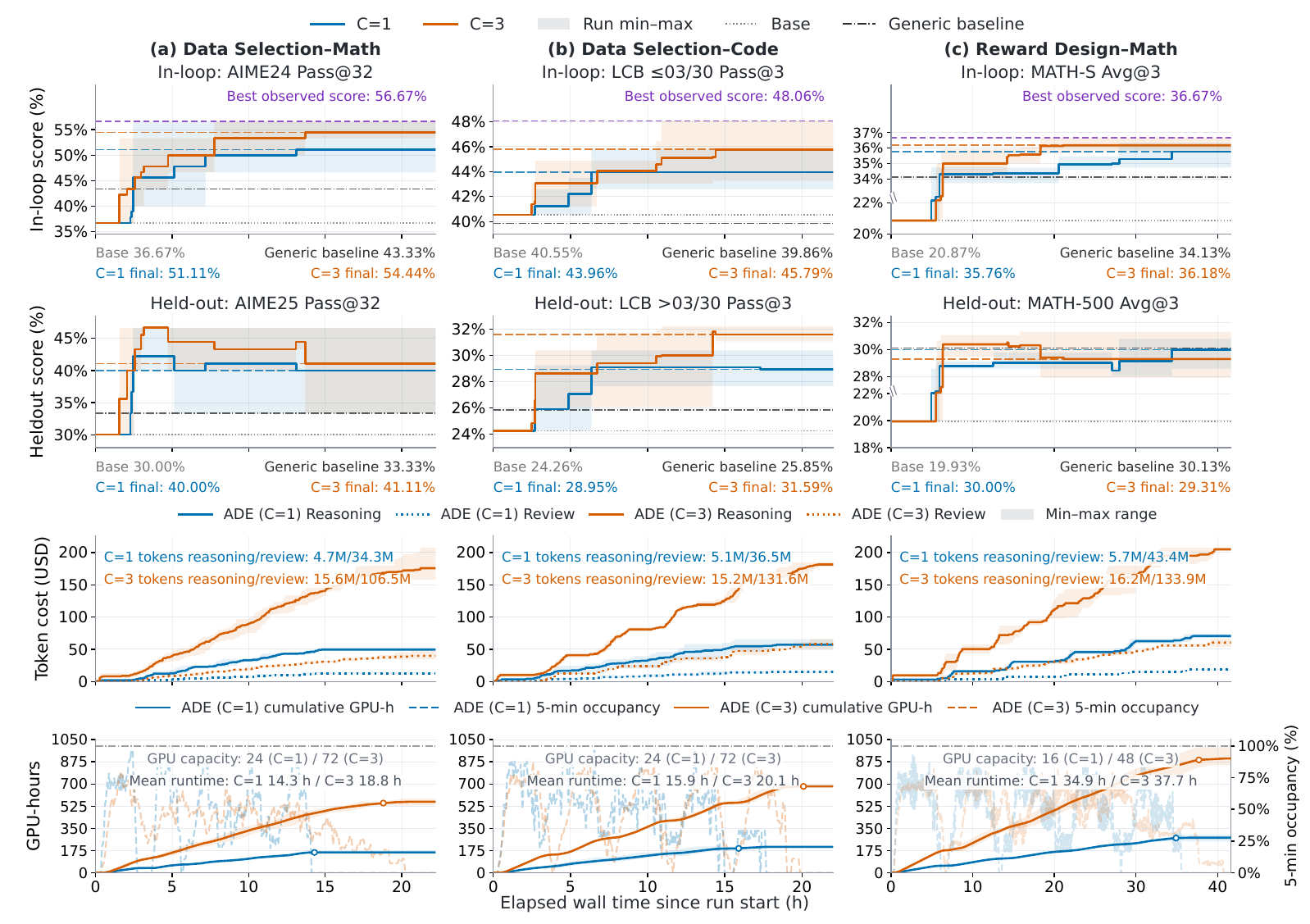}
  \vspace{-1em}
  \caption{Research progress and cost for $C=1$ (blue) and $C=3$ (orange):
  rows show best-so-far in-loop scores, held-out scores of in-loop-selected models,
  cumulative token cost, and GPU-hours with allocation occupancy.
  Lines average three Runs per task and $C$; shading shows min--max ranges.}
  \label{fig:rq2}
\end{figure*}

\begin{figure*}[!t]
  \centering
  \includegraphics[width=0.97\linewidth]{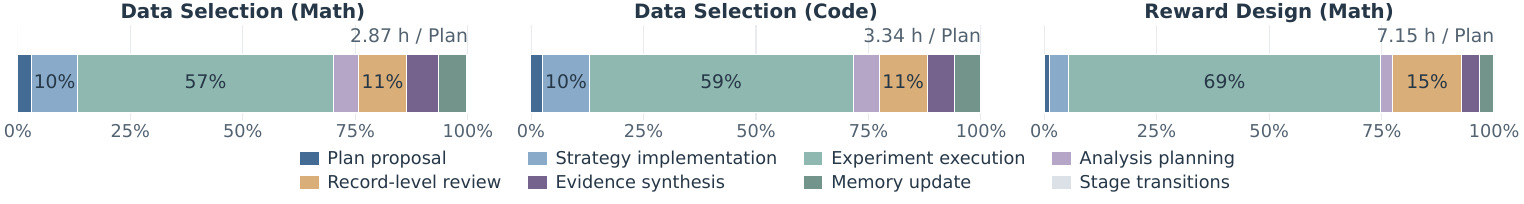}
  \vspace{-1em}
  \caption{Estimated research-stage times at $C=1$. Each bar shows a task's
  stage.}
  \label{fig:plan-stage-time}
  \vspace{-1em}
\end{figure*}

\textbf{Data selection gains extend beyond the in-loop benchmark.}
Math and code selection improve both metrics on AIME25 and the temporally
held-out LCB split, respectively. For math at $C=1$, absolute gains over the
generic baseline reach 13.34\% on AIME25 Pass@32 and 5.85\% in held-out
Macro Avg. Pass@$K$ (Table~\ref{tab:effectiveness-results}).

\textbf{Reward design primarily improves problem coverage.}
At $C=3$, AIME24 Pass@32 rises from 3.33\% to 13.33\%,
while Average@32 rises from 0.21\% to 0.52\%.
The held-out Macro Avg. shows a coverage--correctness tradeoff:
Pass@$K$ increases while Average@$K$ decreases.
The strategy's main benefit is expanding the set of problems the model can solve.

\textbf{Curriculum learning improves performance under a fixed reward and target-model training budget.}
Curriculum learning improves both held-out Macro Avg. metrics at both
capacities while retaining the same model, training-data pool, outcome-only
reward, and 100-step budget.
These results support training-data allocation as a useful intervention
under the same training conditions as the generic baseline.

\subsection{RQ2: Progressive Strategy Improvement}
\label{sec:rq2}

\textbf{Later experiments develop strategies with higher in-loop scores.}
Figure~\ref{fig:rq2} plots the best in-loop score $B_r(t)$ over successful
Trials completed by time $t$, initialized at Base and averaged equally over
three Runs. For math data selection at $C=3$, mean AIME24 Pass@32 rises from
45.56\% after the initial strategies complete (about 2.55 hours) to 54.44\% (about 13.70 hours).

\textbf{Held-out trajectories reveal how search progress generalizes.}
Figure~\ref{fig:rq2} (Row 2) tracks the held-out performance of models selected
during search. For math data selection at $C=3$, mean AIME25 Pass@32
peaks at 46.67\% and finishes at 41.11\%, above the generic baseline's 33.33\%.
During this later phase, mean in-loop AIME24 Pass@32 rises from 47.78\% to 54.44\%.
This divergence is consistent with overfitting to the search signal:
further improvements on that signal can favor different strategies from
those that generalize best. Tracking held-out trajectories exposes this
behavior alongside the overall gains from research.

\textbf{Greater capacity improves in-loop scores, with task-dependent held-out returns.}
Increasing capacity from $C=1$ to $C=3$ yields absolute gains of 0.42--3.33\% in final mean
in-loop scores across the three tasks, at triple the peak GPU allocation and
higher token use (Figure~\ref{fig:rq2}). Mean held-out LCB $>$03/30 Pass@3 rises
from 28.95\% to 31.59\%, and AIME25 Pass@32 from 40.00\% to 41.11\%, but
MATH-500 Average@3 for reward design falls from 30.00\% to 29.31\%.

\textbf{Experiment turnaround shapes research throughput.}
Experiment execution, including waiting, occupies 57--69\% of Plan time at
$C=1$, more than Agent activities (Figure~\ref{fig:plan-stage-time});
Appendix~\ref{app:stage-timing-c3} gives the $C=3$ breakdown.
These time shares identify experiment scheduling as a promising target for
improving research efficiency. In asynchronous research, scheduling can balance
available compute with the value of waiting for informative results before
choosing the next direction.

\vspace{-0.5em}
\subsection{RQ3: Research Organization and Experimental Feedback}
\label{sec:rq3}

\textbf{Comparisons vary role organization, cumulative Memory, and feedback access.}
We compare full ADE ($C=1$) with four alternatives on Data Selection--Math;
all retain the strategy interface (a) and Engine (b) in
Figure~\ref{fig:ade-method-overview}.
(1) \textit{Unified Agent} merges Coordinator, Builder, and Analyzer functions
(d--f) into one persistent session, without structured research Memory.
(2) \textit{Independent Search} removes cumulative research Memory,
so each strategy is developed independently of prior experimental findings.
(3) \textit{Outcome-only Feedback} retains Coordinator and Builder (d,e),
omits Analyzer (f), Review Worker, and Summarizers (g,h), and exposes only
candidate history, scalar scores, and execution facts.
(4) \textit{Raw Evidence Feedback} retains the same roles but also exposes
raw evidence from the Experiment Package (c) directly to the Coordinator (d).
Appendix~\ref{app:comparison-protocol} details the protocols.

\begin{figure}[!t]
\centering
\vspace{-1em}
\includegraphics[width=0.95\linewidth]{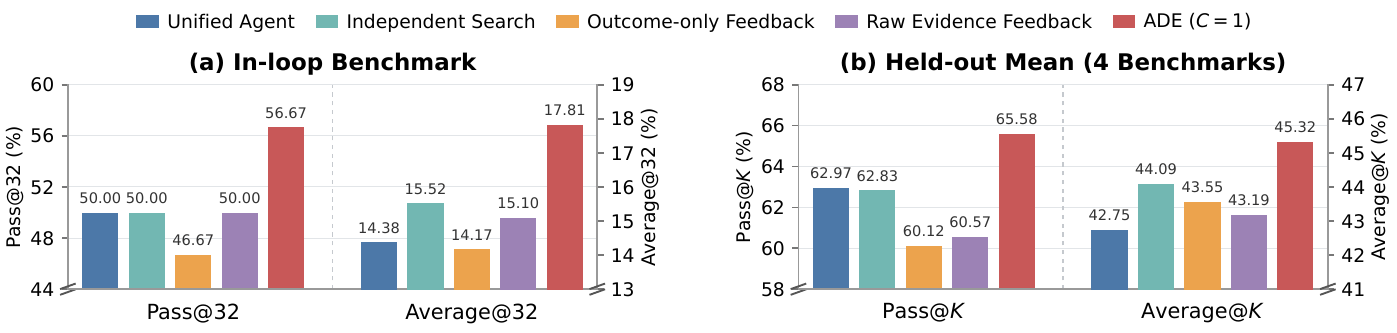}
\vspace{-1em}
\caption{Data Selection--Math research organization and feedback comparisons. Each panel shows Pass@$K$ (left axis) and
Average@$K$ (right axis); held-out scores average four benchmarks.}
\label{fig:math-sft-ablation}
\vspace{-1em}
\end{figure}

\begin{table}[!t]
  \centering
  \fontsize{8.5}{9}\selectfont
  \caption{Selected strategies across four tasks; linked appendices provide detailed strategy scripts.}
  \label{tab:proposed-strategies}
  \resizebox{0.95\linewidth}{!}{%
  \begin{tabular}{@{}L{0.15\textwidth}L{0.78\textwidth}@{}}
    \toprule
    Task & Agent-developed strategy \\
    \midrule
    Data Selection (Math)
      & \textbf{Select valid trajectories with direct, low-redundancy reasoning.}
        Filter by training cutoff and Judge assessments, then deduplicate and
        select under fixed size, length-quota, and token-budget constraints.
        $\rightarrow$ (\hyperref[app:artifact-math-sft]{Appx.~\ref*{app:artifact-math-sft}}) \\
    Data Selection (Code)
      & \textbf{Stratified Repetition Control.}
        Replace highly repetitive records with low-repetition candidates matched
        by prompt subtype, cutoff status, length band, and nearest token count.
        $\rightarrow$ (\hyperref[app:artifact-code-sft]{Appx.~\ref*{app:artifact-code-sft}}) \\
    Reward Design (Math)
      & \textbf{Group-conditioned process credit.}
        Use sibling correctness and corroboration to weight reasoning quality;
        require corroboration in all-wrong groups.
        $\rightarrow$ (\hyperref[app:artifact-math-rft]{Appx.~\ref*{app:artifact-math-rft}}) \\
    Curriculum Learning (Math)
      & \textbf{Difficulty cycling with paired reinforcement.}
        Prioritize unseen or less-used problems; after full-pool coverage,
        pair foundation and bridge batches using same-subject problems.
        Freeze the schedule before training.
        $\rightarrow$ (\hyperref[app:artifact-curriculum]{Appx.~\ref*{app:artifact-curriculum}}) \\
    \bottomrule
  \end{tabular}%
  }
\vspace{-2em}
\end{table}

\textbf{Strategy quality varies with research organization and feedback access.}
Raw Evidence Feedback improves both in-loop metrics over Outcome-only Feedback
and trades higher held-out Macro Avg. Pass@$K$ for lower Average@$K$
(Figure~\ref{fig:math-sft-ablation}). ADE's selected strategy exceeds all four
alternatives on both in-loop metrics and held-out Macro Avg., reaching
65.58\% Pass@$K$ and 45.32\% Average@$K$.
Relative to the strongest alternative on each metric, absolute gains are
6.67\% in in-loop Pass@32 and 2.62\% in Macro Avg. Pass@$K$.

These comparisons favor a workflow that combines evidence interpretation,
cumulative findings, and subsequent research decisions.

\textbf{Research organization and evidence reuse improve strategy quality.}
ADE exceeds Independent Search on both metrics, in-loop and on held-out
Macro Avg. This comparison supports carrying findings across experiments;
Unified Agent examines the organization of planning, implementation, and
analysis. Section~\ref{sec:rq4} traces how these research workflows translate
into subsequent decisions.

\subsection{RQ4: Feedback-Guided Research Trajectories}
\label{sec:rq4}

\begin{figure}[!t]
\centering
\vspace{-1em}
\includegraphics[width=0.95\linewidth]{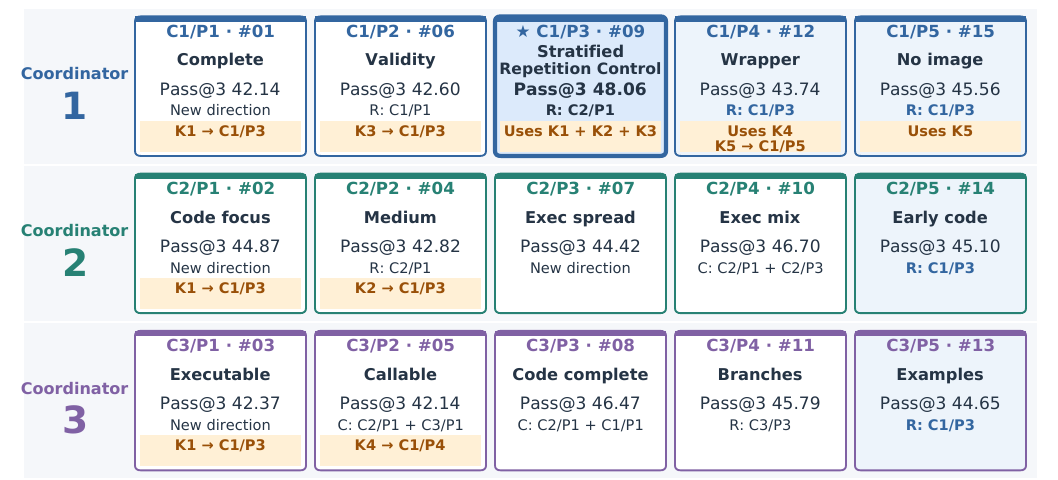}
\vspace{-1em}
\caption{Data Selection (Code) trajectories: R/C denote revisited/combined parents; K1--K5, knowledge links; \#, proposal order; star, selected Plan. Scores are in-loop Pass@3 (\%).}
\vspace{-1.5em}
\label{fig:rq4-code-sft-trace}
\end{figure}

\textbf{Later strategies combine findings across research directions.}
Table~\ref{tab:proposed-strategies} presents strategy examples across four tasks;
Figure~\ref{fig:rq4-code-sft-trace} traces one in-loop-selected Data Selection
(Code) Run. Its selected Plan, C1/P3, builds on C2/P1: redundancy findings
motivate repetition control (K1), C2/P2's unsuccessful shortening motivates
length matching (K2), and C1/P2's uninformative Judge assessments of training
records motivate a deterministic selection signal (K3). These findings jointly
shape \emph{Stratified Repetition Control}, which reaches 48.06\% in-loop
Pass@3, an absolute gain of 3.19\% over its parent, while Average@3 rises
from 32.50\% to 33.03\%.
The question evolves from whether long reasoning harms learning to whether
reducing repetition at comparable length improves learning. C1/P3 turns this
distinction into a selection rule, illustrating how experimental feedback
refines both the research question and its intervention.

\textbf{Useful findings extend beyond inherited strategies.}
C1/P3 seeds four follow-up Plans across all three Coordinators.
C3/P2's callable-defect findings inform C1/P4's function-definition and
invocation checks (K4); C1/P4's image-placeholder finding then informs C1/P5
(K5), which still inherits C1/P3. Findings can therefore guide later decisions
independently of which strategy is adopted.
Appendix~\ref{app:evidence-cases} provides the full trace and supporting evidence.

\vspace{-0.5em}
\section{Conclusion and Discussion}
\label{sec:conclusion}
\vspace{-0.5em}
We studied agent-directed data engineering across four tasks under fixed
experimental protocols, examining strategy effectiveness, research progress,
organization, and feedback use. Selected strategies improve
target-model performance over generic baselines; gains vary across
held-out benchmarks and metrics. Strategies progressively improve on benchmarks
used during research. Organization and feedback ablations favor the complete
system; trajectories link earlier findings to later revisions and combinations
beyond direct strategy inheritance. Results support assessing AI4AI research
capability through progressive improvement and feedback-guided decisions,
alongside held-out effectiveness.

\textbf{Limitations and future work.}
Adaptive search raises a general challenge for AI4AI: repeated optimization
of an in-loop signal can overfit the feedback available to the research agent.
Our held-out trajectories expose this risk alongside the gains from search,
highlighting the value of evaluating generalization throughout research.
LLM Agent stochasticity poses a second challenge: the same inputs can lead
agents to different strategies and experimental paths. Comparing different
research agents could clarify how their reasoning and exploration choices
shape research outcomes. Testing discovered strategies across model families
and downstream tasks would assess their transfer. 
Appendix~\ref{sec:reflections} discusses future directions for AI4AI research.

\begingroup
\def\UrlBreaks{\do\/}
\def\UrlBigBreaks{}
\bibliography{iclr2027_conference}
\bibliographystyle{iclr2027_conference}
\endgroup

\appendix


\clearpage

\section{Extended Related Work}
\label{app:research-boundaries}

This appendix expands the related-work discussion along the three dimensions
in Table~\ref{tab:study-boundaries}: intervention scope, research capability
(Q1), and effectiveness (Q2). Tables~\ref{tab:research-boundaries-systems}
and~\ref{tab:research-boundaries-evaluation} compare the intervention and
execution controls, evidence reuse, and selection and evaluation procedures
reported for representative systems and benchmarks. We then relate these
choices to work on recursive self-improvement (RSI), distinguishing the
component being improved from the process that produces the improvement.

\textbf{Reading the comparison.}
The tables summarize the cited configurations. Source sections refer to the
cited papers, and repository headings refer to the cited software documentation.
An evaluation can be external to training yet still guide adaptive research.
Here, \emph{held-out from research} means that evaluation examples are
separate from those used for in-loop evaluation and that their results do not
guide research or selection.
Where a cited protocol does not specify a separate held-out evaluation, the
table records that reporting limit without inferring that the system cannot
support one.

\begin{table}[!t]
  \centering
  \caption{Research boundaries in representative autonomous research systems.
  Source locations identify the basis for each row.}
  \label{tab:research-boundaries-systems}
  \fontsize{8.5}{10}\selectfont
  \setlength{\tabcolsep}{4pt}
  \renewcommand{\arraystretch}{1.12}
  \resizebox{0.95\linewidth}{!}{%
  \begin{tabularx}{0.95\linewidth}{@{}L{0.14\linewidth}YYY@{}}
    \toprule
    \textbf{Work / source} & \textbf{Intervention and execution}
      & \textbf{Research and evidence reuse}
      & \textbf{Selection and final evaluation} \\
    \midrule
    AutoResearch\newline\citep{autoresearch2026}\newline\emph{program.md}
      & Agent edits training code, including architecture and optimization.
        Data loading, evaluator, and a five-minute training limit are fixed.
      & Agent proposes and runs changes. Git history and a results ledger
        record retained, discarded, and failed~experiments.
      & Validation bits per byte guide retention. The documented loop does
        not specify a separate untouched final audit. \\
    \addlinespace[7pt]
    POISE\newline\citep{poise2026}\newline\S2--3
      & Agent changes RL optimization components under a common training
        interface, aligned budgets, and fixed experiment~hyperparameters.
      & Search selects archive parents; lineage, reference algorithms, and
        reflective analyses inform new proposals. Failed or context-dependent
        designs remain in the archive.
      & A weighted reasoning-suite score guides evolution. The cited protocol
        does not specify an additional untouched evaluation after selection. \\
    \addlinespace[7pt]
    DataMaster\newline\citep{datamaster2026}\newline\S3; App.~B.2
      & Data discovery, composition, and processing change under a fixed
        learning algorithm. Benchmark settings fix the training code or the
        training script and base model.
      & Tree scheduling allocates experiments. Global Memory and a shared
        Data Pool carry findings and reusable artifacts across branches.
      & Downstream node scores guide search; results use best submissions.
        An additional untouched audit after selection is not specified in
        the cited~sections. \\
    \addlinespace[7pt]
    EvoTrainer\newline\citep{evotrainer2026}\newline\S3--4; App.~E.1
      & RL recipes and diagnostic tools evolve. Humans bootstrap the stack
        and approve training launches and version promotion or pruning.
      & Agent diagnoses versions and proposes interventions. Persistent
        memory, reusable skills, and negative evidence support later branches.
      & Versions are compared under shared validation conditions. A separate
        untouched final split is not specified in the cited evaluation protocol. \\
    \bottomrule
  \end{tabularx}%
  }
\end{table}

\textbf{Intervention scope and research authority.}
A fixed evaluator leaves substantial freedom over what produces a model.
AutoResearch admits broad training-code changes, whereas POISE searches RL
algorithms under shared experimental settings. DataMaster fixes the learning
algorithm and changes the data pipeline. EvoTrainer additionally develops its
diagnostic harness and includes human approval at execution boundaries.
These choices address different research questions. \name{} fixes researcher
models, the protocol, and task-specific training~conditions to study the
strategies developed within its declared data-engineering scope.

\clearpage
\begin{table}[!t]
  \centering
  \caption{Research boundaries in data-engineering studies and benchmarks,
  including this work. Freezing a submission, retraining from that submission,
  and evaluating on data withheld from in-loop evaluation are distinct
  protocol choices.}
  \label{tab:research-boundaries-evaluation}
  \fontsize{8.5}{10}\selectfont
  \setlength{\tabcolsep}{4pt}
  \renewcommand{\arraystretch}{1.12}
  \resizebox{0.95\linewidth}{!}{%
  \begin{tabularx}{0.95\linewidth}{@{}L{0.14\linewidth}YYY@{}}
    \toprule
    \textbf{Work / source} & \textbf{Intervention and execution}
      & \textbf{Research and evidence reuse}
      & \textbf{Selection and final evaluation} \\
    \midrule
    RSIBench-Data\newline\citep{rsibenchdata2026}\newline\S3, 5--6
      & Agents construct training experience and choose whitelisted
        configurations. The base model and training, serving, and evaluation
        infrastructure are fixed.
      & Scores, trajectories, verifier outputs, and attempt history inform
        revision, stopping, and checkpoint selection; recorded trajectories
        support process~analysis.
      & Agent selects a checkpoint before official scoring in fresh
        environments. Selection and official evaluation use the same task
        subset; \S6 explicitly distinguishes this from held-out tasks. \\
    \addlinespace[7pt]
    AI4AI-Bench\newline\citep{ai4aibench2026}\newline\S2.3; App.~A
      & Agents edit repository code and run proxy experiments. Submitted code
        is frozen, then executed from the prescribed starting state under a
        fixed final evaluator.
      & The exploration phase permits adaptive proxy use; source code alone
        is submitted. The benchmark accommodates different researcher systems
        rather than prescribing one memory mechanism.
      & Training tasks score up to three recent checkpoints and report the
        best. Proxy/final separation varies by task: OPD uses different
        benchmarks; BTRM's final set includes its exposed proxy subset. \\
    \addlinespace[7pt]
    DataAgent\newline\citep{exploringade2026}\newline\S3--4; App.~C--D
      & Agents synthesize and curate data with a fixed teacher, student,
        and training setup within each configuration.
      & Iterative agents inspect public scores and failure examples to revise
        datasets. Public-score history determines the selected~submission.
      & After research, the submitted dataset is used to retrain the student
        and evaluate on private held-out data. This evaluates the selected
        dataset through a fresh training run. \\
    \midrule
    \name{} (ours)\newline\S\ref{sec:system}--\ref{sec:empirical}
      & Agents change selection, curriculum, or reward strategies. The
        Harness enforces a fixed protocol and task-specific training~conditions.
      & Agents revise or combine strategies using attributed findings,
        including observations from unsuccessful experiments. Records link proposals
        to the evidence available at decision~time.
      & In-loop scores select the strategy and checkpoint. The same frozen
        checkpoint is evaluated on benchmarks or splits withheld from
        in-loop evaluation; results do not return to research or selection. \\
    \bottomrule
  \end{tabularx}%
  }
\end{table}

\textbf{Research capability and evidence reuse (Q1).}
Experiment histories, reflective archives, shared artifacts, and diagnostic
skills provide different routes for carrying evidence into later decisions.
Our study examines both improvement in in-loop strategy scores and recorded
links from earlier findings to subsequent revisions and combinations.
The traces document how prior evidence informs later decisions, complementing
the research-organization and feedback comparisons in Section~\ref{sec:rq3}.

\textbf{Evaluation independence and effectiveness (Q2).}
Final scoring can differ from search feedback in timing, sampled responses,
data, or training realization. RSIBench-Data explicitly reports task-subset
reuse despite a separate official evaluation. AI4AI-Bench's boundary depends
on the task, so a single binary label would hide material differences.
DataAgent evaluates the selected dataset through retraining; \name{}
evaluates the already selected checkpoint. Our effectiveness question concerns
performance on evaluation examples withheld from in-loop evaluation, with
results excluded from research and selection. This tests effectiveness beyond
the evaluation feedback used to develop the strategy.

\textbf{Earlier foundations for feedback-driven data and reward design.}
DataEnvGym formulates data generation as sequential decisions informed by
student errors and skill feedback \citep{dataenvgym2025}.
Curation-Bench fixes the model, training recipe, and evaluation suite while
agents inspect data, implement selection policies, and revise them from
training feedback \citep{curationbench2026}. Its analysis of local policy
tuning and method-guided exploration directly informs the study of research
behavior within a fixed training setup. Eureka uses policy-training feedback
to revise executable rewards for control tasks \citep{eureka2024}; \name{}
studies reward design alongside selection and curricula for LLM training.
These works establish precedents for feedback-driven intervention design and
controlled training. Our study combines these ingredients with agent-directed
research organization, feedback ablations, and evidence-linked trajectories
across the four data-engineering tasks.

\textbf{The scope of a research decision.}
The meaning of strategy revision depends on the permitted intervention.
\citet{whatismissing2026} study choices of training paradigm, data regime,
and stage structure. Curation-Bench examines changes among data-policy
families within a fixed training setup. \name{} also fixes the training
paradigm: its decisions concern data-engineering hypotheses, their
implementation, and the allocation of experiments. Evidence of revision at
this level addresses a complementary question to revising the training
paradigm itself. MARS already studies cross-branch transfer of reflective
lessons \citep{mars2026}; our trajectories examine such evidence reuse within
the controlled data-engineering setting.

\textbf{What changes in recursive self-improvement?}
RSI work places the improvement loop at several levels.
STOP applies a code improver to its own scaffolding, using a fixed language
model \citep{stop2024}. Darwin G\"odel Machine selects coding agents from an
archive and uses them to modify their own implementations, also with frozen
foundation models \citep{dgm2026}. These systems demonstrate that the
improvement procedure can change without updating model weights.
SEAL instead learns to generate data and update directives for its own
weight adaptation, rewarding self-edits by downstream performance
\citep{seal2025}. Autodata's reported meta-optimization evolves the data
scientist's prompts through code edits \citep{autodata2026}; Frontis-MA1
trains research operators from executable experience \citep{frontisma12026}.
A-Evolve-Training revises a rolling research policy across post-training
rounds while retaining an immutable reference stack
\citep{aevolvetraining2026}. Each provides a different route for carrying
experience into the next improvement process.

\textbf{Proposed recursion and demonstrated closure.}
The scope of the experiments is as relevant as the RSI framing.
HELIX proposes build--update--rebuild and demonstrates harness evolution
and verified training-data export; its Section~9.5 states that it does not
train an updated model \citep{helix2026}. ScienceBuddy reports successive
harness-refinement and RL cycles, with a separate fixed auxiliary model
proposing harness edits \citep{sciencebuddy2026}. This evaluates changes to
the task model and its harness while retaining the auxiliary improver.
Beyond its main cross-model benchmark, RSIBench-Data reports an exploratory
same-family experiment in Section~5.5; its adapted candidates remain below
the unadapted reference \citep{rsibenchdata2026}.
Frontis-MA1 distinguishes training the improver from a sustained loop in which
successors improve their own production process \citep{frontisma12026}.
The recent roadmap of \citet{genuinersi2026} likewise distinguishes autonomy
over improvement strategies from recursive meta-improvement.
These distinctions motivate asking what an improved system contributes to
the next research round. In \name{}, strategies and Memory change while the
researcher models and Harness remain fixed. This isolates data-engineering
research as a component of AI4AI and provides experience from which future
systems could learn to conduct better research.

\section{Discussion and Future Directions}
\label{sec:reflections}
\label{sec:limitations}

AI4AI can improve both the models produced by research and the process that
produces them. \name{} studies a concrete connection between these two
objectives: agents develop data-engineering strategies, observe their effects
on learning, and carry findings into later experiments. The results motivate
moving from better strategies within a research campaign to better research
across campaigns. Work on RSI offers several ways to make this transition,
including revising an improver's code, learning research operators, and
alternating model and harness updates
\citep{stop2024,dgm2026,frontisma12026,sciencebuddy2026}.
The following directions concern what should improve, how experience should
persist, and how to measure the value of the resulting research process.

\textbf{From improving a target model to improving the researcher.}
A stronger target model could become a data generator, evaluator, or researcher
in a later campaign. These roles require different capabilities: gains on
math or code benchmarks alone do not establish better experimental judgment.
A direct next experiment would train a researcher from \name{} trajectories,
then compare it with its predecessor on fresh data-engineering tasks under
matched budgets. The record would need to preserve the evidence available
when each decision was made, including unsuccessful interventions and their
subsequent corrections. Frontis-MA1's training of research operators provides
a precedent for learning from executable research experience
\citep{frontisma12026}; SEAL provides a complementary example of learning
which self-generated updates improve a model \citep{seal2025}.
For \name{}, the relevant outcome would be improved research yield: better
strategies, fewer wasted experiments, or more accurate predictions of where
an intervention helps. Repeating this comparison across researcher generations
would test whether better research experience produces better future
research, connecting the present setting to recursive meta-improvement
\citep{genuinersi2026}.

\textbf{Letting the research procedure evolve alongside its models.}
The usefulness of a diagnostic tool or planning policy may change as the
researcher and target models improve. Autodata illustrates improvement of the
data scientist through prompt evolution \citep{autodata2026}, while
A-Evolve-Training revises the policy governing subsequent research rounds
\citep{aevolvetraining2026}. HELIX motivates revisiting harness design after
model updates; ScienceBuddy implements alternating harness revision and RL
\citep{helix2026,sciencebuddy2026}. For \name{}, this suggests allowing
versioned changes to diagnostic tools, Memory retrieval, or coordination
between campaigns. A useful first study would compare the original and
revised research procedures with both the original and updated researcher
models. This would identify whether their benefits transfer or depend on a
particular model--procedure pairing. Each campaign could retain a fixed
Engine and evaluation contract, making successive system versions comparable
while expanding what the outer research loop can~change.

\textbf{Turning experiment histories into transferable research knowledge.}
The code case in \hyperref[sec:rq4]{RQ4} shows why a winning artifact is only
part of the research output. An unsuccessful shortening intervention helps
motivate a later strategy that controls repetition while matching length.
The useful inheritance is a conditional finding about what to test, even
when the original artifact is abandoned. Reflective memory in MARS and the
archives in POISE already preserve evidence across branches
\citep{mars2026,poise2026}. A next step for \name{} is to store findings with
the model, data distribution, training budget, and diagnostic evidence under
which they were obtained. Retrieval should distinguish a replicated result
from an unresolved explanation. Testing this knowledge on new model families
would measure whether agents can predict when an earlier finding applies,
revise it when conditions change, and avoid repeating resolved experiments.
Such records could support both inference-time Memory and training of future
researchers, with their value measured on new campaigns.

\textbf{Valuing experiments that change what the agent knows.}
The feedback comparisons in \hyperref[sec:rq3]{RQ3} and the code case motivate
an objective beyond improving the current best score. An experiment can be
valuable because it separates competing explanations. For example, matched
length exposure can help distinguish a benefit from removing repetition
from a benefit due to shorter reasoning. An Analyzer could make the competing
predictions explicit; the next intervention would test the factor on which
they disagree. This also creates a concrete target for researcher learning:
does the agent choose experiments whose outcomes improve later decisions?
Evaluation could compare such a policy with score-driven refinement under a
common budget, measuring both final performance and the accuracy of
predictions on new interventions. Curation-Bench's distinction between local
tuning and method-guided exploration provides a related view of research
progress \citep{curationbench2026}.

\textbf{Measuring whether research gains repay their cost.}
Greater research capacity improves mean in-loop endpoints in
\hyperref[sec:rq2]{RQ2}, but consumes more resources and encounters plateaus.
For AI4AI, the return depends on how often a discovered strategy will be
reused. An expensive search may be useful if its strategy improves many
subsequent training runs; a cheaper strategy may be preferable for one target.
Future studies could report performance at fixed total budgets that include
researcher inference, candidate training, validation, and researcher updates.
They could also measure the cost of reaching a predeclared performance target
across repeated campaigns. Budget-aware search in MARS offers a starting
point \citep{mars2026}. \name{} adds a scheduling question: when should an
agent launch another experiment, and when should it wait for pending evidence
that could change several branches? Comparing adaptive allocation with fixed
allocation would test whether better coordination increases research yield.

\textbf{Maintaining useful feedback across successive generations.}
As research systems learn from more campaigns, previously reserved results
can become development feedback through memory or researcher training.
The evaluation unit must therefore include the history inherited by each
system version. Studies could partition entire research tasks across
researcher training, development, and final assessment, refreshing final tasks
as generations advance. They could also test whether gains survive changes
in target model, training budget, and objective. RSIBench-Data's exploratory
same-family experiment illustrates why this matters: adapted candidates
remain below the unadapted reference despite improving within the search
\citep{rsibenchdata2026}. For a recursive extension of \name{}, each successor
should be tested both as a task model and in its intended role in the next
improvement cycle. This would establish which improvements carry forward and
which remain specific to one~campaign.

\section{Formal \name{} Research Protocol}
\label{app:formal-protocol}

This appendix specifies the research protocol summarized in
Section~\ref{sec:system}. We first define the roles and research state, then
follow a Plan through proposal, implementation, analysis, and assessment.
The remaining subsections specify Memory publication, recovery, and
information boundaries before summarizing the complete execution protocol
and its invariants. Role instructions appear in
Appendix~\ref{app:agent-instructions}; experimental parameter values appear in
Appendix~\ref{app:experimental-configuration}.
The following schematic summarizes the proposal-to-Memory cycle:
\begin{equation}
\begin{aligned}
  \psi_t &\sim \pi(\,\cdot\mid\mathcal K_t),
  &z_t &= \Gamma_{\mathcal Z}(\psi_t), \\
  (\theta_t,\xi_t,e_t,s_t)
    &= \operatorname{Execute}(z_t;\mathcal B),
  &a_t &= \operatorname{BlindAnalyze}(e_t), \\
  \delta_t &= \operatorname{Adjudicate}(\psi_t,a_t,e_t,s_t),
  &\mathcal K_{t+1} &= U(\mathcal K_t,\delta_t).
\end{aligned}
\label{eq:ade-research-transition}
\end{equation}
Here $\mathcal K_t$ denotes the available research Memory, $\psi_t$ the
accepted Plan commitment inherited by a Trial, $z_t$ its admitted artifact,
$(\theta_t,\xi_t,e_t,s_t)$ the controlled experimental outcome,
$a_t$ the blinded interpretation, and $\delta_t$ the evidence-linked adjudication
used to update research state. The schematic suppresses admission checks,
role-specific views, and intermediate versions, which are specified below.
The Coordinator proposes once per Plan; later Trials inherit that Plan's
commitment rather than resampling $\psi_t$. The experiments in this paper
use one Trial per Plan, while the protocol permits sequential Trials.

\subsection{Role and Information-Boundary Matrix}
\label{app:agent-protocols}

The agent protocol specifies what each role may observe, what decision it
must make, and what it must return before research state can advance.
Role instructions and an execution walkthrough appear in
Appendix~\ref{app:agent-instructions}.

\begin{table}[!t]
\centering\adeappendixtablestyle
\caption{Role-specific evidence and decisions. Only the Harness publishes
canonical state. No research role receives held-out evaluation records as
part of its declared inputs.}
\label{tab:role-contracts}
\resizebox{0.95\linewidth}{!}{%
\begin{tabularx}{0.95\linewidth}{@{}L{0.17\linewidth}Y Y@{}}
\toprule
Role & Available evidence & Required output and authority \\
\midrule
Coordinator & Frozen RM, remaining budget, Plan catalog, ranking and admitted
source~evidence & Research Plan, relation to prior work, and precommitted
expectations. \\
Builder & Accepted Plan, frozen Plan-local evidence, source strategies,
fixed task conditions and development feedback & Executable selection, reward, or
curriculum strategy and implementation~rationale. \\
Analyzer & Realized intervention, permitted training and in-loop behavior;
review results only at synthesis & Review design, then source-linked
analysis and findings; no access to the Plan or its expectations. \\
Plan Summarizer & Parent PM, immutable Plan, current Trial outcome,
realization, comparison records and accepted analysis & Complete successor
PM and hypothesis~adjudication. \\
Run Summarizer & Parent RM, one queue-head PM update, and same-version
Run catalog and~ranking & Complete successor RM with cross-Plan findings
and unresolved questions. \\
Review Worker & Exact records and Analyzer-designed~rubrics & Record-level
judgments; no research choice or memory publication. \\
Engine / operations & Frozen execution command / operational job state &
Experiment records / recovery record; no scientific proposal authority. \\
\bottomrule
\end{tabularx}%
}
\end{table}

Before accepting a delivery, the Harness checks the role, experiment identity,
input versions, source references, required outputs and applicable task
constraints. The agent supplies scientific reasoning; the Harness establishes
that the delivery is correctly bound and admissible.

\subsection{Research State and Admissible Change}
\label{app:state-representation}

Let the scientific-state projection of a Run be
\begin{equation}
  \mathcal S_t =
  \bigl(\mathcal B,\mathcal Z,\mathcal K^R_t,
  \{\mathcal K^P_{p,t}\}_{p\in\mathcal P_t},\mathcal L_t\bigr),
  \label{eq:run-state}
\end{equation}
Here $t$ indexes research-state snapshots and $\mathcal P_t$ is the set of
admitted Plans. Here $\mathcal B$ is the fixed experimental configuration,
$\mathcal Z$ is the
task-specific artifact space, $\mathcal K^R_t$ is Run Memory,
$\mathcal K^P_{p,t}$ is the Plan Memory of research direction $p$, and
$\mathcal L_t$ is the operator-intervention ledger. Only the three latter
states evolve during a run; neither $\mathcal B$ nor $\mathcal Z$ is an agent
decision. The complete controller state additionally records active scopes,
queues, attempts, budget counters, catalog and~ranking; the projection above
does not replace that operational state.

At a planning boundary, a Coordinator receives the immutable view
\begin{equation}
  \mathcal V_p
  =
  \bigl(
    \mathcal K^R_t,\mathcal C_t,\mathcal R_t,\beta_t,
    \mathcal B,\mathcal Z
  \bigr),
  \qquad
  (\eta_p,\rho_p)\sim\pi(\,\cdot\mid\mathcal V_p),
  \label{eq:coordinator-planning-view}
\end{equation}
where $\eta_p$ is the research direction pursued by Plan $p$,
$\mathcal C_t$ contains admitted and active Plan directions,
$\mathcal R_t$ is the current Trial ranking, $\beta_t$ is the remaining
budget, and $\rho_p$ records the new direction's relation to prior or active work.
Planning views are frozen per call. Plan admission is serialized against
$\mathcal C_t$, while an admitted Plan may advance asynchronously.

Trials within a Plan advance serially. When Trial $i$ in Plan $p$ is created,
the Harness freezes the current Plan- and
Run-Memory bases $b_i=(b_i^P,b_i^R)$ and the accepted Plan commitment
$\psi_i=(h_i,\widehat e_i,\phi_i,c_i)$, comprising the hypothesis and
proposed mechanism $h_i$, expected observations $\widehat e_i$, rejection~conditions $\phi_i$, and fixed controls $c_i$. The commitment is inherited from
the immutable Plan; creating a later Trial does not itself launch a new
Coordinator proposal or authorize rewriting that Plan. The indices $b_i^P$
and $b_i^R$ identify immutable Memory versions; $i$
identifies a Trial within the Run. Its scientific record is
\begin{equation}
  \tau_i =
  \bigl(\psi_i,z_i,b_i,\theta_i,\xi_i,e_i,q_i,r_i,a_i,s_i\bigr),
  \label{eq:trial-record}
\end{equation}
where $z_i$ is the admitted artifact, $\theta_i$ the search-selected checkpoint,
$\xi_i$ an immutable execution record, $e_i$ the evidence package,
$(q_i,r_i)$ the frozen review design and Review Worker tool output, $a_i$ the
blinded evidence interpretation, and $s_i$ a typed terminal status. For an
inadmissible artifact or failed execution,
unavailable fields take the value $\bot$ rather than being silently omitted.

\subsection{Hypothesis Formation and Intervention Rules}

The Coordinator must explain why a direction is worth testing in light of
accepted evidence and remaining budget. It may explore a new direction,
revisit or contradict a completed Plan, or combine findings from multiple
completed Plans. No relation is a preset default.

The comparison expectation is specified separately for the task's primary
and secondary metrics. Each can require strict improvement, allow no
regression, or request a diagnostic comparison. A new direction is compared
with the generic baseline; a revisit or contradiction uses its sole source
Plan, while a combination uses its first declared source. The Harness selects
that Plan's eligible representative Trial and freezes a separate portfolio
comparison target. This distinguishes testing the proposed claim from asking
whether the result is the Run's best.

The Builder realizes this commitment under the task-specific contracts in
Appendix~\ref{app:artifact-and-package}. Development feedback may guide a
revision within the same Trial; changing the hypothesis requires a new~Plan.

\subsection{Strategy Implementation and Experiment Packaging}
\label{app:artifact-and-package}

For Trial $i$ in Plan $p$, one \adecomponent{Builder} receives the frozen
commitment $\psi_i$, frozen Plan Memory $\mathcal K^P_{p,b_i^P}$, a
task-declared empirical package $\mathcal D^{\mathrm{build}}_i$, and any
admitted parent artifact. Its first
realization is
\begin{equation}
  z_i^{(0)}
  =
  \Gamma_{\mathcal Z}
  \bigl(\psi_i;\mathcal K^P_{p,b_i^P},\mathcal D^{\mathrm{build}}_i,z_i^{\mathrm{parent}}\bigr),
  \qquad z_i^{(0)}\in\mathcal Z,
  \label{eq:evidence-grounded-realization}
\end{equation}
The empirical package contains the schemas, examples, statistics, and
bounded probes declared by the task. Together with frozen Plan Memory and
admitted source artifacts, it defines the Builder's input view. Held-out evaluation, undeclared files,
and \adecomponent{Engine} logs are outside this view.

The \adecomponent{Harness} checks and realizes the proposal before training.
Starting from draft $z_i^{(0)}$, for at most $J$ reflection revisions,
\begin{equation}
  o_i^{(j)}
  =\operatorname{Realize}
    \bigl(z_i^{(j)};\mathcal D^{\mathrm{build}}_i\bigr),
  \qquad
  z_i^{(j+1)}
  =\operatorname{Reflect}
    \bigl(\psi_i,z_i^{(j)},o_i^{(j)}\bigr),
  \quad j=0,\ldots,J-1,
  \label{eq:builder-reflection-loop}
\end{equation}
while $\psi_i$, the data and source bindings, the intervention space, and the
runtime remain unchanged. These revisions remain within the same logical Trial
and may repair interface or execution failures on fixed task inputs; they may
not consult additional target-model performance results or broaden the
intervention. The Builder may finalize early; $J$ bounds revisions rather
than requiring them, and the final candidate must pass admission before
training. A failure to
admit produces a typed terminal Trial~record rather than an omitted Trial. A
later Builder in the same Plan belongs to a new, sequential Trial.

For an admitted artifact, the fixed binder creates
$x_i=\operatorname{Bind}(z_i;\mathcal B)$. The
\adecomponent{Engine} and \adecomponent{Harness} then form the evidence state
\begin{equation}
  (\theta_i,\xi_i,\mathcal A_i,s_i)
  =\operatorname{Execute}(x_i),
  \qquad
  e_i=\operatorname{Package}
    (\xi_i,\mathcal A_i;\Sigma_{\mathrm{task}}),
  \label{eq:engine-evidence}
\end{equation}
where $\theta_i$ is the selected checkpoint, $\xi_i$ is the immutable execution
record, $\mathcal A_i$ is the raw artifact set, and $s_i$ is the terminal
status. The task-owned specification $\Sigma_{\mathrm{task}}$ selects the
identity-, source-, training-, and model-behavior records available to later
scientific roles. Large records may remain durable references. Operational
telemetry and held-out evaluation are not members of $e_i$.

\textbf{Task-specific intervention contracts.}
The Builder delivers an executable strategy and a scientific account of its
implementation, source inheritance, and limitations. The four experimental
tasks instantiate three intervention spaces, specified below. Candidate-pool
sizes, selection counts, and schedule dimensions for the experiments are
given in Appendix~\ref{app:experimental-configuration}.

\begin{adecontractbox}{Data selection}
\textbf{Inputs.} The fixed candidate inventory, required subset size, and an
optional task-authorized Judge that scores responses against a declared rubric.

\textbf{Output.} The required number of distinct, authorized complete-trajectory
identifiers from the fixed candidate inventory.

\textbf{Constraints.} Selection may combine quality scores, length, task
features and group structure. Ranking, quotas and deterministic backfill are
part of the proposed strategy; training and evaluation remain fixed. The
selector cannot read external files or call external services other than the
provided Judge capability.
\end{adecontractbox}
The Harness checks executable validity, membership and cardinality against
the fixed inventory. When Judge enrichment is enabled, the selector must use
it on a valid execution path. The rubric specifies observable criteria,
allowed score levels in $[0,1]$, and aggregation weights summing to one.
Judge scores are evidence for the selector, not selection decisions.

\begin{adecontractbox}{Reward design}
\textbf{Response-level inputs.} The problem, generated response, extracted
answer, externally verified outcome, response length and maximum length.

\textbf{Group-level inputs.} All sibling responses for one prompt, with
outcome, process, rule and length evidence and their pre-group rewards.

\textbf{Output.} A finite training reward in $[0,1]$ for each response,
with the decision and evidence used to assign it.

\textbf{Constraints.} Group credit is deterministic and invariant to sibling
ordering or identifier changes. It cannot perform external I/O or request
new Judge assessments; it returns rewards, leaving advantage estimation to
the fixed training algorithm.
\end{adecontractbox}
The response-level record keeps the final score separate from the
authoritative outcome, optional process projection, and rule evidence;
unavailable evidence is explicit. With group credit enabled, the row-level
reward retains the outcome, the Engine acquires process evidence, and the
group function assigns the training rewards. Admission probes include
mixed-correctness groups and check the realized GRPO advantage signs:
correct responses retain positive advantages and incorrect responses retain
non-positive advantages in these mixed groups. Passing establishes usable
reward behavior under the probes; its scientific value is evaluated through
controlled training.

\begin{adecontractbox}{Curriculum learning}
\textbf{Inputs.} The tokenizer-eligible problem inventory, subject and level
metadata, fixed schedule length and batch size, and an optional task-authorized
Judge.

\textbf{Output.} A complete schedule with the required number of batches, each
containing the prescribed number of distinct authorized problem identifiers.
Problems may recur across batches.

\textbf{Constraints.} The schedule is realized and frozen before training.
The strategy controls batch composition and cross-step exposure; it cannot
change problem contents, rewards, the optimizer, or evaluation, and receives
no training-time feedback while constructing the schedule.
\end{adecontractbox}
The Harness checks batch count, cardinality and inventory membership, then
canonicalizes the identifiers within each batch. The realized schedule and
consumed problem identities support later analysis of coverage and reuse.

\textbf{Evidence contents and attribution.}
Each role receives a fixed input inventory identifying the task, current
Run/Plan/Trial, research-state version, and sources it may inspect. The
Analyzer's Experiment Package includes an evidence catalog describing the
available interventions, measurements and behavior records. Large records
remain individually accessible through indexed references.
Table~\ref{tab:package-contract} summarizes the package; the role-specific
uses of these records appear in Appendix~\ref{app:agent-protocols}.

\begin{table}[!t]
\centering\adeappendixtablestyle
\caption{Analyzer package contents and exclusions.}
\label{tab:package-contract}
\resizebox{0.95\linewidth}{!}{%
\begin{tabularx}{0.95\linewidth}{@{}L{0.23\linewidth}Y@{}}
\toprule
Record & Bound contents and use \\
\midrule
Identity and catalog & Run/Coordinator/Plan/Trial scope, artifact identity,
kind, availability, source experiment and record-level attribution. \\
Realized intervention & Executed strategy and final realization report; selected trajectories,
reward/group evidence, or curriculum batches and consumed problem identities. \\
Training and behavior & Normalized training telemetry, task-permitted rollout
records and complete fixed-checkpoint in-loop responses with assessments. \\
Review overlay & Frozen packet and coverage, added only for synthesis;
missing and failed rows remain explicit. \\
Excluded context & Accepted Plan and rationale, PM/RM, planning comparators,
objective-comparison record, online validation results, held-out records,
and undeclared operational files. \\
\bottomrule
\end{tabularx}%
}
\end{table}
The bootstrap reference has no Builder realization; its executed baseline
artifacts replace that part of the package. This exception does not apply to
ordinary Search Trials. Package projection defines the role-specific input
boundary described in
Appendix~\ref{app:context-boundaries}.

\subsection{Independent Evidence Interpretation}

The Analyzer uses the Experiment Package in
Table~\ref{tab:package-contract}. Its interpretation is based on the realized
intervention and permitted measurements, with the Plan, expectations, and
Memory excluded. Analysis has two stages: first the Analyzer designs a
record-level review; then it interprets the frozen review results alongside
the original evidence. The Review Worker executes the requested assessments
between these stages.

SFT review covers all selected training examples and all in-loop evaluation
responses. Candidate data supports comparison with the selected subset but
is not itself a record-level Review pool. RFT review likewise covers all
in-loop responses and requires at least the task-declared 25\% coverage of
training rollout evidence using group-based selection. Requests may select
all records, specified records, or complete groups where the task permits.
The Harness validates coverage and scope, binds the request to exact records,
executes the Review Worker, and freezes the results and coverage before
resuming the Analyzer.

SFT analysis connects candidate and selected distributions to supervision
quality and downstream behavior. Reward-design analysis connects outcomes, process evidence, assigned rewards
and realized advantages to training and validation. Curriculum analysis joins
scheduled and consumed problem identities to rollout evidence, examining
subject/level composition, coverage, reuse, and cohort outcomes. Each analysis
links its findings to the underlying measurements.

\subsection{Hypothesis Assessment and Memory Updates}

The Plan Summarizer receives both the original commitment and the blinded
interpretation, together with the parent PM, objective Trial outcome,
realization record and frozen comparisons. Missing analysis or unavailable
evidence remains explicit.

Hypothesis-aware summarization proposes the adjudicated findings
\begin{equation}
  \begin{aligned}
  \widetilde\delta_i
    &=\operatorname{Adjudicate}(\psi_i,a_i,e_i,s_i)\\
    &=\{(\gamma_{ij},v_{ij},\mathcal R_{ij})\}_{j=1}^{m_i}.
  \end{aligned}
  \label{eq:typed-findings}
\end{equation}
Here $\gamma_{ij}$ is a claim, $v_{ij}$ the Summarizer's model-generated
assessment of how the evidence relates to it, and $\mathcal R_{ij}$ its
source references; $m_i$ is the number of
such claims. These claim assessments form part of the Memory update alongside
observations and unresolved questions.
Admission then returns $\delta_i=\operatorname{Admit}
(\widetilde\delta_i;\tau_i)$ when the identities, source references, coverage,
and status bindings are valid. Each element of $\delta_i$ retains its sources
$\mathcal R_{ij}$ in $\tau_i$. Thus an interpretation can qualify evidence but
cannot replace its intervention, execution status, or measurements, and an
adjudication cannot retrospectively rewrite $\psi_i$.

The Trial keeps three distinct conclusions:
\begin{itemize}[leftmargin=*,itemsep=0.2em]
\item \textbf{Realization:} verified, deviated, or unverified.
\item \textbf{Hypothesis:} the Summarizer's assessment against the precommitted
expectation and rejection~condition.
\item \textbf{Portfolio:} new best, not new best, or not comparable.
\end{itemize}
Only the second is a model-generated interpretation. Each finding remains
attributable to its supporting evidence.

The Harness admits and publishes each memory version. An agent's proposed
text does not change shared research state until that publication succeeds.

\subsection{Versioned State Publication and Queue Semantics}
\label{app:update-queue}

Agents propose Memory updates; the Harness admits and publishes them.
Here $k$ indexes Plan-Memory versions for Plan $p$, and $j$ indexes
Run-Memory versions in their serialized publication order. After Trial $i$
in Plan $p$, the
\adecomponent{Plan Summarizer} proposes a successor from the parent Plan Memory,
immutable Plan, Trial Record, and blinded interpretation:
\begin{equation}
  \widetilde{\mathcal K}^{P}_{p,k+1}
  =U_P\!\left(
    \mathcal K^P_{p,k};\psi_i,\tau_i,a_i,\delta_i
  \right),
  \qquad
  \mathcal K^P_{p,k+1}
  =\operatorname{Publish}_P
    (\widetilde{\mathcal K}^{P}_{p,k+1};\tau_i).
  \label{eq:versioned-plan-memory-update}
\end{equation}
Publication creates an immutable Plan-Memory version containing its identifier,
parent, revision, source Trial identifiers, source digests, and admitted
findings. All Trials remain append-only records, including unsuccessful
experiments whose observations can inform later proposals. The resulting lineage is
\begin{equation}
  (\mathcal K^P_{p,k},\mathcal K^R_j)
  \longrightarrow (\psi_i,z_i,\tau_i,a_i,\delta_i)
  \longrightarrow
  (\mathcal K^P_{p,k+1},\mathcal K^R_{j+1}).
  \label{eq:research-memory-lineage}
\end{equation}

After $\mathcal K^P_{p,k+1}$ is accepted, that completed Trial's Plan Memory
result is enqueued exactly once under a persistent commit key. If $Q_j$ is the
durable FIFO queue and $\Delta_{\sigma(j)}=\operatorname{head}(Q_j)$ denotes its
next Plan Memory update, the single-writer transition is
\begin{equation}
  \mathcal K^R_{j+1}
  =\operatorname{Publish}_R\!\left(
    U_R(\mathcal K^R_j,\Delta_{\sigma(j)})
  \right),
  \qquad
  Q_{j+1}=\operatorname{pop}(Q_j).
  \label{eq:ordered-run-memory-commit}
\end{equation}
Only the queue head may change Run Memory. Publication and queue removal form
one logical commit: after recovery, an already published commit key is
acknowledged rather than merged again. New planning calls see
$\mathcal K^R_{j+1}$; already running calls retain their frozen bases. Thus
parallel Trial completion order affects commit order, but cannot create a
partially visible or duplicate \mbox{global update}.

\subsection{Recovery and Operator Semantics}
\label{app:recovery-semantics}

A logical Trial may have several physical attempts. Let
$\alpha_{i,m}=(i,m,x_i,b_i,\ell_{i,m},\upsilon_{i,m},s_{i,m})$ record attempt
$m$, where $x_i$ is the frozen command, $b_i$ the frozen Memory bases,
$\ell_{i,m}$ the physical resource assignment, $\upsilon_{i,m}$ the attempt
telemetry, and $s_{i,m}$ its status. A retry is admissible only for a declared
retryable failure and remaining recovery budget, and must satisfy
\begin{equation}
  \operatorname{Retry}(\alpha_{i,m})=\alpha_{i,m+1}
  \quad\Longrightarrow\quad
  (x_{i,m+1},b_{i,m+1},\psi_{i,m+1},z_{i,m+1})
  =(x_i,b_i,\psi_i,z_i).
  \label{eq:recovery-replay}
\end{equation}
The new attempt has a fresh identity and may receive different physical
resources, but cannot alter the logical experiment. The failed attempt remains
fenced and durable; exhausting the recovery budget yields a terminal failure or
suspends the Run according to the declared policy.

Human pause, resume, graceful-finish, cancel, and held-out-evaluation actions
append to the operator ledger. For operator action $o_t$,
\begin{equation}
  \mathcal L_{t+1}
  =\operatorname{Append}(\mathcal L_t,o_t),
  \qquad
  \operatorname{Rewrite}_{o_t}
  (\psi_i,z_i,\tau_i,\mathcal K^P,\mathcal K^R)
  =\emptyset.
  \label{eq:operator-boundary}
\end{equation}
Operational components may observe health, allocate resources, or initiate an
admissible replay, but have no scientific write authority. Held-out evaluation
results are operational records, not research state, and cannot become inputs to
proposal, ranking, adjudication, Memory publication, or Plan and Trial selection.

\subsection{Context and Information-Access Boundaries}
\label{app:context-boundaries}

Every Call receives a role-specific instruction set and a fixed inventory of
readable evidence. The shared instruction is to inspect only declared inputs,
use permitted attempt-local working material, and return the required
semantic output. On retry, the agent receives its prior delivery and targeted
feedback, repairs only the identified parts, and preserves unaffected content.
Reflection, staged analysis and physical retries retain their original
logical research commitment.

Context is selected by role and research-state version; large records remain
accessible through indexed references. Model and decoding settings are given
in Appendix~\ref{app:experimental-configuration}. When analysis attempts are
exhausted, the failure is recorded and unavailable evidence retains its status.

Role-specific input packages separate the information supplied to each agent.
The Analyzer's package excludes the Plan, hypothesis, and research Memory;
canonical state changes pass through Harness admission and publication.

\subsection{Stage Semantics}
\label{app:stage-semantics}

Table~\ref{tab:ade-stage-semantics} summarizes the transitions defined above.
It separates scientific decisions from the checks and publication steps that
make their outputs available to subsequent roles.

\begin{table}[!t]
  \centering\adeappendixtablestyle

  \caption{Stage semantics of a \name{} research iteration.}
  \label{tab:ade-stage-semantics}
  \resizebox{0.95\linewidth}{!}{%
  \begin{tabularx}{0.95\linewidth}{@{}L{0.17\linewidth}C{0.30\linewidth}Y@{}}
    \toprule
    Stage & Abstract transition & Scientific boundary \\
    \midrule
    Hypothesis commitment
      & $\mathcal V_p\mapsto(\eta_p,\rho_p)$
      & The Plan commits its question, mechanism, expectations, rejection~conditions, and controls. Each Trial inherits that commitment.
        The catalog in $\mathcal V_p$ exposes active directions to help avoid
        duplicate proposals. \\
    Intervention admission
      & $(\psi_i,\Gamma_{\mathcal Z})\mapsto z_i\in\mathcal Z$
      & Interface and execution checks establish that the artifact can run
        under the fixed configuration; the permission boundary restricts it to
        the declared selection, reward, or curriculum surface. \\
    Measurement
      & $(z_i,\mathcal B)\mapsto(\theta_i,\xi_i,e_i,s_i)$
      & Training and evaluation execute outside the proposing agent. Recovery
        may replay the same frozen logical experiment, but cannot change
        $\psi_i$, $z_i$, or $\mathcal B$. \\
    Blinded evidence interpretation
      & $(z_i,\xi_i,e_i)\mapsto(q_i,r_i,a_i)$
      & Without receiving $\psi_i$, the Analyzer commits its review design
        before the Review Worker tool is called, then binds the returned
        review to primary evidence in the blinded interpretation. \\
    Hypothesis adjudication
      & $(\psi_i,a_i,e_i,s_i)\mapsto\widetilde\delta_i$
      & Plan-level synthesis compares the independently formed interpretation
        with the precommitted expectation and rejection~condition. It may
        qualify the claim but cannot rewrite it. \\
    Plan update
      & $(\mathcal K^P_{p},\tau_i,\delta_i)
        \mapsto\mathcal K^{P+}_{p}$
      & Findings with valid source references cross the local knowledge
        boundary with their assessments preserved; the Trial is admitted
        exactly once. \\
    Run update
      & $(\mathcal K^R_j,\mathcal K^P_{p,k+1})
        \mapsto\mathcal K^R_{j+1}$
      & Every completed Trial's Plan Memory result enters one serialized merge
        order even when Plans run concurrently. \\
    Operator boundary
      & $(\mathcal L_t,o_t)\mapsto\mathcal L_{t+1}$
      & Pause, resume, graceful finish, and held-out evaluation are durable and
        auditable; they cannot rewrite canonical scientific state. \\
    \bottomrule
  \end{tabularx}%
  }
\end{table}

\subsection{Controlled and Cumulative Execution}
\label{app:protocol-pseudocode}

Algorithm~\ref{alg:ade-loop} gives the logical protocol. ``Asynchronously''
means that construction, training, review, and analysis for different Plans may
overlap. Planning admission and Run-Memory commitment remain serialized because
they update shared research state. $B_P$ is the Plan budget and $B_T$ is the
maximum number of Trials per Plan. The planning-opportunity check includes
remaining budget, available capacity, and the absence of a stop boundary.
Build includes realization, bounded reflection, and artifact admission;
execution includes the recovery rules above. Memory updates below denote
accepted, published versions. Unavailable review or analysis remains explicit
in the Trial~record.

\begin{algorithm}[!t]
\small
\refstepcounter{algorithm}
\hrule
\vspace{0.5ex}
\noindent\textbf{Algorithm \thealgorithm}\quad Controlled and cumulative \name{} research
\label{alg:ade-loop}
\vspace{0.5ex}
\hrule
\vspace{0.6ex}
\begin{algorithmic}[1]
\Require fixed experimental configuration $\mathcal B$; artifact space $\mathcal Z$;
         budgets $B_P,B_T$; declared parallel capacity; initial Run Memory $\mathcal K^R_0$
\State Initialize active Plans $\mathcal P\gets\emptyset$ and ledger
       $\mathcal L\gets\emptyset$; initialize the RM update queue and budget counters
\While{a planning opportunity remains, a Plan is active, or an RM update is queued}
  \Statex \emph{Serialized research-direction commitment}
  \If{a planning opportunity remains}
    \State Freeze $\mathcal V_p$; $(\eta_p,\rho_p)\gets\Call{ProposeDirection}{\mathcal V_p}$
    \If{the proposal passes Plan admission against the current catalog}
      \State admit Plan $p$ with its frozen basis; initialize PM and its Trial counter
      \State add $p$ to $\mathcal P$ and account for its Plan budget
    \EndIf
  \EndIf
  \Statex \emph{Asynchronous experimental advancement}
  \ForAll{Plans $p\in\mathcal P$ eligible to start their next Trial under $B_T$}
    \State $\psi_i\gets\Call{InheritPlanCommitment}{\eta_p}$;
           freeze bases $b_i=(b_i^P,b_i^R)$ and $\psi_i$; account for this Trial
    \State $z_i\gets\Call{Build}{\psi_i,\mathcal K^P_{p,b_i^P},\mathcal D^{\mathrm{build}}_i,z_i^{\mathrm{parent}},\mathcal Z}$
    \If{$z_i\notin\mathcal Z$}
      \State $(\theta_i,\xi_i,e_i,q_i,r_i,a_i,s_i)
        \gets(\bot,\bot,\bot,\bot,\bot,\bot,\textsc{BuilderFailed})$
    \Else
      \State $(\theta_i,\xi_i,e_i,s_i)\gets\Call{Execute}{z_i,\mathcal B}$
      \If{interpretable evidence is available}
        \State $q_i\gets\Call{DesignReview}{z_i,\xi_i,e_i}$
        \State $r_i\gets\Call{ReviewWorker}{q_i,e_i}$
        \State $a_i\gets\Call{BlindAnalyze}{z_i,\xi_i,e_i,r_i}$
      \Else
        \State $(q_i,r_i,a_i)\gets(\bot,\bot,\bot)$
      \EndIf
    \EndIf
    \State $\tau_i\gets
      (\psi_i,z_i,b_i,\theta_i,\xi_i,e_i,q_i,r_i,a_i,s_i)$
    \State $\widetilde\delta_i\gets
      \Call{Adjudicate}{\psi_i,a_i,e_i,s_i}$;
      $\delta_i\gets\Call{Admit}{\widetilde\delta_i,\tau_i}$
    \State $\mathcal K^P_p\gets
      \Call{UpdatePlan}{\mathcal K^P_p,\tau_i,\delta_i}$;
           archive $\tau_i$ exactly once
    \State $\Delta_i\gets\mathcal K^P_p$;
           enqueue this Trial's accepted PM version exactly once
    \State Mark $p$ terminal if its Trial budget is exhausted or its stopping rule applies
    \State Block this Plan's next Trial until its RM merge commits
  \EndFor
  \Statex \emph{Serialized scientific commitment}
  \ForAll{queued $\Delta_i$ in commit order}
    \State $\mathcal K^R\gets\Call{Merge}{\mathcal K^R,\Delta_i}$ exactly once
    \State Acknowledge and dequeue the committed update
    \State Remove its Plan from $\mathcal P$ if terminal; otherwise unblock its next Trial
  \EndFor
\EndWhile
\State \Return $(\mathcal K^R,\{\tau_i\}_{i},\mathcal L)$
\end{algorithmic}
\vspace{0.4ex}
\hrule
\end{algorithm}

\subsection{Protocol Invariants}
\label{app:protocol-invariants}

The protocol maintains five invariants. First, scientific control requires
$\mathcal B_t=\mathcal B_0$ and $\mathcal Z_t=\mathcal Z_0$ throughout a
qualifying run. Second, $\psi_i$ is frozen before execution and is absent from
the \adecomponent{Analyzer}'s information set. Third, proposal, measurement, blinded
interpretation, semantic adjudication, and mechanical admission have distinct
authorities: only Harness admission and publication change canonical
knowledge. Fourth, each terminal Trial is archived and contributes exactly
one Plan-Memory update and one Run-Memory update, including failed Trials.
Suspended work resumes from its preserved state under the recovery policy. Fifth,
held-out results do not influence search:
\begin{equation}
  e^{\mathrm{hold}}_i \notin \operatorname{View}_{\mathrm{agent}}(t),
  \qquad
  o_t \text{ may update }\mathcal L_t
  \text{ but may not rewrite }(\psi_i,z_i,\tau_i,\mathcal K_t).
  \label{eq:heldout-noninterference}
\end{equation}
These invariants make research decisions, experimental outcomes, and their
use in subsequent proposals traceable throughout a controlled and cumulative Run.

\section{Experimental Configuration}
\label{app:experimental-configuration}
\label{app:experimental-specification}

This appendix specifies the models, data, training recipes, research budgets,
and comparison protocols used in Section~\ref{sec:experimental-design}.
Task interfaces are defined in Appendix~\ref{app:artifact-and-package};
benchmark splits, metrics, decoding, and checkpoint selection are specified
in Appendix~\ref{app:benchmark-details}. Unless stated otherwise,
the experiment configuration uses a fixed seed of 42.

\subsection{Target-Model and Data Configuration}

\begin{table}[!t]
  \centering\adeappendixtablestyle

  \caption{Target-learning configuration.}
  \label{tab:target-learning-configuration}
  \setlength{\tabcolsep}{3pt}
  \resizebox{0.95\linewidth}{!}{%
  \begin{tabularx}{0.95\linewidth}{@{}L{0.16\linewidth}Y Y Y@{}}
    \toprule
    Parameter & Data Selection--Math & Data Selection--Code & Math RFT tasks \\
    \midrule
    Target model & Qwen2.5-7B-Instruct & Qwen2.5-7B-Instruct & Qwen2.5-0.5B \\
    Training source & OpenThoughts mixed pool & OpenThoughts mixed pool & Fixed MATH training set \\
    Source size & 3,840 trajectories & 3,840 trajectories & 3,072 problems \\
    Intervention & Select 384 trajectories & Select 384 trajectories & Reward/group credit or data~schedule \\
    Learning method & Full-parameter SFT & Full-parameter SFT & GRPO \\
    \bottomrule
  \end{tabularx}%
  }
\end{table}

\textbf{Data selection.}
The shared source is OpenThoughts-114k. The pool builder allocates 3,840
rows proportionally across matched math, code and science groups using
largest-remainder counts, per-group seeded reservoir sampling, and a final
seeded shuffle. Both SFT tasks use this same fixed mixed-domain pool;
math versus code denotes the downstream evaluation objective. The builder
records availability and realized group counts.

Each OpenThoughts candidate is one complete user--assistant trajectory. The
assistant target contains one paired \code{<think>}/\code{</think>} reasoning
span followed by a nonempty final answer. A selector must return exactly 384
distinct complete trajectories; it cannot split a trajectory or compensate for
cardinality with a post-hoc token-matched subset. Training examples are capped
at 16,384 tokens.

\textbf{Reward design and curriculum learning.}
Curriculum Learning--Math uses the same target model, 3,072-problem MATH
pool, and 100-step GRPO budget as Reward Design--Math. Its intervention
is a precomputed schedule of 100 batches, each containing 256 distinct
authorized problem identifiers. Problems may recur across batches; the
outcome-only reward remains fixed.
The tokenizer limit excludes three overlength prompts, leaving 3,069 eligible
problems for scheduling. Construction of the MATH training and evaluation
splits is described in Appendix~\ref{app:benchmark-details}. The GRPO settings in Table~\ref{tab:rft-training-configuration}
apply to both math RFT tasks; curriculum learning disables data and actor
shuffling to preserve its precomputed batches.

\subsection{Generic Baseline Configuration}

The base target is evaluated before task-specific training. The generic
baseline is the reference intervention trained under the same task conditions
as the agent-developed strategies.

For both data-selection settings, the generic reference is one immutable
random collection of 384 complete trajectories from the same 3,840-item
pool. The collection is sampled once before research and reused rather than
redrawn for each proposal.

For reward design, let $O(x,y)\in\{0,1\}$ be externally verified answer
correctness for response $y$ to problem $x$. The generic reference is
\begin{equation}
  R_{\mathrm{ref}}(x,y)=O(x,y),
  \qquad
  R^{\mathrm{group}}_{\mathrm{ref}}(x,y_j)=O(x,y_j),
  \label{eq:outcome-only-reference}
\end{equation}
so neither response-level shaping nor cross-response redistribution changes the
outcome reward. It is trained with the same GRPO and evaluation configuration
as every proposed reward.

Curriculum learning shares the fixed base model and trained generic
reference with reward design. That reference uses outcome-only rewards and
randomly shuffled training data.

The configuration is frozen at Run initialization and retained with its results.

\subsection{SFT Training Configuration}

Table~\ref{tab:sft-training-configuration} reports the common training
configuration applied to the fixed-random reference and every agent-selected
dataset in both SFT settings.

\begin{table}[!t]
  \centering\adeappendixtablestyle

  \caption{Shared SFT configuration for the math and code data-selection
  settings.}
  \label{tab:sft-training-configuration}
  \resizebox{0.85\linewidth}{!}{%
  \begin{tabularx}{0.85\linewidth}{@{}L{0.38\linewidth}Y@{}}
    \toprule
    Parameter & Value \\
    \midrule
    Training type & Full-parameter supervised fine-tuning \\
    Selected examples & 384 \\
    Maximum epochs & 13 \\
    Effective global batch & 32 examples \\
    Optimizer steps per full epoch & 12 \\
    Optimizer & AdamW \\
    Optimizer betas / epsilon & $(0.9,0.999)$ / $1\times10^{-8}$ \\
    Weight decay / maximum gradient norm & $0$ / $1.0$ \\
    Learning rate & $1\times10^{-5}$ \\
    Schedule / warmup & Cosine / 0.1 ratio \\
    Input cutoff & 16,384 tokens \\
    Checkpoints & Every epoch; all retained \\
    Online early stopping & Four consecutive non-improving epochs \\
    Output format & Learned tagged reasoning followed by a final~answer \\
    \bottomrule
  \end{tabularx}%
  }
\end{table}

The SFT backend uses LlamaFactory full-parameter training with the Qwen
chat template, DeepSpeed ZeRO-3 and bfloat16. The configuration adds
\code{<think>} and \code{</think>} as special tokens and resizes the
vocabulary; their generation is learned through SFT. With eight training GPUs,
one example per device and four accumulation steps, the effective batch is
32. User prompts are not training targets.
The learned reasoning protocol is separate from the base model's disabled
native thinking~protocol.

\subsection{RFT/GRPO Training Configuration}

Table~\ref{tab:rft-training-configuration} gives the common GRPO recipe.
Reward design changes the reward and group-credit functions; curriculum
learning changes the training-data schedule while retaining outcome-only credit.
Answer correctness is computed externally. Proposed response rewards remain
in $[0,1]$, and group credit is deterministic within each prompt group.

\begin{table}[!t]
  \centering\adeappendixtablestyle

  \caption{Shared math RFT training configuration.}
  \label{tab:rft-training-configuration}
  \resizebox{0.85\linewidth}{!}{%
  \begin{tabularx}{0.85\linewidth}{@{}L{0.38\linewidth}Y@{}}
    \toprule
    Parameter & Value \\
    \midrule
    Algorithm & GRPO \\
    Training problems & 3,072 \\
    Actor updates / data epochs & 100 / at most 20 \\
    Prompt groups per update & 256 \\
    Sampled responses per prompt & 8 \\
    Generated trajectories per update & 2,048 \\
    PPO mini-batch / epochs & 256 prompt groups / 1 \\
    Optimizer / betas / gradient clip & AdamW / $(0.9,0.999)$ / 1.0 \\
    Learning rate / schedule & $5\times10^{-7}$ / constant \\
    Warmup ratio / weight decay & 0 / 0.01 \\
    Prompt / response limits & 1,024 / 8,192 tokens \\
    Rollout sampling & Temperature 1, top-$p=1$ \\
    PPO clip / entropy coefficient & 0.2 / 0.001 \\
    Actor KL loss & Coefficient $1\times10^{-4}$ \\
    Group credit & Enabled; deterministic within prompt group \\
    Checkpoints & Every 4 updates; 25 positions through~update~100 \\
    \bottomrule
  \end{tabularx}%
  }
\end{table}

\subsection{Research Budget and Parallel Capacity}

\textbf{Candidate budget.}
For the reported research settings, the budget is five Plans per Coordinator
with one Trial per Plan. Table~\ref{tab:complete-search-budgets} therefore
reports a capacity of five
agent-developed candidates for $C=1$ and fifteen for $C=3$, after the single
generic reference. The parallel settings have no round barrier; Coordinators
may advance their own Plans while admitted evidence is accumulated in shared
Run Memory.

\begin{table}[!t]
  \centering\adeappendixtablestyle

  \caption{Research capacity per Run, shared by all four tasks. Counts exclude
  the base-target evaluation and generic reference.}
  \label{tab:complete-search-budgets}
  \setlength{\tabcolsep}{3pt}
  \resizebox{0.95\linewidth}{!}{%
  \begin{tabularx}{0.95\linewidth}{@{}Y C{0.15\linewidth}C{0.17\linewidth}C{0.14\linewidth}C{0.16\linewidth}@{}}
    \toprule
    Capacity & Coordinators & Plans per Coordinator & Trials per Plan
      & Maximum candidates \\
    \midrule
    $C=1$ & 1 & 5 & 1 & 5 \\
    $C=3$ & 3 & 5 & 1 & 15 \\
    \bottomrule
  \end{tabularx}%
  }
\end{table}

Research budgets are specified by candidate counts. Elapsed time, agent-token
usage, and GPU-hours are measured outcomes in \hyperref[sec:rq2]{RQ2}.
Moving from one to three Coordinators increases candidate capacity from five
to fifteen for every task.

\textbf{Research agents and evidence review.}
All research-agent roles use \code{gpt-5.6-sol} with \code{xhigh} reasoning
effort. Evidence review uses \code{Qwen3.6-35B-A3B} to interpret experimental records.

The Review provider uses temperature 0, disabled thinking, a
4,096-token completion limit, at most 128 concurrent requests and three
attempts per request. Builder reward replay is configured for 32 prompt
groups.

A Builder may reflect at most twice before delivery. Agent execution has a
configured retry allowance of three, and Engine automatic recovery permits
at most three attempts.

\textbf{Hardware and resource accounting.}
The Run-level GPU lease is $24C$ for SFT and $16C$ for RFT, where $C$
is the number of Coordinators. Within each Coordinator's lease, both training paradigms
use eight GPUs for training and eight for checkpoint-selection evaluation.
In-loop evaluation uses 24 GPUs for SFT and 16 for RFT; Operator held-out
evaluation uses the same respective allocations. Training may overlap
checkpoint-selection evaluation; in-loop and Operator held-out evaluation
do not overlap under the declared resource policy. These counts describe
leased capacity. Experiments use NVIDIA H20 GPUs hosted in eight-GPU workers.
Figure~\ref{fig:rq2} reports estimated allocated GPU-hours and elapsed time; model-token usage and API-equivalent costs appear in
Appendix~\ref{app:scaling-accounting}.

\subsection{Protocol for the Compared Research Methods}
\label{app:comparison-protocol}

The Data Selection--Math comparisons in Section~\ref{sec:rq3} use the same task-defined
candidate pool, target-model initialization, training recipe, per-candidate
training budget, checkpoint rule, benchmark splits, and in-loop metric ordering.
The alternatives compare different research organizations and feedback settings.

\textbf{Unified Agent.}
One persistent session plans, implements, and analyzes candidates using raw
experimental evidence and execution facts. Prescribed role handoffs and
structured Plan/Run research memory are removed, while the agent may retain
its own notes. The ADE Controller still allocates five search slots, executes
candidates, and resumes the session after each experiment. Failed candidates
consume their allocated slots; same-Trial retries do not create extra search
opportunities. This comparison changes role decomposition and explicit
Memory organization jointly. Separate Builder reflection calls are disabled,
while candidate realization, validation, and repair remain.

\textbf{Independent Search.}
Each repetition consists of five independently developed initial candidates,
with no access to findings from the other candidates and no later-Plan
refinement. Repetition and selection follow the shared reporting protocol below.

\textbf{Outcome-only Feedback and Raw Evidence Feedback.}
Both retain the Coordinator/Planner and Builder and permit iterative revision
within five search slots. Outcome-only Feedback supplies candidate history,
scalar scores, and execution facts. Raw Evidence Feedback additionally exposes
accepted raw experimental evidence to the Planner, which performs its own
analysis. Both remove the independent
Analyzer, Review, and research summarization workflow. Full ADE combines these
roles with structured research memory.
In both reduced variants, Plan/Run memory versions contain deterministic outcome
records rather than model-generated research summaries, and existing role
conversations remain available. Raw Evidence Feedback exposes accepted historical
experiment packages, including the bootstrap baseline, to the Planner; Builder retains its current
and source Plans, code, fixed task inputs, and candidate implementation feedback.
Analyzer Review is distinct from the task Judge: the reduced variants retain
the task-defined Judge capability and fixed training/evaluation interfaces.

\subsection{Reporting Protocol and Failure Accounting}
\label{app:reporting-protocol}

\textbf{Repetitions and Best selection.}
Each full-ADE setting and each research-method alternative has three independent
repetitions. Each iterative Run has five Plans per Coordinator and one Trial
per Plan; each Independent Search repetition has five independent candidates.
Across three Runs, ADE budgets 15 candidates for $C=1$ and 45 for $C=3$;
each research-method alternative budgets 15 candidates.

Within each Run, the task's ordered in-loop metrics select the best Trial
and its Plan. The same ordering selects the \emph{Best} strategy across the
three Run winners for Table~\ref{tab:effectiveness-results} and the
research-method comparisons. SFT prioritizes Pass@$K$, with Average@$K$
breaking ties; RFT uses the reverse ordering. Each selected checkpoint is
then evaluated on all four held-out benchmarks.

\textbf{Across-Run summaries and trajectories.}
Table~\ref{tab:average-results} reports equal-weight means and sample standard
deviations of the three Run winners. Figure~\ref{fig:rq2} averages three
Runs, with shading from their minimum to maximum. Its first row tracks
best-so-far in-loop scores; the second tracks each Run's current selected
model on the held-out benchmark named in the panel. Both start at Base.
Appendix~\ref{app:complete-effectiveness} reports individual Run results and
per-benchmark research-method comparisons.

\textbf{Failures and retries.}
A failed build, training job, or required evaluation consumes its Plan and
remains recorded with its failure status. An Engine retry re-executes the
frozen intervention and configuration within that Trial. Required evaluations
that fail are recorded as missing; best-so-far curves update on successfully
scored candidates.

\section{Benchmark and Evaluation Details}
\label{app:benchmark-details}

This appendix specifies online checkpoint evaluation, in-loop Trial evaluation,
and held-out evaluation. The first two belong to the adaptive research loop;
the third does not. The same generation, grading, and aggregation rules apply
to the base target, generic baseline, and agent-developed strategies within
each setting and evaluation purpose. Training and research budgets are
specified in Appendix~\ref{app:experimental-configuration}.

\subsection{Benchmark Suites and Evaluation Purposes}

Table~\ref{tab:benchmark-purpose-spec} lists the benchmarks and sampling
counts for each evaluation stage. Data Selection--Math uses AIME 2024
in-loop; Data Selection--Code uses the earlier portion of a temporal split
of LiveCodeBench v5. Reward Design--Math and Curriculum Learning--Math use
MATH-S, a 500-problem subset of MATH, in-loop. Each task is evaluated on four
held-out benchmarks or splits after strategy~selection.

\begin{table}[!t]
  \centering\adeappendixtablestyle

  \caption{Benchmarks by evaluation stage. $K$ is the number of generated
  responses per problem.}
  \label{tab:benchmark-purpose-spec}
  \resizebox{0.95\linewidth}{!}{%
  \begin{tabularx}{0.95\linewidth}{@{}L{0.18\linewidth}Y Y c c@{}}
    \toprule
    Setting & Evaluation stage & Benchmark & Problems & $K$ \\
    \midrule
    Data Selection--Math & Online checkpoint evaluation & AIME 2024 & 30 & 8 \\
    Data Selection--Math & In-loop Trial evaluation & AIME 2024 & 30 & 32 \\
    Data Selection--Math & Held-out evaluation & AIME 2025, HMMT & 30 each & 32 \\
    Data Selection--Math & Held-out evaluation & MATH-500, MATH-S & 500 each & 3 \\
    \midrule
    Data Selection--Code & Online checkpoint evaluation & LiveCodeBench v5, on/before 2024-03-30
      & 439 & 1 \\
    Data Selection--Code & In-loop Trial evaluation & LiveCodeBench v5, on/before 2024-03-30
      & 439 & 3 \\
    Data Selection--Code & Held-out evaluation & LiveCodeBench v5, after 2024-03-30
      & 441 & 3 \\
    Data Selection--Code & Held-out evaluation & HumanEval+ & 164 & 3 \\
    Data Selection--Code & Held-out evaluation & Codeforces & 467 & 3 \\
    Data Selection--Code & Held-out evaluation & CodeElo & 408 & 3 \\
    \midrule
    Math RFT tasks & Online checkpoint evaluation & MATH-S & 500 & 1 \\
    Math RFT tasks & In-loop Trial evaluation & MATH-S & 500 & 3 \\
    Math RFT tasks & Held-out evaluation & MATH-500 & 500 & 3 \\
    Math RFT tasks & Held-out evaluation & AIME 2024/2025, HMMT & 30 each & 32 \\
    \bottomrule
  \end{tabularx}%
  }
\end{table}

The Math RFT rows apply to both Reward Design--Math and Curriculum
Learning--Math.

\textbf{LiveCodeBench temporal split.}
Table~\ref{tab:lcb-temporal-split} gives the exact row counts and observed date
ranges on the two sides of the LiveCodeBench cutoff.

\begin{table}[!t]
  \centering\adeappendixtablestyle

  \caption{LiveCodeBench v5 temporal split used for code data selection.}
  \label{tab:lcb-temporal-split}
  \resizebox{0.85\linewidth}{!}{%
  \begin{tabularx}{0.85\linewidth}{@{}Y C{0.10\linewidth}C{0.20\linewidth}C{0.20\linewidth}@{}}
    \toprule
    Dataset & Rows & Start date & End date \\
    \midrule
    LiveCodeBench v5, in-loop split
      & 439 & 2023-05-07 & 2024-03-30 \\
    LiveCodeBench v5, held-out split
      & 441 & 2024-04-06 & 2025-01-04 \\
    \bottomrule
  \end{tabularx}%
  }
\end{table}

\textbf{MATH splits and evaluation boundaries.}
The dataset builder starts from the 7,500-example MATH training split.
It first samples the 500 problems in MATH-S to match the joint subject--difficulty
counts of MATH-500, using largest-remainder quotas and seeded sampling within
strata. It then samples 3,072 training problems from the remaining 7,000 rows.
In Data Selection--Math, MATH-S is used for final evaluation.

\subsection{Metrics, Graders, and Aggregation}

Let $c_{ij}\in\{0,1\}$ indicate whether sample $j$ for problem $i$ is judged
correct. For $N$ problems and $K$ samples per problem, we report
\begin{equation}
  \operatorname{Average@}K
  =\frac{1}{N}\sum_{i=1}^{N}\frac{1}{K}\sum_{j=1}^{K}c_{ij},
  \qquad
  \operatorname{Pass@}K
  =\frac{1}{N}\sum_{i=1}^{N}
    \mathbb{I}\!\left[\max_j c_{ij}=1\right].
  \label{eq:benchmark-metrics}
\end{equation}
Pass@$K$ measures the fraction of problems with at least one correct
response among the $K$ samples; Average@$K$ measures the mean fraction of
correct responses per problem. Both weight problems equally and are reported
as percentages.

\textbf{Aggregation across benchmarks.}
For each task, Macro Avg. is computed separately for Pass@$K$ and
Average@$K$ as the arithmetic mean over its four held-out benchmarks. Each
benchmark has equal weight regardless of its number of problems, and uses
its own $K$ from Table~\ref{tab:benchmark-purpose-spec}. We average unrounded
scores before formatting the results. Across-Run aggregation is specified in
Appendix~\ref{app:experimental-configuration}.

\textbf{Failed responses and evaluations.}
Missing or unextractable final answers, responses
without an extractable code block, code-execution failures, and per-sample
grader timeouts or exceptions receive zero credit and remain in the metric
denominator. A process-level failure that prevents the required evaluation from
finishing is instead retained as a missing evaluation and cannot improve a
best-so-far result.

\textbf{Benchmark graders.}
The AIME evaluator extracts the final boxed answer and applies the
fixed mathematical-answer comparison; HMMT uses MathArena answer extraction
and equivalence~checking. The MATH-S and
MATH-500 evaluators share Qwen Math answer extraction and
mathematical-equivalence grading, with a three-second grading timeout.
LiveCodeBench uses the last fenced code block, its public/private test
records and a fixed execution procedure with a
six-second test timeout and a 30-second outer execution timeout. HumanEval+
uses EvalPlus tests; Codeforces and CodeElo execute the generated Python
program against the benchmark's standard-input/standard-output tests.

Benchmark graders determine task accuracy; Judge and Review assessments
provide auxiliary evidence for strategy development and analysis.

\subsection{Generation and Checkpoint Selection}

\textbf{Generation.}
All evaluation purposes use stochastic decoding with temperature 0.6,
top-$p=0.95$, a maximum context length of 32,768 tokens, and at most 16,384 newly
generated tokens. Math and code SFT reserve up to 8,192 tokens for reasoning;
the RFT evaluator sets no separate reasoning-token limit.
Table~\ref{tab:benchmark-purpose-spec} gives the purpose-specific $K$.

The SFT system instruction requests step-by-step reasoning within explicit
reasoning tags, followed by a final answer in the problem's requested format.
RFT uses the model's default assistant system instruction; its mathematics
problem prompt requests a boxed final answer. The same task-specific prompt
format applies to the base target, baseline and proposed strategies.

\textbf{Selection and final evaluation.}
Every published checkpoint undergoes online checkpoint evaluation. Within
each task, checkpoints are ranked by the primary metric, with the secondary
metric breaking ties: SFT prioritizes Pass@$K$ then Average@$K$, whereas RFT
prioritizes Average@$K$ then Pass@$K$. The selected checkpoint undergoes
in-loop Trial evaluation on the same benchmark or split with the larger $K$
specified in Table~\ref{tab:benchmark-purpose-spec}. Those in-loop scores
support Trial and Plan ranking under the same metric ordering. Checkpoint
selection and strategy~selection both use adaptive research feedback.

Final evaluation applies the held-out suite to the selected checkpoint.

\section{Complete Effectiveness Results}
\label{app:complete-effectiveness}

This appendix complements the selected-strategy results in
Table~\ref{tab:effectiveness-results} with across-Run summaries, individual
Run results, and per-benchmark research-method comparisons. All score pairs
report Pass@$K$ / Average@$K$ in percentages. Benchmark-specific sampling
counts and grading rules are given in Appendix~\ref{app:benchmark-details}.

\subsection{Across-Run Means and Standard Deviations}
\label{app:effectiveness-summary}

Table~\ref{tab:average-results} summarizes each ADE task and capacity under
the reporting protocol in Appendix~\ref{app:reporting-protocol}.

Within each Run, Macro Avg. averages each metric separately over the four
held-out benchmarks, using the benchmark-specific $K$. We then compute the
mean and sample STD of these Run-level macro scores. Bold entries identify
per-metric best scores within each task, excluding STD rows.

\begin{table}[!t]
\centering
\caption{Across-Run means and sample STD (\%). Cells report Pass@$K$ / Average@$K$ for all tasks. Each ADE mean row is followed by its sample STD over three Runs; bold marks per-metric best scores, excluding STD rows.}
\label{tab:average-results}
\scriptsize\setlength{\tabcolsep}{3pt}
\renewcommand{\arraystretch}{1.05}
\resizebox{0.95\linewidth}{!}{%
\begin{tabular}{@{}lc|ccccc@{}}
\toprule
\textbf{Method} & \textbf{In-loop} & \multicolumn{5}{c}{\textbf{Held-out}} \\
\midrule
\textbf{Data Selection--Math} & \multicolumn{1}{c|}{AIME24} & \multicolumn{1}{c}{MATH-500} & \multicolumn{1}{c}{MATH-S} & \multicolumn{1}{c}{AIME25} & \multicolumn{1}{c}{HMMT} & \multicolumn{1}{c}{Macro Avg.} \\
\addlinespace[2pt]
Base model & 36.67 / 11.15 & 85.80 / 76.00 & 86.00 / 78.00 & 30.00 / 7.71 & 16.67 / 2.29 & 54.62 / 41.00 \\
Generic Baseline & 43.33 / 12.60 & 87.60 / \textbf{78.00} & 88.00 / 77.87 & 33.33 / 11.15 & 30.00 / \textbf{7.40} & 59.73 / 43.61 \\
ADE ($C=1$) & 51.11 / \textbf{16.46} & \textbf{87.87} / 77.98 & \textbf{88.60} / \textbf{78.71} & 40.00 / \textbf{13.85} & \textbf{31.11} / 6.18 & \textbf{61.89} / \textbf{44.18} \\
\quad STD & 5.09 / 1.17 & 2.08 / 1.82 & 0.53 / 0.92 & 6.67 / 2.76 & 5.09 / 0.47 & 3.22 / 1.25 \\
ADE ($C=3$) & \textbf{54.44} / 16.01 & 87.40 / 77.00 & 87.53 / 77.24 & \textbf{41.11} / 12.57 & \textbf{31.11} / 5.87 & 61.79 / 43.17 \\
\quad STD & 1.93 / 1.34 & 1.51 / 1.46 & 0.31 / 1.69 & 6.94 / 3.19 & 5.09 / 1.20 & 2.27 / 1.53 \\
\midrule
\textbf{Data Selection--Code} & \multicolumn{1}{c|}{LCB $\leq$03/30} & \multicolumn{1}{c}{LCB $>$03/30} & \multicolumn{1}{c}{HumanEval+} & \multicolumn{1}{c}{Codeforces} & \multicolumn{1}{c}{CodeElo} & \multicolumn{1}{c}{Macro Avg.} \\
\addlinespace[2pt]
Base model & 40.55 / 27.49 & 24.26 / 15.95 & 85.98 / \textbf{74.39} & 7.28 / 3.93 & 8.58 / 4.25 & 31.53 / 24.63 \\
Generic Baseline & 39.86 / 26.35 & 25.85 / 18.44 & 85.98 / 70.53 & 8.57 / 4.85 & 9.07 / 5.23 & 32.37 / 24.76 \\
ADE ($C=1$) & 43.96 / 30.37 & 28.95 / 19.65 & 84.76 / 71.41 & 7.92 / 4.38 & 8.33 / 4.49 & 32.49 / 24.98 \\
\quad STD & 1.64 / 1.54 & 1.37 / 0.42 & 2.44 / 1.84 & 0.77 / 0.54 & 1.30 / 0.59 & 1.30 / 0.66 \\
ADE ($C=3$) & \textbf{45.79} / \textbf{32.35} & \textbf{31.59} / \textbf{22.32} & \textbf{88.21} / 73.92 & \textbf{8.92} / \textbf{5.23} & \textbf{10.54} / \textbf{5.86} & \textbf{34.82} / \textbf{26.83} \\
\quad STD & 2.40 / 1.74 & 0.57 / 0.27 & 0.70 / 1.89 & 0.33 / 0.04 & 0.98 / 0.60 & 0.28 / 0.37 \\
\midrule
\textbf{Reward Design--Math} & \multicolumn{1}{c|}{MATH-S} & \multicolumn{1}{c}{MATH-500} & \multicolumn{1}{c}{AIME24} & \multicolumn{1}{c}{AIME25} & \multicolumn{1}{c}{HMMT} & \multicolumn{1}{c}{Macro Avg.} \\
\addlinespace[2pt]
Base model & 39.40 / 20.87 & 36.60 / 19.93 & 0.00 / 0.00 & 3.33 / 0.10 & 0.00 / 0.00 & 9.98 / 5.01 \\
Generic Baseline & 49.00 / 34.13 & \textbf{46.80} / \textbf{30.13} & 3.33 / 0.21 & 0.00 / 0.00 & 0.00 / 0.00 & 12.53 / 7.59 \\
ADE ($C=1$) & 51.07 / 35.76 & 45.60 / 30.00 & 7.78 / 0.35 & 3.33 / 0.10 & \textbf{1.11} / \textbf{0.03} & 14.46 / \textbf{7.62} \\
\quad STD & 1.33 / 0.89 & 0.92 / 1.22 & 5.09 / 0.22 & 3.34 / 0.11 & 1.92 / 0.06 & 0.49 / 0.30 \\
ADE ($C=3$) & \textbf{52.20} / \textbf{36.18} & 44.07 / 29.31 & \textbf{11.11} / \textbf{0.56} & \textbf{4.44} / \textbf{0.14} & 0.00 / 0.00 & \textbf{14.91} / 7.50 \\
\quad STD & 0.72 / 0.45 & 2.47 / 1.79 & 3.85 / 0.26 & 1.93 / 0.06 & 0.00 / 0.00 & 0.66 / 0.39 \\
\midrule
\textbf{Curriculum Learning--Math} & \multicolumn{1}{c|}{MATH-S} & \multicolumn{1}{c}{MATH-500} & \multicolumn{1}{c}{AIME24} & \multicolumn{1}{c}{AIME25} & \multicolumn{1}{c}{HMMT} & \multicolumn{1}{c}{Macro Avg.} \\
\addlinespace[1pt]
Base model & 39.40 / 20.87 & 36.60 / 19.93 & 0.00 / 0.00 & \textbf{3.33} / \textbf{0.10} & 0.00 / 0.00 & 9.98 / 5.01 \\
Generic Baseline & 49.00 / 34.13 & \textbf{46.80} / 30.13 & 3.33 / 0.21 & 0.00 / 0.00 & 0.00 / 0.00 & 12.53 / 7.59 \\
ADE ($C=1$) & 51.20 / 35.76 & 46.60 / \textbf{30.47} & 7.78 / 0.38 & 0.00 / 0.00 & \textbf{2.22} / 0.07 & 14.15 / \textbf{7.73} \\
\quad STD & 0.53 / 0.90 & 0.72 / 0.42 & 1.92 / 0.22 & 0.00 / 0.00 & 1.92 / 0.06 & 0.90 / 0.17 \\
ADE ($C=3$) & \textbf{53.20} / \textbf{37.07} & 45.73 / 30.29 & \textbf{10.00} / \textbf{0.42} & 0.00 / 0.00 & \textbf{2.22} / \textbf{0.14} & \textbf{14.49} / 7.71 \\
\quad STD & 1.06 / 0.07 & 1.67 / 0.67 & 3.33 / 0.00 & 0.00 / 0.00 & 1.92 / 0.16 & 0.74 / 0.17 \\
\bottomrule
\end{tabular}}
\end{table}

\subsection{Run-Level Results}
\label{app:run-results}
Tables~\ref{tab:runs-math-sft}--\ref{tab:runs-curriculum} report the individual
results underlying Table~\ref{tab:average-results}. Each ADE row uses one
Run's in-loop-selected checkpoint for all benchmark scores. Run numbers are
local to each task and capacity. The marker $\ast$ identifies the Run selected
for the corresponding Best result in Table~\ref{tab:effectiveness-results}.

\begin{table}[!t]
\centering\scriptsize\setlength{\tabcolsep}{3pt}
\caption{Data Selection--Math run-level results (\%). $\ast$ marks the Best Run within each \mbox{ADE setting}.}
\label{tab:runs-math-sft}
\resizebox{0.95\linewidth}{!}{%
\begin{tabular}{@{}lcc|ccccc@{}}
\toprule
Method & Run & In-loop & \multicolumn{5}{c}{Held-out} \\
\midrule
 & & AIME24 & MATH-500 & MATH-S & AIME25 & HMMT & Macro Avg. \\
Base model & 1 & 36.67 / 11.15 & 85.80 / 76.00 & 86.00 / 78.00 & 30.00 / 7.71 & 16.67 / 2.29 & 54.62 / 41.00 \\
Generic Baseline & 1 & 43.33 / 12.60 & 87.60 / 78.00 & 88.00 / 77.87 & 33.33 / 11.15 & 30.00 / 7.40 & 59.73 / 43.61 \\
ADE ($C=1$) & 1 & 50.00 / 15.83 & 87.20 / 77.47 & 88.00 / 77.93 & 40.00 / 15.94 & 26.67 / 6.15 & 60.47 / 44.37 \\
ADE ($C=1$) & 2$^{\ast}$ & 56.67 / 17.81 & 90.20 / 80.00 & 88.80 / 79.73 & 46.67 / 14.90 & 36.67 / 6.67 & 65.58 / 45.32 \\
ADE ($C=1$) & 3 & 46.67 / 15.73 & 86.20 / 76.47 & 89.00 / 78.47 & 33.33 / 10.73 & 30.00 / 5.73 & 59.63 / 42.85 \\
ADE ($C=3$) & 1 & 53.33 / 14.48 & 86.00 / 75.40 & 87.80 / 75.40 & 33.33 / 10.73 & 30.00 / 5.83 & 59.28 / 41.84 \\
ADE ($C=3$) & 2 & 53.33 / 16.56 & 89.00 / 78.27 & 87.20 / 77.60 & 46.67 / 10.73 & 26.67 / 4.69 & 62.39 / 42.82 \\
ADE ($C=3$) & 3$^{\ast}$ & 56.67 / 16.98 & 87.20 / 77.33 & 87.60 / 78.73 & 43.33 / 16.25 & 36.67 / 7.08 & 63.70 / 44.85 \\
\bottomrule
\end{tabular}}
\end{table}
\begin{table}[!t]
\centering\scriptsize\setlength{\tabcolsep}{3pt}
\caption{Data Selection--Code run-level results (\%). $\ast$ marks the Best Run within each \mbox{ADE setting}.}
\label{tab:runs-code-sft}
\resizebox{0.95\linewidth}{!}{%
\begin{tabular}{@{}lcc|ccccc@{}}
\toprule
Method & Run & In-loop & \multicolumn{5}{c}{Held-out} \\
\midrule
 & & LCB $\leq$03/30 & LCB $>$03/30 & HumanEval+ & Codeforces & CodeElo & Macro Avg. \\
Base model & 1 & 40.55 / 27.49 & 24.26 / 15.95 & 85.98 / 74.39 & 7.28 / 3.93 & 8.58 / 4.25 & 31.53 / 24.63 \\
Generic Baseline & 1 & 39.86 / 26.35 & 25.85 / 18.44 & 85.98 / 70.53 & 8.57 / 4.85 & 9.07 / 5.23 & 32.37 / 24.76 \\
ADE ($C=1$) & 1$^{\ast}$ & 45.79 / 31.74 & 30.39 / 20.11 & 87.20 / 72.76 & 7.71 / 4.00 & 8.82 / 4.66 & 33.53 / 25.38 \\
ADE ($C=1$) & 2 & 42.60 / 28.70 & 28.80 / 19.27 & 84.76 / 72.15 & 8.78 / 5.00 & 9.31 / 4.98 & 32.91 / 25.35 \\
ADE ($C=1$) & 3 & 43.51 / 30.68 & 27.66 / 19.58 & 82.32 / 69.31 & 7.28 / 4.14 & 6.86 / 3.84 & 31.03 / 24.22 \\
ADE ($C=3$) & 1 & 43.28 / 30.37 & 31.52 / 22.60 & 87.80 / 72.36 & 9.21 / 5.21 & 9.56 / 5.64 & 34.52 / 26.45 \\
ADE ($C=3$) & 2 & 46.01 / 33.64 & 31.07 / 22.30 & 87.80 / 73.37 & 8.99 / 5.21 & 11.52 / 6.54 & 34.85 / 26.86 \\
ADE ($C=3$) & 3$^{\ast}$ & 48.06 / 33.03 & 32.20 / 22.07 & 89.02 / 76.02 & 8.57 / 5.28 & 10.54 / 5.39 & 35.08 / 27.19 \\
\bottomrule
\end{tabular}}
\end{table}
\begin{table}[!t]
\centering\scriptsize\setlength{\tabcolsep}{3pt}
\caption{Reward Design--Math run-level results (\%). $\ast$ marks the Best Run within each \mbox{ADE setting}.}
\label{tab:runs-math-rft}
\resizebox{0.95\linewidth}{!}{%
\begin{tabular}{@{}lcc|ccccc@{}}
\toprule
Method & Run & In-loop & \multicolumn{5}{c}{Held-out} \\
\midrule
 & & MATH-S & MATH-500 & AIME24 & AIME25 & HMMT & Macro Avg. \\
Base model & 1 & 39.40 / 20.87 & 36.60 / 19.93 & 0.00 / 0.00 & 3.33 / 0.10 & 0.00 / 0.00 & 9.98 / 5.01 \\
Generic Baseline & 1 & 49.00 / 34.13 & 46.80 / 30.13 & 3.33 / 0.21 & 0.00 / 0.00 & 0.00 / 0.00 & 12.53 / 7.59 \\
ADE ($C=1$) & 1 & 50.40 / 34.73 & 44.60 / 28.60 & 13.33 / 0.52 & 0.00 / 0.00 & 0.00 / 0.00 & 14.48 / 7.28 \\
ADE ($C=1$) & 2$^{\ast}$ & 50.20 / 36.33 & 46.40 / 30.80 & 3.33 / 0.10 & 6.67 / 0.21 & 3.33 / 0.10 & 14.93 / 7.80 \\
ADE ($C=1$) & 3 & 52.60 / 36.20 & 45.80 / 30.60 & 6.67 / 0.42 & 3.33 / 0.10 & 0.00 / 0.00 & 13.95 / 7.78 \\
ADE ($C=3$) & 1 & 52.80 / 36.07 & 46.80 / 31.33 & 6.67 / 0.31 & 3.33 / 0.10 & 0.00 / 0.00 & 14.20 / 7.94 \\
ADE ($C=3$) & 2$^{\ast}$ & 52.40 / 36.67 & 42.00 / 27.93 & 13.33 / 0.52 & 6.67 / 0.21 & 0.00 / 0.00 & 15.50 / 7.17 \\
ADE ($C=3$) & 3 & 51.40 / 35.80 & 43.40 / 28.67 & 13.33 / 0.83 & 3.33 / 0.10 & 0.00 / 0.00 & 15.02 / 7.40 \\
\bottomrule
\end{tabular}}
\end{table}

\begin{table}[!t]
\centering\scriptsize\setlength{\tabcolsep}{3pt}
\caption{Curriculum Learning--Math run-level results (\%). $\ast$ marks the Best Run within each \mbox{ADE setting}.}
\label{tab:runs-curriculum}
\resizebox{0.95\linewidth}{!}{%
\begin{tabular}{@{}lcc|ccccc@{}}
\toprule
Method & Run & In-loop & \multicolumn{5}{c}{Held-out} \\
\midrule
 & & MATH-S & MATH-500 & AIME24 & AIME25 & HMMT & Macro Avg. \\
Base model & 1 & 39.40 / 20.87 & 36.60 / 19.93 & 0.00 / 0.00 & 3.33 / 0.10 & 0.00 / 0.00 & 9.98 / 5.01 \\
Generic Baseline & 1 & 49.00 / 34.13 & 46.80 / 30.13 & 3.33 / 0.21 & 0.00 / 0.00 & 0.00 / 0.00 & 12.53 / 7.59 \\
ADE ($C=1$) & 1 & 51.00 / 36.13 & 47.20 / 30.80 & 10.00 / 0.62 & 0.00 / 0.00 & 3.33 / 0.10 & 15.13 / 7.88 \\
ADE ($C=1$) & 2 & 50.80 / 34.73 & 46.80 / 30.00 & 6.67 / 0.21 & 0.00 / 0.00 & 0.00 / 0.00 & 13.37 / 7.55 \\
ADE ($C=1$) & 3$^{\ast}$ & 51.80 / 36.40 & 45.80 / 30.60 & 6.67 / 0.31 & 0.00 / 0.00 & 3.33 / 0.10 & 13.95 / 7.75 \\
ADE ($C=3$) & 1 & 52.40 / 37.00 & 44.40 / 29.93 & 13.33 / 0.42 & 0.00 / 0.00 & 3.33 / 0.31 & 15.27 / 7.67 \\
ADE ($C=3$) & 2$^{\ast}$ & 54.40 / 37.13 & 47.60 / 31.07 & 6.67 / 0.42 & 0.00 / 0.00 & 3.33 / 0.10 & 14.40 / 7.90 \\
ADE ($C=3$) & 3 & 52.80 / 37.07 & 45.20 / 29.87 & 10.00 / 0.42 & 0.00 / 0.00 & 0.00 / 0.00 & 13.80 / 7.57 \\
\bottomrule
\end{tabular}}
\end{table}

\subsection{Research Organization and Feedback Comparisons (Math Data Selection)}
\label{app:math-sft-ablation}

Table~\ref{tab:math-sft-ablation-full} gives the per-benchmark scores behind
Figure~\ref{fig:math-sft-ablation}. Results follow the Best-selection protocol
in Appendix~\ref{app:reporting-protocol}.
The ADE row uses $C=1$. Bold marks per-metric maxima among the compared~methods.

\begin{table}[!t]
\centering\scriptsize
\caption{Data Selection--Math comparison details (\%). Cells report Pass@$K$ / Average@$K$; $K=32$ on AIME24, AIME25, and HMMT, and $K=3$ on MATH-500 and MATH-S.}
\label{tab:math-sft-ablation-full}
\setlength{\tabcolsep}{4pt}
\resizebox{0.95\linewidth}{!}{%
\begin{tabular}{@{}lcccccc@{}}
\toprule
Method & AIME24 & MATH-500 & MATH-S & AIME25 & HMMT & Macro Avg. \\
\midrule
Unified Agent & 50.00 / 14.38 & 87.00 / 76.33 & 88.20 / 76.33 & 43.33 / 12.81 & 33.33 / 5.52 & 62.97 / 42.75 \\
Independent Search & 50.00 / 15.52 & 89.20 / 78.60 & \textbf{88.80} / 77.13 & 36.67 / 13.33 & \textbf{36.67} / \textbf{7.29} & 62.83 / 44.09 \\
Outcome-only Feedback & 46.67 / 14.17 & 85.80 / 76.33 & 88.00 / 76.93 & 40.00 / 14.58 & 26.67 / 6.35 & 60.12 / 43.55 \\
Raw Evidence Feedback & 50.00 / 15.10 & 88.00 / 78.00 & 87.60 / 77.47 & 36.67 / 11.77 & 30.00 / 5.52 & 60.57 / 43.19 \\
ADE ($C=1$) & \textbf{56.67} / \textbf{17.81} & \textbf{90.20} / \textbf{80.00} & \textbf{88.80} / \textbf{79.73} & \textbf{46.67} / \textbf{14.90} & \textbf{36.67} / 6.67 & \textbf{65.58} / \textbf{45.32} \\
\bottomrule
\end{tabular}}
\end{table}

\section{Model Cost and Research Time}
\label{app:scaling-accounting}
\label{app:model-cost}

This appendix reports model-call costs and the distribution of research time
for the two data-selection tasks and reward design. Curriculum learning is
outside this analysis. Hardware allocations and research budgets are specified
in Appendix~\ref{app:experimental-configuration}.

\textbf{Token usage and cost estimation.}
Table~\ref{tab:model-cost-summary} reports equal-weight Run means, with the
number of Runs shown for each task and capacity. GPT columns cover research-agent
execution using GPT-5.6 Sol; Qwen columns cover Analyzer-requested record-level
review using Qwen3.6-35B-A3B. The estimates use the five-Plan-per-Coordinator
budget and accompany the cost curves in Figure~\ref{fig:rq2}.

Reported token volumes sum uncached input and output tokens; output includes
reasoning tokens. For each model, API-equivalent cost is
\begin{equation}
  \mathrm{Cost}=10^{-6}\bigl(T_u r_u+T_c r_c+T_o r_o\bigr),
\end{equation}
where $T_u$, $T_c$, and $T_o$ count uncached input, observed cached input,
and output tokens. The corresponding rates $r_u$, $r_c$, and $r_o$ appear in
Table~\ref{tab:model-api-rates}. Qwen cache usage and its cached-input rate
are unavailable; its estimate applies the input rate to reported input tokens.
The self-hosted Qwen service is valued at the China (Beijing) API rate.
These estimates cover model calls; Figure~\ref{fig:rq2} reports GPU resources.

\begin{table}[!t]
\centering\adeappendixtablestyle
\setlength{\tabcolsep}{3pt}
\caption{Estimated mean token usage and API-equivalent cost per Run. Token counts are in millions (M); costs are in USD. Total sums the two model-cost columns before rounding.}
\label{tab:model-cost-summary}
\resizebox{0.95\linewidth}{!}{%
\begin{tabularx}{0.95\linewidth}{@{}L{0.22\linewidth}ZZZZZZZ@{}}
\toprule
Task & $C$ & Runs & GPT (M) & Qwen (M) & GPT (\$) & Qwen (\$) & Total (\$) \\
\midrule
Data Selection--Math & 1 & 3 & 4.67 & 34.32 & 49.58 & 12.55 & 62.13 \\
Data Selection--Math & 3 & 3 & 15.58 & 106.46 & 175.64 & 39.91 & 215.56 \\
Data Selection--Code & 1 & 3 & 5.07 & 36.51 & 57.09 & 14.92 & 72.01 \\
Data Selection--Code & 3 & 3 & 15.19 & 131.59 & 181.41 & 58.49 & 239.90 \\
Reward Design--Math & 1 & 2 & 5.71 & 43.39 & 70.57 & 18.57 & 89.13 \\
Reward Design--Math & 3 & 2 & 16.16 & 133.87 & 204.91 & 60.29 & 265.20 \\
\bottomrule
\end{tabularx}%
}
\end{table}
\begin{table}[!t]
\centering\adeappendixtablestyle
\setlength{\tabcolsep}{12pt}
\caption{Fixed API rates used for cost estimation, dated September 9, 2026 (USD per \mbox{million tokens}).}
\label{tab:model-api-rates}
\resizebox{0.65\linewidth}{!}{%
\begin{tabular}{@{}lccc@{}}
\toprule
Model & Input & Cached input & Output \\
\midrule
GPT-5.6 Sol & 4.00 & 0.40 & 20.00 \\
Qwen3.6-35B-A3B & 0.248 & -- & 1.485 \\
\bottomrule
\end{tabular}%
}
\end{table}

\subsection{Research-Stage Timing}
\label{app:stage-timing-c3}
\begin{figure}[!t]
  \centering
  \includegraphics[width=0.97\linewidth]{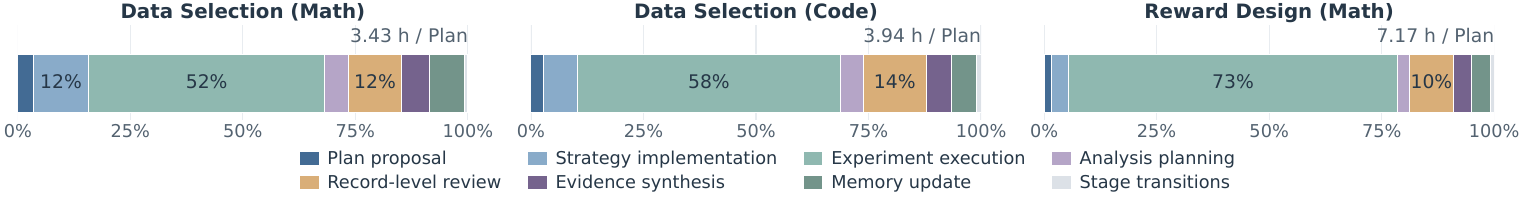}
  \caption{Estimated research-stage times at $C=3$. Bars show time shares;
  labels above each bar give mean hours per Plan, including waiting.}
  \label{fig:plan-stage-time-c3}
\end{figure}

Figure~\ref{fig:plan-stage-time-c3} gives the $C=3$ counterpart to the $C=1$
breakdown in Figure~\ref{fig:plan-stage-time}. Both estimate the full sequence
from Plan proposal through Memory update over five Plan slots per Coordinator.
Stage durations are averaged within each Run and then equally across Runs.
The Memory update stage combines Plan and Run summarization.

Durations include waiting. Each stage's share is its mean duration divided
by the mean total Plan duration. Because Plans can overlap, these per-Plan
times are not the elapsed wall-clock time of the entire Run. Experiment
execution accounts for 52--73\% of estimated Plan time at $C=3$.

\section{Case Study: Feedback-Guided Code Data Selection}
\label{app:evidence-cases}

\subsection{Performance and Experimental Setting}

This case expands the research trajectory in Section~\ref{sec:rq4} and
Figure~\ref{fig:rq4-code-sft-trace}. It follows three Coordinators from their
initial proposals through feedback-guided revisions and combinations to the
selected strategy, \emph{Stratified Repetition Control} (C1/P3), and its
subsequent reuse. The focus is how findings from different branches change
what agents propose next.

We examine the in-loop-selected best Data Selection--Code Run at $C=3$.
Its selected strategy achieves \textbf{48.06\% Pass@3 and 33.03\% Average@3}
on the in-loop LiveCodeBench split and \textbf{32.20\% Pass@3 and 22.07\%
Average@3} on the temporally held-out split. Relative to the generic SFT
baseline, the absolute gains are 8.20\% / 6.68\% in-loop and 6.35\% / 3.63\%
held-out. Table~\ref{tab:effectiveness-results} reports the broader benchmark
comparison. Unless otherwise stated, scores below refer to \mbox{in-loop evaluation}.

The Run contains 15 Plans: three Coordinators each develop five Plans with
one Trial per Plan. Every strategy selects 384 complete trajectories from
the same candidate pool, preserving their contents and the training and
evaluation procedures specified in
Appendices~\ref{app:benchmark-details} and~\ref{app:experimental-configuration}.

\subsection{Strategy Ancestry, Evidence, and Selection}

The trace distinguishes \textbf{strategy inheritance} from \textbf{knowledge
transfer}. R/C labels in Figure~\ref{fig:rq4-code-sft-trace} and the parent
column in Table~\ref{tab:code-sft-case-plans} identify the strategies a Plan
revisits or combines. K1--K5 identify findings that inform its proposal.
A Plan can inherit one strategy while drawing on observations from several
other experiments; a new direction can use prior findings without inheriting
an artifact.

C1--C3 denote Coordinators, and P1--P5 denote each Coordinator's local Plan
order. The figure's numbered entries indicate global proposal order, not
completion order or synchronized rounds. The discussion groups related
branches; evidence from a completed experiment is distinguished from knowledge
of another Coordinator's still-active proposal. All formal parents had
completed before their strategies were reused.

The selected strategy is determined by in-loop scores. The discussion
follows how observations and scores guide subsequent proposals.

\subsection{Initial Exploration}

The initial proposals address three distinct potential weaknesses in the baseline
selection. All begin from the same fixed pool and evaluation setting, but they
operationalize different \mbox{training-data properties}.

\textbf{C1/P1 --- Complete targets.} C1 first asks whether training on cutoff-exceeding responses is a source of avoidable supervision loss. The baseline contains 26 selected records marked to exceed the training cutoff; its evaluation also contains 197 length-terminated generations without extracted executable code. The proposal replaces cutoff-exceeding baseline records with unused, non-truncating candidates of approximately matched length, preserving the 384-record budget and the baseline's 142/107/135 short/medium/long allocation. It reaches 42.14\% Pass@3 and 29.76\% Average@3. Its immediate successor is C1/P2, but its complete-target idea also contributes to C3/P3. The complete-target direction thus contributes to later work across search lanes.

\textbf{C2/P1 --- Code-oriented allocation.} C2 instead asks whether a mixed-domain candidate pool should be selected using observable alignment with the downstream coding task. It prioritizes prompts explicitly requesting executable code through a named interface or a standard-input/standard-output contract. It uses length and cutoff strata as matching controls rather than relying on noisy semantic domain labels. The resulting subset reaches 44.87\% Pass@3 and 32.50\% Average@3, the strongest of the three initial directions. Inspection characterizes 368 of its 384 prompts as executable-Python-wrapper requests. This stronger subset becomes the direct parent of C2/P2 and C1/P3 and contributes to three combination Plans: C3/P2, C3/P3, and C2/P4.

\textbf{C3/P1 --- Executable-target integrity.} C3 investigates whether definite defects in machine-checkable programming targets can be removed using extraction, parsing/compilation, requested-interface checks, and runnable prompt examples. It proposes replacing definite failures with matched passes while retaining uncheckable and non-programming records. The result is 42.37\% Pass@3 and 28.63\% Average@3. Its interface checks give C3/P2 a concrete component to combine with C2/P1's stronger code-oriented subset.

\subsection{Feedback, Refinement, and Recombination}

The next proposals test explanations suggested by the initial results and combine
components from different branches. Two unsuccessful interventions---shortening and
Judge-based filtering---later help shape C1/P3.

\textbf{C2/P2 --- Distinguishing long reasoning from unproductive reasoning.} The code-oriented parent contains substantial long-form supervision, and its failed evaluation generations are longer than successful ones. C2 tests a moderate-length preference by replacing non-truncating long records with medium-length records of the same prompt-contract subtype. The realized intervention moves 43 records from the long band to the medium band, but both metrics decline, to 42.82\% Pass@3 and 31.21\% Average@3. This decline later motivates C1/P3 to match length while targeting repetition. The subsequent search therefore separates reasoning length from repetition within the reasoning.

\textbf{C3/P2 --- Combining task alignment with callable integrity.} C3 explicitly combines C2/P1's code-oriented source with C3/P1's executable-integrity direction. Inspection of the stronger source found parseable Python blocks that nevertheless omitted a requested top-level function. The new proposal therefore targets exact requested-callable presence, rather than treating successful parsing as sufficient compliance. It obtains 42.14\% Pass@3 and 30.52\% Average@3. The broader definition-and-invocation defects identified during this experiment are later cited by C1/P4 when formulating a more specific wrapper-closure experiment.

\textbf{C1/P2 --- Testing semantic validity as a selection signal.} The task Judge scores candidate question--response pairs against a rubric, separately from benchmark grading. Building on C1/P1, C1 uses this Judge to identify substantively invalid targets and replace them with same-family valid candidates. Reviews of the complete-target subset had flagged major errors, including an association between severe meandering and correctness problems. The new result slightly increases Pass@3 to 42.60\%, but lowers Average@3 to 28.78\%, violating the proposal's secondary-metric condition. The candidate judgments contain thousands of passes but no fails, providing no failing group for the proposed replacement rule. These observations later influence C1/P3's choice not to use Judge output as its selection criterion.

\textbf{C2/P3 and C2/P4 --- Separating an exploratory direction from its combination.} After the unsuccessful length refinement, C2/P3 starts a formally new direction: prioritize programming targets qualified by execution and prompt-example checks, then spread them across prompt-family/length cells. Its proposal consults the earlier executable-integrity findings and the failed moderate-length experiment, but does not declare either as an artifact parent. It reaches 44.42\% Pass@3 and 30.07\% Average@3, below C2/P1. C2/P4 then explicitly combines the code-oriented allocation of C2/P1 with the execution qualification and family coverage of C2/P3. This combination reaches 46.70\% Pass@3 and 32.12\% Average@3, compared with C2/P1's 44.87\% and 32.50\%. It extends problem coverage while largely retaining the parent's average correctness.

\textbf{C3/P3 and C3/P4 --- Reusing completeness, then testing structural complexity.} C3/P3 combines C2/P1's code-oriented allocation with C1/P1's complete-target eligibility. It reaches 46.47\% Pass@3 and 31.89\% Average@3, becoming the primary-metric leader. C3/P4, proposed before C1/P3 has completed, starts from the leader available at that time. Weak hard-problem performance motivates a bounded preference for greater branch complexity in the Python abstract syntax tree (AST) among otherwise matched execution-qualified targets. It reaches 45.79\% Pass@3 and 32.27\% Average@3. C3/P4 therefore uses the strongest completed source available at proposal time.

\subsection{Cross-Coordinator Findings and Stratified Repetition Control}
\label{app:case-knowledge-transfer}

C1/P3 adopts C2/P1 as its formal parent and proposes \emph{Stratified
Repetition Control}. Three findings, marked K1--K3 in
Figure~\ref{fig:rq4-code-sft-trace}, explain the choice of intervention:
\begin{itemize}[leftmargin=*,itemsep=0.2em]
\item \textbf{K1: Target repetition.} Reviews of C2/P1 and the initial C1/C3
subsets identify redundancy or meandering as plausible supervision defects.
\item \textbf{K2: Control length exposure.} C2/P2's unsuccessful shortening
motivates matching length while changing repetition.
\item \textbf{K3: Use a deterministic signal.} C1/P2's uninformative Judge gate
motivates an observable repetition measure instead of another semantic threshold.
\end{itemize}
The proposal also takes account of active Plans already exploring execution
qualification and code-oriented complete targets. Their proposed directions
help C1/P3 focus on repetition while other branches explore complementary
properties of the training data.

The strategy replaces highly repetitive parent records with low-repetition
candidates matched jointly by prompt-contract subtype, truncation status,
and length band, then by nearest token count within each stratum.
Appendix~\ref{app:artifact-code-sft} gives the executed
selection logic and thresholds.

The selected subset reaches 48.06\% Pass@3 and 33.03\% Average@3, improving
on C2/P1 by absolute gains of 3.19\% and 0.53\%. Its short/medium/long counts
remain 142/107/135, and its selected-token total is 2,662,517 compared with
2,662,461 for the parent. These aggregate counts support approximate exposure
matching: the search retains the parent's broad length allocation while using
repetition to distinguish candidates within the matching strata.

\subsection{Follow-up Experiments from the Selected Strategy}

Once C1/P3 completes, all three Coordinators propose further refinements from it.
Figure~\ref{fig:rq4-code-sft-trace} highlights these four direct revisits in blue
alongside C1/P3, showing reuse of the same artifact across all three search lanes. The
planning records also show agents using the visible allocation of other proposals to
avoid testing the same change.

\textbf{C1/P4 --- Definition-and-invocation closure.} Inspection of the winning subset finds that all 368 final Python blocks parse, but only 280 define a function and 279 invoke a defined function despite the prompt requirements. Together with the unresolved callable experiment in C3/P2 (K4), this motivates a narrower static wrapper-closure proposal, checking that the requested function is both defined and called: replace definite definition/call mismatches with matched wrapper-closed targets. It reaches 43.74\% Pass@3 and 32.95\% Average@3.

\textbf{C3/P5 --- Correctness on prompt-supplied examples.} C3 also revisits C1/P3, but targets the 804 wrong-answer outcomes that dominate the leader's residual failures. Because C1/P4 is already testing wrapper closure, C3/P5 leaves static wrapper mismatches untouched and instead tests available, unambiguous prompt examples within wrapper-closed records. It proposes matched replacements of example-failing targets with example-passing targets. It reaches 44.65\% Pass@3 and 30.90\% Average@3, below the leader. The proposal illustrates both reuse of a shared artifact and an explicit division of experimental scope.

\textbf{C2/P5 --- Earlier executable-answer onset.} C2 observes that the selected leader still has 25 generation-limit failures without extractable code, compared with 15 in C2/P1. Its proposal explicitly leaves wrapper mismatches to C1/P4 and prompt-example correctness to the active C3/P5. Instead, it tests whether final Python appearing late in the target can be replaced by otherwise matched records with earlier code onset, restricting changes to records without a complete runnable prompt example. This produces 45.10\% Pass@3 and 31.97\% Average@3. The proposal uses the visible division of work to explore a complementary failure mode.

\textbf{C1/P5 --- Removing literal image placeholders.} The final C1 proposal is informed by inspection of the completed C1/P4 subset (K5), which contains 22 prompts with a literal \texttt{<image>} marker. C1/P5 returns to C1/P3 as its formal parent and proposes replacing marker-bearing records with matched marker-free candidates, preserving the contents of each selected trajectory. It reaches 45.56\% Pass@3 and 31.66\% Average@3. This separates diagnostic evidence from artifact inheritance: a later experiment supplies a new diagnostic observation, while an earlier, better artifact remains the starting point.

\subsection{Summary of the Research Trajectory}

Table~\ref{tab:code-sft-case-plans} summarizes all 15 Plans. Scores are in-loop
percentages. Parentage records artifact-level revisit/combine relations; the preceding
discussion identifies additional diagnostic and coordination influences. A dash denotes
a formally new direction, not absence of access to prior evidence.

\begin{table}[!t]
\centering\adeappendixtablestyle
\setlength{\tabcolsep}{3pt}
\caption{All 15 Plans in the Data Selection--Code case-study Run. Scores are in-loop percentages. A dash denotes a new direction with no formal Plan parent. Bold scores mark the in-loop-selected~strategy.}
\label{tab:code-sft-case-plans}
\resizebox{0.95\linewidth}{!}{%
\begin{tabularx}{0.95\linewidth}{@{}lYccc@{}}
\toprule
Plan & Strategy & Formal parent(s) & Pass@3 & Average@3 \\
\midrule
C1/P1 & Complete targets & --- & 42.14 & 29.76 \\
C1/P2 & Judge validity & C1/P1 & 42.60 & 28.78 \\
C1/P3 & Stratified Repetition Control & C2/P1 & \textbf{48.06} & \textbf{33.03} \\
C1/P4 & Wrapper closure & C1/P3 & 43.74 & 32.95 \\
C1/P5 & Literal image-marker removal & C1/P3 & 45.56 & 31.66 \\
\addlinespace[3pt]
C2/P1 & Code-contract priority & --- & 44.87 & 32.50 \\
C2/P2 & Moderate-length replacement & C2/P1 & 42.82 & 31.21 \\
C2/P3 & Execution qualification and spread & --- & 44.42 & 30.07 \\
C2/P4 & Code-oriented execution mix & C2/P1 + C2/P3 & 46.70 & 32.12 \\
C2/P5 & Earlier code onset & C1/P3 & 45.10 & 31.97 \\
\addlinespace[3pt]
C3/P1 & Executable-target integrity & --- & 42.37 & 28.63 \\
C3/P2 & Requested callable & C2/P1 + C3/P1 & 42.14 & 30.52 \\
C3/P3 & Code-oriented complete targets & C2/P1 + C1/P1 & 46.47 & 31.89 \\
C3/P4 & AST branch complexity & C3/P3 & 45.79 & 32.27 \\
C3/P5 & Prompt-example correctness & C1/P3 & 44.65 & 30.90 \\
\bottomrule
\end{tabularx}%
}
\end{table}

\subsection{Analysis of Feedback and Strategy Reuse}

\textbf{Unsuccessful experiments sharpen the next question.}
C2/P2 lowers both metrics after shifting long records toward medium-length
ones. C1/P2 produces a Judge signal with no failing candidate group and
reduces Average@3. C1/P3 brings these observations together: retain the
length allocation of the stronger code-oriented parent, but use a
deterministic repetition measure to select among comparable candidates.
The progression is from broad concerns about long or invalid targets to a
more specific selection rule informed by what earlier attempts revealed.

\textbf{Useful findings travel beyond strategy ancestry.}
C1/P3 inherits C2/P1's subset while drawing on repetition diagnostics and
failed refinements from other branches. C1/P5 later uses the image-marker
observation from C1/P4 while returning to C1/P3 as its parent. In both
cases, the artifact chosen as a starting point and the experiment supplying
the next research question are different. Shared findings therefore support
reuse of a strong strategy together with insights from lower-scoring ones.

\textbf{Parallel branches divide the remaining questions.}
After C1/P3 becomes the leader, all three Coordinators revisit it, targeting
wrapper closure, prompt-example correctness, and executable-answer onset.
Their proposals explicitly account for work already underway, turning a
shared starting point into complementary experiments. These follow-up experiments explore distinct
residual failure modes. Together with the earlier
combinations, this shows how the search alternates between exploring different
data properties and refining a strategy that proves useful across branches.

\section{Agent Instructions and Protocol Walkthrough}
\label{app:agent-instructions}

This appendix presents condensed instructions for each research role and
then follows their handoffs in a data-selection Trial. The boxes summarize
the implemented skills; they are explanatory renderings rather than verbatim
prompts. Appendix~\ref{app:agent-protocols} specifies the corresponding role
contracts and information boundaries.

\subsection{Hypothesis-Formation and Intervention Instructions}

\begin{adepromptbox}[adeviolet]{Coordinator: propose a testable research direction}
Use the supplied Run Memory, Plan catalog, ranking, source evidence and
remaining budget. Propose one research direction and explain how it differs
from existing Plans, including work already underway.

State the hypothesis and proposed mechanism, intervention, expected
observations, fixed controls and rejection criteria. Identify the prior
findings that motivate the proposal and the competing explanations it
should distinguish.

Return a Plan that states whether it explores a new direction, revisits or
contradicts a completed Plan, or combines prior work. Identify the relevant
source Plans and justify this choice from the evidence and available
budget. Leave training execution to the Engine.
\end{adepromptbox}

\begin{adepromptbox}[adeteal]{Builder: implement the planned intervention}
Read the accepted Plan, supplied Plan Memory, authorized data, source
strategies and fixed task conditions. Inspect the available schemas and
examples before implementing the selection, reward, or curriculum logic.
Preserve the target model, training recipe and evaluation procedure.

Return an executable strategy with its implementation rationale, identifying
which components are inherited and which are changed. Use development
feedback to decide whether to finalize or revise it within the same Trial
and revision budget. Keep the Plan's hypothesis and intervention scope fixed
through these revisions; development feedback precedes target-model training.
\end{adepromptbox}

\subsection{Independent Evidence Interpretation Instructions}

The Analyzer receives the Experiment Package without the Plan, its
expectations, or Memory. Between Stages A and B, the Harness validates the
review request and dispatches the Review Worker with the specified records
and rubric. The returned record-level judgments become inputs to Stage~B.

\begin{adepromptbox}{Analyzer, Stage A: design the evidence review}
Inspect the evidence catalog, realized strategy, training measurements and
representative complete records. Form provisional explanations from these
observations.

For each review batch, specify the question, evidence population, selection
scope, observable review criteria and allowed labels. Meet the task-specific
coverage requirements. Return the review design before receiving any
record-level review judgments.
\end{adepromptbox}

\begin{adepromptbox}{Analyzer, Stage~B: interpret the experimental evidence}
Combine the original measurements and behavior records with the returned
review results. Analyze the realized selection, rewards, or curriculum and
relate their observed properties to training and in-loop~performance.

Return an analysis with findings linked to their source evidence. Explain
patterns, differences and contradictions in the observations, and consider
alternative explanations. Use review judgments as advisory evidence. Leave
comparison with the Plan's expectations to the Plan Summarizer.
\end{adepromptbox}

\subsection{Hypothesis Assessment and Memory Update Instructions}

\begin{adepromptbox}[adeviolet]{Plan Summarizer: connect the findings to the research question}
Read the accepted Plan, parent Plan Memory, Trial results and independent
analysis. Compare the expected observations and rejection criteria with the
realized intervention and measured outcomes. Preserve the original research
question when interpreting the result.

Return a complete successor Plan Memory that retains relevant prior
knowledge and integrates the new findings, their sources, contradictions
and next research questions. Include your assessment of how the findings
relate to the hypothesis. Preserve the Harness-provided implementation and
performance-comparison outcomes.
\end{adepromptbox}

\begin{adepromptbox}[adeviolet]{Run Summarizer: integrate findings across Plans}
Read the supplied Run Memory, one accepted Plan Memory update, and the
associated Plan catalog and ranking. Connect the new findings to prior work:
identify agreement, contradiction, reusable strategies and questions for
subsequent proposals.

Return a complete successor Run Memory with source attribution. Preserve
the incoming Plan's findings, hypothesis assessment and implementation and
performance-comparison outcomes. Integrate the supplied update in order;
the Harness publishes the resulting Memory for subsequent research calls.
\end{adepromptbox}

\subsection{End-to-End Protocol Handoffs}

The following data-selection walkthrough connects these instructions across
one Trial. It illustrates the protocol; Appendix~\ref{app:evidence-cases}
analyzes an observed Data Selection--Code research trajectory.
\begin{enumerate}[leftmargin=*,itemsep=0.3em]
\item \textbf{Research question to Plan.} The Coordinator uses the supplied
Run-level findings to propose an intervention. The Harness freezes its
commitment $\psi_i$ and comparison target.
\item \textbf{Plan to executable strategy.} The Builder implements a selector
from that commitment and the supplied data and source strategies. Development
feedback describes the selected examples and diagnostics; bounded revisions
produce the final admitted artifact $z_i$.
\item \textbf{Strategy to experimental evidence.} The Engine trains under
fixed conditions $\mathcal B$, selects a checkpoint using online validation,
and evaluates it at the higher in-loop budget. The Harness packages the
realized selection, training measurements and evaluation records for analysis,
excluding the Plan and its expectations.
\item \textbf{Evidence to findings.} The Analyzer designs the record-level
review. After the Review Worker returns its judgments, the Analyzer combines
them with the original evidence to produce the independent analysis $a_i$.
\item \textbf{Findings to Plan Memory.} The Plan Summarizer compares the
commitment $\psi_i$ with the analysis $a_i$ and measured results. The Harness
admits the resulting findings and publishes the successor Plan Memory.
\item \textbf{Plan Memory to subsequent research.} The Run Summarizer merges
the accepted update into Run Memory, making the findings available to later
proposals. Calls already underway retain their original input versions.
Held-out evaluation remains outside this research loop.
\end{enumerate}

\clearpage
\flushbottom

\section{Selected Strategies and Implementations}
\label{app:selected-artifacts}

The following implementations accompany the selected strategies discussed in
Section~\ref{sec:rq4}. For data selection and reward design, the listings
retain the core decision rules and mark omitted helpers in comments.
The curriculum listing gives the complete scheduling function. Each
subsection explains the strategy before presenting its implementation.

\subsection{Data Selection (Math)}
\label{app:artifact-math-sft}

The selector first applies quality gates, then ranks eligible records by
directness, progress and low redundancy. The same ranking determines which
records survive deduplication.
Task-family labels assess validity, while length bands determine allocation.

Records with at least 16,384 tokens are excluded. Short records have at most
4,096 tokens, medium records have 4,097--8,192, and long records have
8,193--16,383. The selector targets 384 trajectories with short/medium/long
quotas of 156/120/108 and a total of 2,589,626 tokens within 2\%. After the
initial selection, within-band exchanges reduce the token-count difference
without worsening evidence tier. Exchanges may change the finer quality
ranking and stop when the tolerance is met or no improving swap remains.

\begin{adestrategycode}{Data Selection--Math Strategy}
async def select_trajectories(candidate_inventory, select_size, judge):

    # ... Omitted: rubric/schema, record extraction, and batched Judge calls.
    # Records with token_count >= 16384 are excluded; bands are <=4096,
    # <=8192, and <16384. Family labels assess validity, not allocation.

    gate_dimensions = [
        "correctness",
        "coherence_uncorrupted",
        "self_contained",
        "answer_contract",
        "completion",
        "contradiction_free",
    ]

    # ... Omitted: allowed-score/schema validation.

    for index, record in enumerate(records):

        # ... Omitted: unpack Judge evidence; compute complete, family_valid,
        # projected_valid, and scores (including exactly one task family).

        eligible = (
            family_valid
            and all(scores.get(dimension) == 1.0 for dimension in gate_dimensions)
            and scores.get("directness") in [0.5, 1.0]
            and scores.get("progress_fraction", 0.0) > 0.0
            and scores.get("low_redundancy") in [0.5, 1.0]
        )

        if eligible:
            evidence_tier = 0
        elif complete:
            evidence_tier = 1
        else:
            evidence_tier = 2
        record["evidence_tier"] = evidence_tier
        record["scores"] = scores
        record["projected"] = projected if projected_valid else None
        if evidence_tier == 0:
            record["selection_key"] = (
                0,
                -scores["directness"],
                -scores["progress_fraction"],
                -scores["low_redundancy"],
                record["surface_restart_count"],
                record["problem_id"],
                record["trajectory_id"],
            )
        else:
            projected_rank = projected if projected_valid else -1.0
            record["selection_key"] = (
                evidence_tier,
                -scores.get("correctness", -1.0),
                -scores.get("coherence_uncorrupted", -1.0),
                -scores.get("self_contained", -1.0),
                -scores.get("answer_contract", -1.0),
                -scores.get("completion", -1.0),
                -scores.get("contradiction_free", -1.0),
                -scores.get("directness", -1.0),
                -scores.get("progress_fraction", -1.0),
                -scores.get("low_redundancy", -1.0),
                -projected_rank,
                record["problem_id"],
                record["trajectory_id"],
            )

    # The assessment precedes stable-problem and whitespace-exact deduplication,
    # so each duplicate class keeps its highest-ranked assessed representative.
    ranked_records = sorted(records, key=lambda record: record["selection_key"])
    unique_problem_records = []
    seen_problem_ids = set()
    for record in ranked_records:
        if record["problem_id"] in seen_problem_ids:
            continue
        seen_problem_ids.add(record["problem_id"])
        unique_problem_records.append(record)

    unique_records = []
    seen_questions = set()
    for record in unique_problem_records:
        if record["normalized_question"] in seen_questions:
            continue
        seen_questions.add(record["normalized_question"])
        unique_records.append(record)

    band_order = ["A", "B", "C"]
    base_quotas = {"A": 156, "B": 120, "C": 108}

    # ... Omitted: proportional quota scaling for select_size != 384.
    # In the selected run, select_size == 384 and quotas = dict(base_quotas).

    by_band = {band: [] for band in band_order}
    for record in unique_records:
        by_band[record["band"]].append(record)
    for band in band_order:
        by_band[band] = sorted(
            by_band[band], key=lambda record: record["selection_key"]
        )
        for rank_position, record in enumerate(by_band[band]):
            record["rank_position"] = rank_position

    selected_by_band = {}
    for band in band_order:
        selected_by_band[band] = list(by_band[band][: quotas[band]])

    exposure_target = (2589626 * select_size) // 384
    total_tokens = sum(
        record["token_count"]
        for band in band_order
        for record in selected_by_band[band]
    )
    while abs(total_tokens - exposure_target) * 100 > exposure_target * 2:
        current_distance = abs(total_tokens - exposure_target)
        best_exchange = None
        best_key = None
        for band in band_order:
            selected_trajectory_ids = {
                record["trajectory_id"] for record in selected_by_band[band]
            }
            available = [
                record
                for record in by_band[band]
                if record["trajectory_id"] not in selected_trajectory_ids
            ]
            for removed in selected_by_band[band]:
                for added in available:
                    if added["evidence_tier"] > removed["evidence_tier"]:
                        continue
                    revised_total = (
                        total_tokens - removed["token_count"] + added["token_count"]
                    )
                    revised_distance = abs(revised_total - exposure_target)
                    if revised_distance >= current_distance:
                        continue
                    exchange_key = (
                        revised_distance,
                        added["rank_position"],
                        removed["rank_position"],
                        added["problem_id"],
                        added["trajectory_id"],
                    )
                    if best_key is None or exchange_key < best_key:
                        best_key = exchange_key
                        best_exchange = (band, removed, added, revised_total)
        if best_exchange is None:
            break
        band, removed, added, total_tokens = best_exchange
        selected_by_band[band] = [
            added if record is removed else record
            for record in selected_by_band[band]
        ]

    # ... Omitted: sort selected records by problem/trajectory ID within
    # each band and return their trajectory IDs.
\end{adestrategycode}

\subsection{Data Selection (Code)}
\label{app:artifact-code-sft}

This is the in-loop-selected C1/P3 strategy from the Data Selection--Code
case in Appendix~\ref{app:evidence-cases}, called
\emph{Stratified Repetition Control} in Section~\ref{sec:rq4}. The listing
below presents its core selection script; the case study traces how earlier
findings informed its design.

The selector starts from the parent subset of 384 trajectories. It replaces
parent records whose repetition fraction is at least 0.10 with candidates
whose fraction is below 0.05, choosing the nearest token count within the
same prompt subtype, truncation status and length band.

Prompt subtypes are \texttt{named\_callable},
\texttt{stdin\_stdout}, and \texttt{not\_code\_contract}, determined by
textual contract markers. The repetition helper, omitted from the listing,
scans sliding 16-word windows inside each closed \texttt{<think>} region, marks later
occurrences whose starts are at least 16 words after the first occurrence,
and divides the number of marked words by all words in the closed regions.
It returns \texttt{None} when no closed region exists or fewer than 16
words are available. This fraction measures literal repetition. Final assembly
uses a deterministic order and preserves distinct trajectory IDs.

\begin{adestrategycode}{Data Selection--Code Strategy}
async def select_trajectories(candidate_inventory, select_size, judge):

    # ... Omitted: the 384 frozen parent IDs, conversation extraction,
    # prompt-subtype classification, literal repetition measurement,
    # record construction, and a non-decisional Judge probe.

    def selection_block(record):
        token_count = record["token_count"]
        if token_count <= 4096:
            length_band = 0
        elif token_count <= 8192:
            length_band = 1
        else:
            length_band = 2
        will_truncate = 1 if token_count > 16384 else 0
        return record["subtype"], will_truncate, length_band

    parent_records = []
    outside_records = []
    for record in records:
        inherited = record["candidate_id"] in parent_ids
        if not inherited:
            for trajectory_id in record["trajectory_ids"]:
                if trajectory_id in parent_ids:
                    inherited = True
                    break
        if inherited:
            parent_records.append(record)
        else:
            outside_records.append(record)

    selected_by_candidate = {}
    for record in parent_records:
        selected_by_candidate[record["candidate_id"]] = record

    eligible_sources = []
    for record in parent_records:
        fraction = record["repetition_fraction"]
        if fraction is not None and fraction >= 0.10:
            eligible_sources.append(record)
    eligible_sources.sort(
        key=lambda record: (-record["repetition_fraction"], record["candidate_id"])
    )

    candidates_by_block = {}
    for record in outside_records:
        fraction = record["repetition_fraction"]
        if fraction is None or fraction >= 0.05:
            continue
        block = selection_block(record)
        if block not in candidates_by_block:
            candidates_by_block[block] = []
        candidates_by_block[block].append(record)

    used_replacements = {}
    for source in eligible_sources:
        block = selection_block(source)
        best = None
        best_key = None
        for candidate in candidates_by_block.get(block, []):
            if candidate["candidate_id"] in used_replacements:
                continue
            candidate_key = (
                abs(candidate["token_count"] - source["token_count"]),
                candidate["candidate_id"],
            )
            if best is None or candidate_key < best_key:
                best = candidate
                best_key = candidate_key
        if best is not None:
            del selected_by_candidate[source["candidate_id"]]
            selected_by_candidate[best["candidate_id"]] = best
            used_replacements[best["candidate_id"]] = True

    # ... Omitted: deterministic output assembly and shortfall filling.
    # Sort chosen candidates by ID; reject overlapping trajectory IDs
    # and groups that would exceed select_size (never split a group).
    # If short, scan remaining records parent-ID-first, then token count
    # and candidate ID, applying the same constraints; return selected_ids.
\end{adestrategycode}

\clearpage
\subsection{Reward Design (Math)}
\label{app:artifact-math-rft}

The reward strategy combines verifier outcomes with process quality,
computed from derivation soundness (0.50), conclusion support (0.35), and
substantive relevance (0.15). Process-based shaping operates on groups of
sibling responses.

For an all-wrong group, a response is eligible only when all three process
dimensions receive full scores. Shaping requires at least half the siblings
(rounded up) to be eligible; an all-eligible tie retains verifier outcomes.
For mixed groups, quality is centered separately within correct and incorrect
responses, preserving their outcome separation. The listing retains the
group-specific thresholds, tie handling and reward formulas.

\begin{adestrategycode}{Reward Design--Math Strategy}
import math

_PROCESS_DIMENSIONS = (
    "conclusion_support",
    "derivation_soundness",
    "substantive_relevance",
)
_PROCESS_WEIGHTS = {
    "conclusion_support": 0.35,
    "derivation_soundness": 0.50,
    "substantive_relevance": 0.15,
}

# ... Omitted: _outcome validates the verifier score; row-level normal
# and fallback entry points return that outcome unchanged, with no rule signal.

def _process_quality(record: dict) -> float:
    dimensions = record["evidence"]["process"]["dimensions"]
    quality = 0.0
    for name in _PROCESS_DIMENSIONS:
        value = float(dimensions[name])
        if not math.isfinite(value) or not 0.0 <= value <= 1.0:
            raise ValueError(f"process dimension {name!r} must be within [0.0, 1.0]")
        quality += _PROCESS_WEIGHTS[name] * value
    return quality


def _is_absolutely_eligible(record: dict) -> bool:
    dimensions = record["evidence"]["process"]["dimensions"]
    return all(float(dimensions[name]) == 1.0 for name in _PROCESS_DIMENSIONS)

# ... Omitted: _group_result packages record IDs, rewards, and decision
# metadata into the ADE output schema; it does not alter rewards.

def assign_group_credit(group_input: dict) -> dict:
    if group_input["schema_version"] != "ade.group_credit_input.v3":
        raise ValueError("unsupported group-credit input schema")

    records = group_input["records"]
    outcomes = [_outcome(record["outcome_score"]) for record in records]
    process_available = all(
        record["evidence"]["process"]["status"] == "available"
        for record in records
    )
    if not process_available:
        return _group_result(
            records,
            outcomes,
            "abstained",
            ["outcome"],
            [],
            "process_unavailable",
        )

    qualities = [_process_quality(record) for record in records]
    correct = [bool(record["is_correct"]) for record in records]

    if group_input["group_type"] == "all_wrong":
        eligible = [_is_absolutely_eligible(record) for record in records]
        eligible_count = sum(eligible)
        corroboration_count = (len(records) + 1) // 2
        if eligible_count < corroboration_count:
            return _group_result(
                records,
                outcomes,
                "abstained",
                ["outcome", "process"],
                list(_PROCESS_DIMENSIONS),
                "insufficient_sibling_corroboration",
            )
        if eligible_count == len(records):
            return _group_result(
                records,
                outcomes,
                "identity",
                ["outcome", "process"],
                list(_PROCESS_DIMENSIONS),
                "eligible_process_tie",
            )
        rewards = [0.75 if is_eligible else 0.25 for is_eligible in eligible]
        return _group_result(
            records,
            rewards,
            "shaped",
            ["outcome", "process"],
            list(_PROCESS_DIMENSIONS),
            "sibling_corroborated_eligibility",
        )

    if group_input["group_type"] == "mixed":
        correct_qualities = [q for q, is_correct in zip(qualities, correct) if is_correct]
        wrong_qualities = [q for q, is_correct in zip(qualities, correct) if not is_correct]
        correct_varies = max(correct_qualities) > min(correct_qualities)
        wrong_varies = max(wrong_qualities) > min(wrong_qualities)
        if not correct_varies and not wrong_varies:
            return _group_result(
                records,
                outcomes,
                "identity",
                ["outcome", "process"],
                list(_PROCESS_DIMENSIONS),
                "process_tie",
            )

        correct_mean = math.fsum(sorted(correct_qualities)) / len(correct_qualities)
        wrong_mean = math.fsum(sorted(wrong_qualities)) / len(wrong_qualities)
        process_scale = 0.4 / len(records)
        rewards = [
            (0.75 + process_scale * (quality - correct_mean))
            if is_correct
            else (0.25 + process_scale * (quality - wrong_mean))
            for quality, is_correct in zip(qualities, correct)
        ]
    else:
        if max(qualities) == min(qualities):
            return _group_result(
                records,
                outcomes,
                "identity",
                ["outcome", "process"],
                list(_PROCESS_DIMENSIONS),
                "process_tie",
            )
        rewards = [0.25 + 0.5 * quality for quality in qualities]

    return _group_result(
        records,
        rewards,
        "shaped",
        ["outcome", "process"],
        list(_PROCESS_DIMENSIONS),
        "process_quality_separation",
    )
\end{adestrategycode}

\subsection{Curriculum Learning (Math)}
\label{app:artifact-curriculum}

\textbf{Foundation--bridge--challenge cycling with paired reinforcement.}
The schedule fixes 100 batches of 256 distinct problems before training.
Canonical levels 0--2 form the foundation band, level 3 the bridge band,
and levels 4--5 the challenge band. Rotating four-step blocks produce
25 foundation, 50 bridge, and 25 challenge batches. Foundation and bridge
selection prioritizes unseen problems, then less-used problems, with a
fixed tie order.

After every eligible problem has appeared, an adjacent foundation--bridge
pair reserves 36 positions in each batch for the same subject. These pairs
cycle through Counting \& Probability, Geometry, Intermediate Algebra, and
Precalculus; the remaining 220 positions come from other subjects in the
corresponding band. The first eight challenge batches contain 192 level-4
and 64 level-5 problems; the remaining 17 contain 128 of each. Challenge
selection preserves the inherited template's subject totals and favors
less-used problems within each subject--level cell.

The complete implementation is reproduced below.

\begin{adestrategycode}{Curriculum Learning--Math: complete implementation}
async def build_curriculum(
    candidate_inventory, total_steps, prompts_per_step, judge_batch
):
    band_for_level = {
        0: "foundation",
        1: "foundation",
        2: "foundation",
        3: "bridge",
        4: "challenge",
        5: "challenge",
    }
    band_ids = {
        "foundation": [],
        "bridge": [],
        "challenge": [],
    }
    cell_ids = {}
    subject_by_id = {}

    ordered_inventory = sorted(
        candidate_inventory,
        key=lambda item: (item["source_order"], item["problem_id"]),
    )
    for item in ordered_inventory:
        problem_id = item["problem_id"]
        band = band_for_level[item["level"]]
        band_ids[band].append(problem_id)
        subject_by_id[problem_id] = item["subject"]
        if item["level"] in (4, 5):
            cell = (item["subject"], item["level"])
            cell_ids.setdefault(cell, []).append(problem_id)

    # Expand the fixed Harness root seed into one candidate tie rank.
    tie_order = [item["problem_id"] for item in ordered_inventory]
    generator_state = 42
    for index in range(len(tie_order) - 1, 0, -1):
        generator_state = (
            1664525 * generator_state + 1013904223
        ) % 4294967296
        swap_index = generator_state % (index + 1)
        tie_order[index], tie_order[swap_index] = (
            tie_order[swap_index],
            tie_order[index],
        )
    tie_rank = {
        problem_id: rank for rank, problem_id in enumerate(tie_order)
    }

    subjects_by_band = {}
    for band, problem_ids in band_ids.items():
        subjects_by_band[band] = sorted(
            {subject_by_id[problem_id] for problem_id in problem_ids}
        )

    block_types = (
        ("foundation", "bridge", "challenge", "bridge"),
        ("bridge", "challenge", "bridge", "foundation"),
        ("challenge", "bridge", "foundation", "bridge"),
        ("bridge", "foundation", "bridge", "challenge"),
    )
    step_bands = []
    for step_index in range(total_steps):
        block_index = step_index // 4
        position_in_block = step_index % 4
        step_bands.append(
            block_types[block_index % len(block_types)][position_in_block]
        )

    def select_ids(band, slots, visits, allowed_subjects=None):
        if allowed_subjects is None:
            allowed_subjects = subjects_by_band[band]
        allowed_subjects = list(allowed_subjects)
        allowed_set = set(allowed_subjects)
        available = [
            problem_id
            for problem_id in band_ids[band]
            if subject_by_id[problem_id] in allowed_set
        ]
        chosen = []
        chosen_set = set()

        unseen = [
            problem_id
            for problem_id in available
            if visits[problem_id] == 0
        ]
        if unseen:
            for subject in allowed_subjects:
                subject_unseen = [
                    problem_id
                    for problem_id in unseen
                    if subject_by_id[problem_id] == subject
                ]
                if subject_unseen and len(chosen) < slots:
                    selected = min(
                        subject_unseen,
                        key=lambda problem_id: tie_rank[problem_id],
                    )
                    chosen.append(selected)
                    chosen_set.add(selected)

            remaining_unseen = sorted(
                (
                    problem_id
                    for problem_id in unseen
                    if problem_id not in chosen_set
                ),
                key=lambda problem_id: tie_rank[problem_id],
            )
            for problem_id in remaining_unseen:
                if len(chosen) == slots:
                    break
                chosen.append(problem_id)
                chosen_set.add(problem_id)

        if len(chosen) < slots:
            represented = {
                subject_by_id[problem_id] for problem_id in chosen
            }
            for subject in allowed_subjects:
                if subject in represented or len(chosen) == slots:
                    continue
                subject_candidates = [
                    problem_id
                    for problem_id in available
                    if subject_by_id[problem_id] == subject
                    and problem_id not in chosen_set
                ]
                selected = min(
                    subject_candidates,
                    key=lambda problem_id: (
                        visits[problem_id],
                        tie_rank[problem_id],
                    ),
                )
                chosen.append(selected)
                chosen_set.add(selected)

            remaining = sorted(
                (
                    problem_id
                    for problem_id in available
                    if problem_id not in chosen_set
                ),
                key=lambda problem_id: (
                    visits[problem_id],
                    tie_rank[problem_id],
                ),
            )
            chosen.extend(remaining[: slots - len(chosen)])

        return chosen

    # Recover the inherited template's exact subject totals at each challenge
    # step. Challenge visits are independent of the other two bands.
    template_visits = {
        problem_id: 0 for problem_id in band_ids["challenge"]
    }
    challenge_subject_totals = {}
    for step_index, band in enumerate(step_bands):
        if band != "challenge":
            continue
        template_ids = select_ids(
            "challenge", prompts_per_step, template_visits
        )
        subject_counts = {
            subject: sum(
                1
                for problem_id in template_ids
                if subject_by_id[problem_id] == subject
            )
            for subject in subjects_by_band["challenge"]
        }
        challenge_subject_totals[step_index] = subject_counts
        for problem_id in template_ids:
            template_visits[problem_id] += 1

    challenge_subjects = subjects_by_band["challenge"]
    subject_tie_rank = {
        subject: min(
            tie_rank[problem_id]
            for problem_id in band_ids["challenge"]
            if subject_by_id[problem_id] == subject
        )
        for subject in challenge_subjects
    }

    schedule_visits = {
        problem_id: 0
        for problem_ids in band_ids.values()
        for problem_id in problem_ids
    }
    remediation_subjects = (
        "Counting & Probability",
        "Geometry",
        "Intermediate Algebra",
        "Precalculus",
    )
    remediation_index = 0
    pending_bridge_subject = None
    challenge_ordinal = 0
    schedule = []

    for step_index, band in enumerate(step_bands):
        if band == "challenge":
            challenge_ordinal += 1
            if challenge_ordinal <= 8:
                level_four_numerator = 3
                denominator = 4
            else:
                level_four_numerator = 1
                denominator = 2

            subject_counts = challenge_subject_totals[step_index]
            level_four_target = (
                prompts_per_step * level_four_numerator // denominator
            )
            level_four_by_subject = {
                subject: (
                    subject_counts[subject]
                    * level_four_numerator
                    // denominator
                )
                for subject in challenge_subjects
            }
            unassigned = level_four_target - sum(
                level_four_by_subject.values()
            )
            remainder_order = sorted(
                challenge_subjects,
                key=lambda subject: (
                    -(
                        subject_counts[subject]
                        * level_four_numerator
                        % denominator
                    ),
                    subject_tie_rank[subject],
                ),
            )
            for subject in remainder_order[:unassigned]:
                level_four_by_subject[subject] += 1

            step_ids = []
            for subject in challenge_subjects:
                cell_targets = (
                    (4, level_four_by_subject[subject]),
                    (
                        5,
                        subject_counts[subject]
                        - level_four_by_subject[subject],
                    ),
                )
                for level, slots in cell_targets:
                    selected = sorted(
                        cell_ids[(subject, level)],
                        key=lambda problem_id: (
                            schedule_visits[problem_id],
                            tie_rank[problem_id],
                        ),
                    )[:slots]
                    step_ids.extend(selected)
        else:
            target_subject = None
            if pending_bridge_subject is not None:
                target_subject = pending_bridge_subject
                pending_bridge_subject = None
            elif (
                band == "foundation"
                and step_index + 1 < total_steps
                and step_bands[step_index + 1] == "bridge"
                and all(
                    schedule_visits[problem_id] > 0
                    for problem_id in schedule_visits
                )
            ):
                target_subject = remediation_subjects[
                    remediation_index % len(remediation_subjects)
                ]
                remediation_index += 1
                pending_bridge_subject = target_subject

            if target_subject is None:
                step_ids = select_ids(
                    band,
                    prompts_per_step,
                    schedule_visits,
                )
            else:
                target_ids = sorted(
                    (
                        problem_id
                        for problem_id in band_ids[band]
                        if subject_by_id[problem_id] == target_subject
                    ),
                    key=lambda problem_id: (
                        schedule_visits[problem_id],
                        tie_rank[problem_id],
                    ),
                )[:36]
                other_subjects = [
                    subject
                    for subject in subjects_by_band[band]
                    if subject != target_subject
                ]
                other_ids = select_ids(
                    band,
                    prompts_per_step - 36,
                    schedule_visits,
                    other_subjects,
                )
                step_ids = target_ids + other_ids

        for problem_id in step_ids:
            schedule_visits[problem_id] += 1
        schedule.append(sorted(step_ids))

    return schedule
\end{adestrategycode}

\end{document}